\documentclass{article}
\usepackage[preprint]{colm2025_conference}

\usepackage{microtype}
\usepackage{hyperref}
\usepackage{url}
\usepackage{booktabs}
\usepackage{graphicx}
\usepackage{amsmath}
\usepackage{multirow}
\usepackage{subcaption}
\usepackage{wrapfig}

\usepackage{lineno}

\definecolor{darkblue}{rgb}{0, 0, 0.5}
\hypersetup{colorlinks=true, citecolor=darkblue, linkcolor=darkblue, urlcolor=darkblue}

\title{Triple Phase Transitions: Understanding the Learning \\Dynamics of Large Language Models from a Neuroscience Perspective}

\author{Yuko Nakagi\textsuperscript{1,2}\thanks{Equal first author.}  \thanks{Team lead.} , Keigo Tada\textsuperscript{1,2}\footnotemark[1] , Sota Yoshino\textsuperscript{1,2}\footnotemark[1] , Shinji Nishimoto\textsuperscript{1,2}\thanks{Equal last author.} , Yu Takagi\textsuperscript{1,2,3}\footnotemark[3]
\\
\\
\textsuperscript{1}Osaka University, Japan\\
\textsuperscript{2}National Institute of Information and Communications Technology, Japan\\
\textsuperscript{3}National Institute of Informatics, Japan\\\\
\textbf{Correspondence:} \href{mailto:email@domain}{nishimoto.shinji.fbs@osaka-u.ac.jp, yutakagi322@gmail.com}
}

\begin{document}

\ifcolmsubmission
\linenumbers
\fi

\maketitle

\begin{abstract}
\label{sec:abstract}

Large language models (LLMs) often exhibit abrupt emergent behavior, whereby new abilities arise at certain points during their training. This phenomenon, commonly referred to as a ``phase transition'', remains poorly understood. In this study, we conduct an integrative analysis of such phase transitions by examining three interconnected perspectives: the similarity between LLMs and the human brain, the internal states of LLMs, and downstream task performance. We propose a novel interpretation for the learning dynamics of LLMs that vary in both training data and architecture, revealing that three phase transitions commonly emerge across these models during training: (1) alignment with the entire brain surges as LLMs begin adhering to task instructions (\textit{Brain Alignment and Instruction Following}), (2) unexpectedly, LLMs diverge from the brain during a period in which downstream task accuracy temporarily stagnates (\textit{Brain Detachment and Stagnation}), and (3) alignment with the brain reoccurs as LLMs become capable of solving the downstream tasks (\textit{Brain Realignment and Consolidation}). These findings illuminate the underlying mechanisms of phase transitions in LLMs, while opening new avenues for interdisciplinary research bridging AI and neuroscience.

\end{abstract}
    
\section{Introduction}
\label{sec:introduction}

Large language models (LLMs) often exhibit abrupt emergent behaviors, whereby new abilities arise from scaling the model size, the amount of training data, or the number of training steps \citep{Wei2022-gq}. This behavior has predominantly been discovered through evaluations of model outputs (e.g., downstream task performance), leading to numerous breakthroughs \citep{wei2022chain, caballero2023broken, du2024understanding}. Recent efforts in mechanistic interpretability have begun to uncover the underlying internal changes that occur during training process, aiming to elucidate these emergent phenomena \citep{Olsson2022-of, Chen2023-wl}. However, prior studies have largely examined these transitions in isolated perspectives, making it unclear how these transitions, along with other additional aspects, interact with each other.

In seeking to clarify the underlying principles of deep neural networks (DNNs), neuroscience research has yielded human-centered insights into DNNs through comparisons between their internal representations and human brain activity \citep{doi:10.1073/pnas.1403112111, G10005}. This approach has recently been extended to LLMs \citep{Jain2018-st, Goldstein2022-yw, Oota2022-gd}, revealing that brain activity has a stronger alignment with larger- and higher-performance models~\citep{Antonello2023-pi}. These findings highlight the potential of the human brain, an intricate system with diverse functions, to serve as a tool for model evaluation. Because most existing studies have only examined trained models, however, they offer limited insight into how emergent phenomena arise during the learning process.

In this study, we focus on the learning dynamics of LLMs in an attempt to provide a comprehensive understanding of emergent phenomena by integrating three analytical perspectives: \textit{brain encoding} analysis, which assesses alignment with human brain activity; \textit{probing} analysis, which detects shifts in internal representations; and \textit{benchmark} analysis, which measures downstream task performance\footnote{All experimental resources (source code, training data, and model checkpoints) will be publicly available to promote reproducibility and future research. See Sections~\ref{appendix:additional_methods:llms} and \ref{appendix:additional_methods:brain-encoding-model} for the details.}. We demonstrate that multiple LLMs, each characterized by distinct architectures and training data, exhibit a robust, common three-stage phase transition in their learning dynamics, including the precise timing of these transitions. Specifically, (1) alignment with the entire brain surges once the LLMs begin to follow downstream task instructions (\textit{\textbf{Brain Alignment and Instruction Following}}); (2) surprisingly, the LLMs diverge from the brain during a period in which their downstream task accuracy temporarily stagnates (\textit{\textbf{Brain Detachment and Stagnation}}); and (3) alignment with the brain reoccurs once LLMs become capable of solving the downstream tasks (\textit{\textbf{Brain Realignment and Consolidation}}). Although prior work has reported stronger brain alignment with larger and higher performance models~\citep{Antonello2023-pi}, our findings reveal that the learning trajectory of LLMs is more complex than previously thought, consisting of multiple phases. By examining multiple LLMs trained on distinct language datasets, we highlight the influence of the training data on these learning dynamics. Taken together, our findings demonstrate that incorporating human brain activity as a biologically grounded benchmark reveals how the emergent capabilities of LLMs form and consolidate during training, offering essential insights for safer, more interpretable language models.

\begin{figure}[t]
\begin{center}
\includegraphics[width=1.0\linewidth]{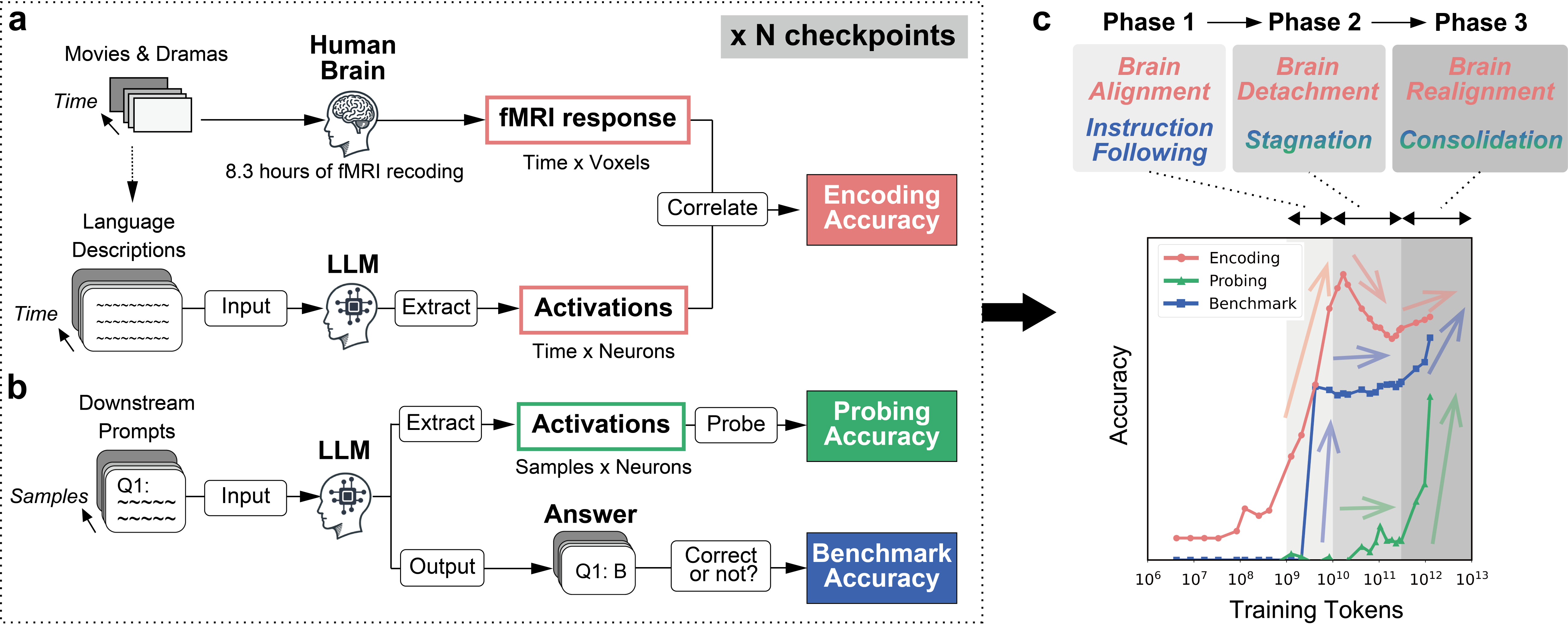}
\caption{\textbf{Overview of the study.} \textbf{a} \textit{Brain encoding} analysis. \textbf{b} \textit{Probing} analysis (top) and \textit{benchmark} analysis (bottom). \textbf{c} Three phase-transition phenomena during the learning process of LLMs, as identified through the results of the encoding, probing, and benchmark analyses. Red, green, and blue lines indicate encoding, probing, and benchmark accuracies, respectively, across all LLM checkpoints.}
\label{fig:methods-overview}
\end{center}
\end{figure}

\section{Related work}
\label{sec:related}

\paragraph{Phase transitions in LLMs}

LLMs acquire distinct abilities throughout their training process as their model size becomes larger, the amount of training data rises, and the number of training steps increases. While some abilities emerge gradually \citep{Kaplan2020-ul, Hoffmann2022-ye}, others appear abruptly \citep{10.1145/3531146.3533229, Wei2022-gq}. Previous studies have identified ``emergent abilities'', such as the onset of chain-of-thought prompting beyond a certain scale \citep{Wei2022-gq}. To understand LLMs internally, mechanistic interpretability seeks to unveil the internal computations learned by neural networks \citep{elhage2021mathematical, dai-etal-2022-knowledge}. Internal phase transitions in the learning process have also recently been observed in this field \citep{Olsson2022-of, Chen2023-wl, park2024emergencehiddencapabilitiesexploring}, shedding light on how abrupt performance gains align with shifts in internal representations. For example, \citet{Olsson2022-of} found ``phase changes'' in Transformers, where induction heads suddenly arise and enable extended contexts to be handled through in-context learning. However, prior studies have largely focused on transitions in isolated perspectives, leaving it unclear how these shifts, alongside other relevant aspects, interact.

\paragraph{Interpretability from neuroscience insights}
From a neuroscience perspective, related work has compared the internal representations of DNNs with human brain activity \citep{doi:10.1073/pnas.1403112111, G10005}, demonstrating that the hierarchical structure of high-performing DNNs mirrors the hierarchical processing of the human visual cortex. More recent efforts have employed linear mappings from LLM internal representations to brain activity, probing language processing similarities between LLMs and the human brain \citep{Jain2018-st, Schrimpf2021-vr, Goldstein2022-yw, Oota2022-gd}. Notable examples include comparisons of learning characteristics between LLMs and the human brain \citep{millet2022toward, Caucheteux2022, Caucheteux2023-az, Aw2023-zc, aw2023training, Antonello2023-pi, annurev:/content/journals/10.1146/annurev-neuro-120623-101142, Tuckute2024,  10.1162/nol_a_00087, de-Varda2025.02.01.636044} and discussions of the changes in latent dimensionality within LLMs \citep{Cheng2024-nu}. 
\section{Methods}
\label{sec:methods}

\subsection{Large language models}
\label{sec:methods-llms}
We analyze the learning dynamics of LLMs from three distinct perspectives. For this, we use four pre-trained models with available training checkpoints: OLMo-2 \citep{olmo20242olmo2furious}, OLMo-0724 \citep{Groeneveld2023OLMo}, and LLM-jp \citep{llmjp2024llmjpcrossorganizationalprojectresearch} for the main analysis, and Amber \citep{Liu2023-ky} for the control analysis. The number of checkpoints used for each model ranges from 18 to 28. Table \ref{tab:method-llms} presents an overview of these LLMs. OLMo-2, OLMo-0724, and Amber are English-centric LLMs trained on publicly accessible datasets, while LLM-jp is a bilingual Japanese-English model trained on a roughly equal mix of Japanese and English. Each model uses a different training corpus and tokenizer, with vocabulary sizes ranging from 32,000 to 100,352. These LLMs have between 6.74–7.3B parameters, 32 hidden layers, and a hidden dimension of 4,096. They are all based on a decoder-only Transformer architecture \citep{NIPS2017_3f5ee243}, but each LLM incorporates a few critical modifications. There are also notable differences across these LLMs in terms of their layer normalization and attention mechanisms. See Section \ref{appendix:additional_methods:llms} for further details on the models used. The multilayer perceptron (MLP) layers require most parameters and represent essential features \citep{bereska2024mechanisticinterpretabilityaisafety, geva-etal-2021-transformer}. Thus, we use their neural activation as the internal representations of each model.

\subsection{Brain encoding models}
\label{sec:methods-encoding}
Our initial approach for analyzing the learning dynamics of the LLMs involves an investigation of  how their activations progressively align with brain activity during the training process. Specifically, for each checkpoint described in Sections \ref{sec:methods-llms} and \ref{appendix:additional_methods:llms}, we perform an encoding analysis by evaluating the prediction accuracy of a learned linear mapping from each layer's activations to brain activity~\citep{Naselaris2011-wm, Nishimoto2011-uc, Huth2012-bh} (Figure \ref{fig:methods-overview}a). These analyses are conducted separately for each participant.

\paragraph{fMRI datasets}
\label{sec:methods-fmri-datasets}
We use the Narrative Movie fMRI Dataset \citep{Yamaguchi2024-is, nakagi-etal-2024-unveiling} for our brain encoding analysis. This dataset provides brain activity data from six healthy participants with normal or corrected-normal vision (three females; ages 22–40, mean = 28.7), while they freely watched 8.3 hours of movies or drama series inside a 3T functional Magnetic Resonance Imaging (fMRI) scanner. All participants were native Japanese speakers. The dataset includes nine video clips of movies or drama series as stimuli: eight international titles and one Japanese title. The international titles were dubbed into Japanese, allowing the participants to view all clips in Japanese. The dataset also contains three types of natural language annotation from the videos. We use the \textit{Narrative Content} annotation, which describes the background story of the scene at 5-s intervals, for the main analysis and the \textit{Objective Information} annotation, which describes the objects in the scene every second, for the control analysis, both in English and Japanese. See Section \ref{appendix:additional_methods:fmri_datasets} for additional dataset details. In this study, we use 29,993 seconds of data covering all six participants. We divide the data into training and test datasets. All prediction performance results are computed using the test dataset. Specifically, we use the fMRI scanning sessions corresponding to the last split of each movie or drama series (7,737 seconds in total) as the test data. The remainder of the sessions (22,262 seconds in total) forms the training data.

\paragraph{Model construction}
\label{sec:methods-brain-encoding-models}
We extract the activations of the LLMs for the annotations from each hidden layer. They consist of several tokens for each time point. We then average the activations across tokens. Because multiple annotations exist for each second, we extract the activations for each annotator and then average them across all annotators. We then train an L2-regularized linear model to predict the voxel-level brain activity from the corresponding activations of the LLM neurons. We estimate the model weights from the training data, then apply them to the test data. The regularization parameters are determined for each voxel by cross-validation during training. We then evaluate the prediction accuracy by computing the Pearson's correlation coefficients between the predicted and measured fMRI signals. Statistical significance is assessed using a blockwise permutation test that compares the correlations between predicted and measured signals against the correlations obtained after shuffling the measured signals. We set the threshold for statistical significance to $p < 0.05$ and correct for multiple comparisons using the FDR procedure. We model the hemodynamic delay in the BOLD signal, assumed to be 8–10 seconds. See Section \ref{appendix:additional_methods:brain-encoding-model} for additional details about model construction and region-of-interest (ROI) analyses. In the main analysis, we focus on later layers within each LLM. Specifically, we focus on layer 25 in both OLMo-2 and LLM-jp, and layer 30 in OLMo-0724, which show the greatest checkpoint-wise changes and are strongly involved in each model's emergent behavior across all evaluations (encoding, probing, and benchmark analyses; see Section \ref{appendix:additional_methods:select-layer} for details). We confirm that the layers surrounding the main focal layer exhibit similar patterns of results (see Figure \ref{sec:additional_results:select-layer}).

\subsection{Probing}
\label{sec:methods-probing}
As the second approach, we investigate how downstream task-relevant representations are progressively acquired within the LLMs during the training process. To this end, we perform a probing analysis by evaluating how well linear models can predict neuron activations from the answer labels of downstream tasks across all hidden layers (Figure~\ref{fig:methods-overview}b, top). These analyses are conducted separately for each LLM checkpoint.

\paragraph{Downstream datasets}
\label{sec:methods-downstream-datasets}
We use four downstream tasks: Massive Multitask Language Understanding (MMLU)~\citep{Hendrycks2020-ev}, CommonsenseQA (CSQA)~\citep{Talmor2019-or}, AI2 Reasoning Challenge (ARC)~\citep{allenai:arc}, and HellaSwag~\citep{zellers2019hellaswag}. For both the probing and benchmark analyses (see Section~\ref{sec:methods-benchmark}), we use 5-shot prompts in both English and Japanese. We exclude prompts exceeding each model's maximum context length. See Section \ref{appendix:additional_methods:Downstream_datasets} for additional details about downstream datasets and Section~\ref{appendix:additional_methods:prompts-examples} for examples of the prompts used in these datasets. In the probing analysis, each dataset is split into training and test datasets at a 4:1 ratio. All prediction performance results are computed using the test dataset.

\paragraph{Probing for MLP activations}
\label{sec:methods-brain-inspired-Probing}
We first feed the downstream task prompts into the LLMs and extract the final-token neuron activations in each layer. For each sample in the downstream tasks, we create an answer matrix (the number of samples \(\times\) the number of choices), where the correct and incorrect choices are labeled as 1 and 0, respectively. Our objective is to learn a mapping from this answer matrix to the activations across all layers of the LLMs. To this end, we use L2-regularized linear regression to estimate the linear weights that transform the answer matrix corresponding to the training data into the observed activations in the LLMs, and then apply these learned weights to the test data. The regularization parameter is optimized via 4-fold cross-validation for each neuron in the training dataset. We then evaluate the prediction accuracy by computing the Pearson's correlation coefficients between the predicted and actual activations. The amount of information necessary for each task that is retained at each LLM checkpoint is quantitatively evaluated at the neuron level. In this study, we focus on the same layers as in the encoding analysis (Section~\ref{sec:methods-encoding}). 

\subsection{Benchmark}
\label{sec:methods-benchmark}
As the third approach for analyzing the learning dynamics of LLMs, we investigate how the ability to solve downstream tasks emerges as the training process continues. Specifically, we use each LLM checkpoint to evaluate performance on the downstream task datasets from Section~\ref{sec:methods-downstream-datasets} (Figure~\ref{fig:methods-overview}b, bottom). The evaluation metric is the ratio of correctly answered items, i.e., the fraction of samples for which the model outputs exactly match the correct answers. We use the \textit{llm-jp-eval} library\footnote{\url{https://github.com/llm-jp/llm-jp-eval}} for this analysis.

\subsection{Direct analysis of MLP activations}
\label{sec:methods-activations-weights-analysis}
To investigate how internal representations in LLMs change over the course of training, we compare their properties across checkpoints using the correlation and dimensionality of activations. We compute the Pearson's correlation coefficients of the activations across LLM checkpoints. Specifically, we use the same dataset as for the encoding analysis, extract the activations from the same layers, and calculate their correlations across different checkpoints. We also quantify the dimensionality of the activations at each checkpoint and examine changes across checkpoints. Here, we adopt the Intrinsic Dimension (ID), which has garnered attention as a means of characterizing the nature of LLM internal representations. We estimate the IDs at each checkpoint by applying the Generalized Ratios Intrinsic Dimension Estimator (GRIDE)~\citep{Denti2022} to each layer's activations of the same dataset used for the encoding analysis (see Section~\ref{appendix:activaions-dimensionality-analysis} for details).
\section{Results}

\subsection{Learning dynamics of LLMs}
\label{sec:results-dynamics}

\begin{figure}[t]
\centering
\includegraphics[width=1.0\linewidth]{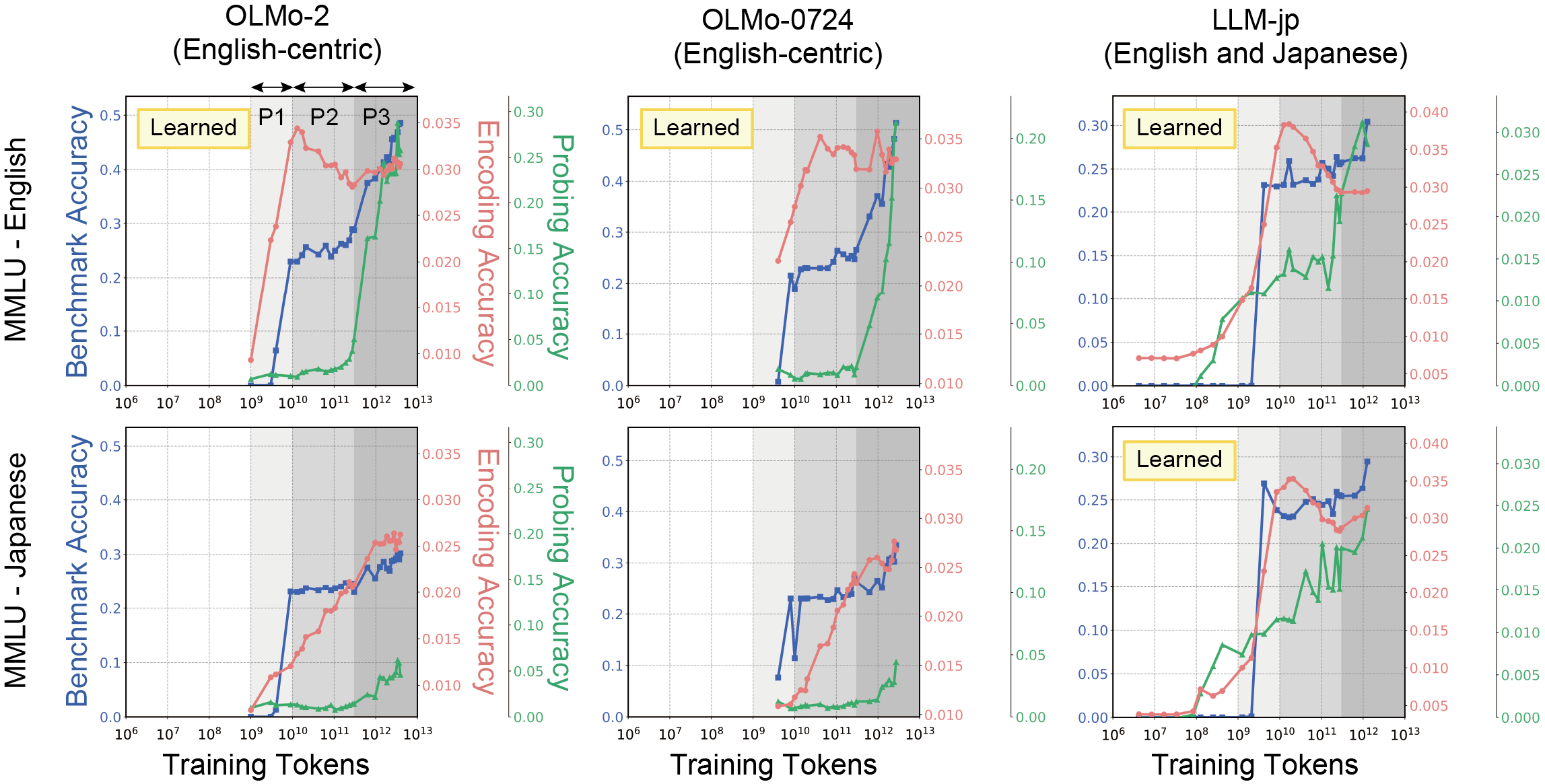}
\caption{\textbf{Learning dynamics of LLMs exhibiting three phase transitions.} The horizontal axis denotes the number of training tokens. The vertical axis denotes the average encoding accuracy for all voxels of a single participant (DM06) (red lines), the benchmark accuracy (blue lines), and the average probing accuracy for all LLM neurons calculated using MMLU (green lines). We select layers 25, 30, and 25 from OLMo-2, OLMo-0724, and LLM-jp, respectively, to capture the transitions that occur at each phase of the learning dynamics. The background color indicates the LLM phase. The legend indicates whether the language has been learned sufficiently by the model. No checkpoints preceding the 10\textsuperscript{9} training tokens have been made publicly available aside from LLM-jp.}
\label{fig:result-phase-transtions}
\end{figure}

We investigate how each accuracy metric evolves over the course of training using the three analytical approaches from Sections~\ref{sec:methods-encoding}, \ref{sec:methods-probing}, and \ref{sec:methods-benchmark}. Figure~\ref{fig:result-phase-transtions} clearly shows that later layers of the LLMs have three phase transitions during training. The first phase transition emerges after around 10\textsuperscript{9}--10\textsuperscript{10} training tokens, where both the encoding and benchmark accuracy suddenly surge. In particular, the benchmark accuracy (blue line) reveals that the model's ability to perform downstream tasks improves, suggesting that the LLMs begin following task instructions. Simultaneously, the brain encoding accuracy (red line) reveals enhanced overall alignment of the LLMs with the entire brain (see Section~\ref{sec:results-encoding}). We refer to this first phase as the \textit{\textbf{Brain Alignment and Instruction Following}} Phase. The second phase transition arises after around 10\textsuperscript{10}--$3\cdot10^{11}$ training tokens, where the benchmark accuracy stagnates. Strikingly, in this phase transition, the brain encoding accuracy \textbf{declines}, indicating reduced alignment between the LLMs and the brain (see Section~\ref{sec:results-encoding}). Accordingly, we label this second phase as the \textit{\textbf{Brain Detachment and Stagnation}} Phase. The third phase transition occurs beyond approximately $3\cdot10^{11}$ training tokens, where the benchmark and probing accuracy increase sharply, accompanied by a slight upward trend in the brain encoding accuracy. At this point, the increase in the benchmark accuracy suggests that the LLMs gradually acquire the capability to solve tasks, whereas the change in the brain encoding accuracy implies a renewed enhancement in alignment with the brain (see Section~\ref{sec:results-encoding}). We thus refer to this third phase as the \textit{\textbf{Brain Realignment and Consolidation}} Phase. Notably, these dynamics are only observed when the LLMs process a language that they have learned sufficiently: English and Japanese for LLM-jp, and English only for OLMo-2 and OLMo-0724.

Figure~\ref{fig:result-phase-transtions} presents the results for OLMo-2, OLMo-0724, and LLM-jp when using MMLU for the probing and benchmark analyses. The brain encoding result shows the average prediction accuracy across all brain voxels inferred from LLM neurons in the specified layer, whereas the probing result shows the average prediction accuracy across all LLM neurons. Section~\ref{appendix:additional_results:dyamics} provides analogous results for adjacent layers, other downstream tasks (CSQA, ARC, HellaSwag), other LLMs (Amber), other languages (Chinese), and other annotations.

\begin{figure*}[t]
\centering
\includegraphics[width=1.0\linewidth]{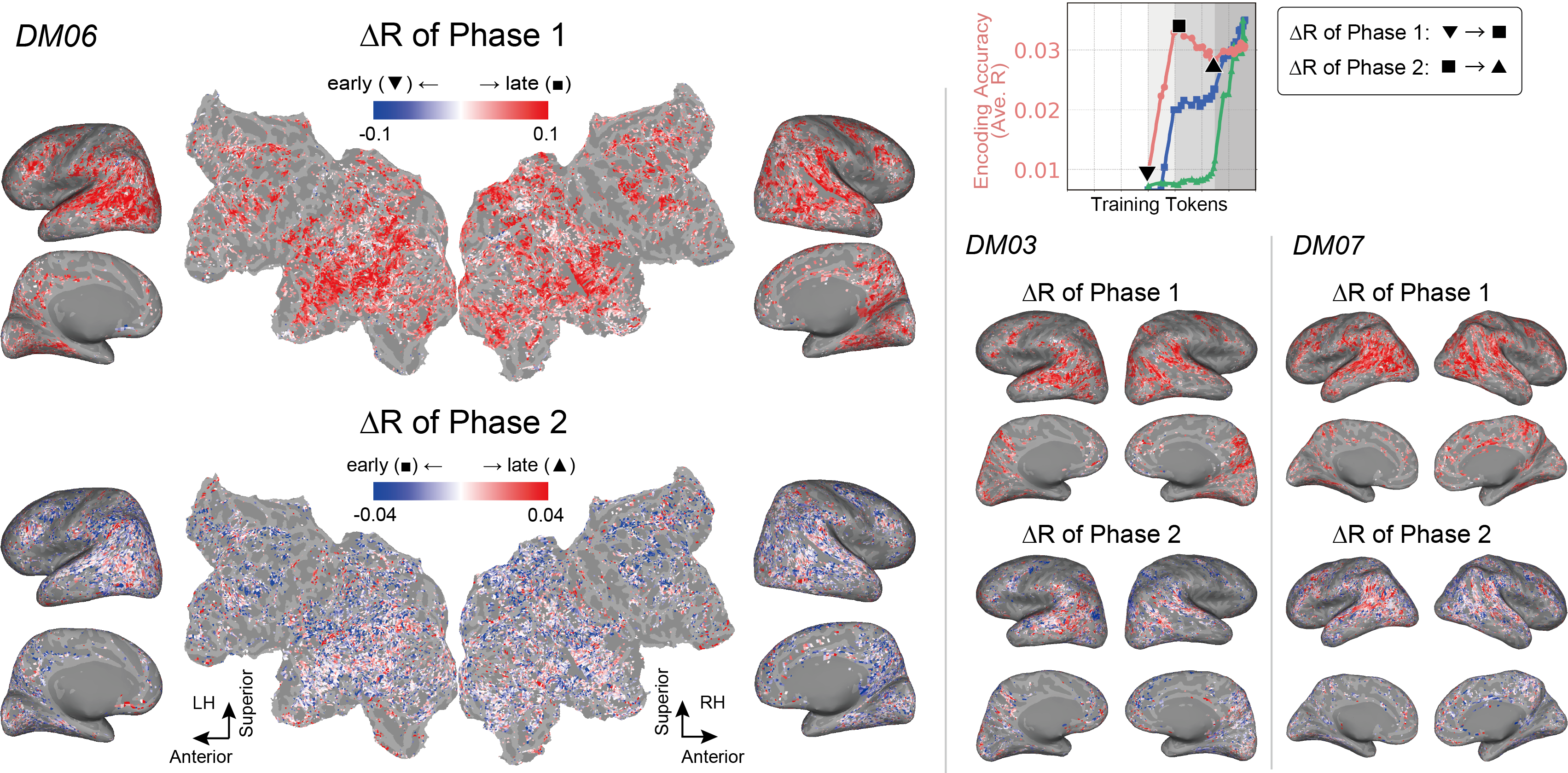}
\caption{\textbf{Changes in the relationship with the brain.} Differences in encoding accuracy among checkpoints for three participants (DM06, DM03, and DM07), projected onto the inflated (top, lateral, and medial views) and flattened cortical surface (occipital areas are at the center, only for DM06), for both the left and right hemispheres. We have chosen OLMo-2's checkpoints that capture the transitions at each phase of the learning dynamics. Brain regions with significant accuracy at either of the two checkpoints are colored ($p<0.05$, FDR-corrected). Voxels exhibiting higher accuracy at the later checkpoint are indicated in red, whereas those exhibiting higher accuracy at the earlier checkpoint are indicated in blue.}
\label{fig:result-flatmap}
\end{figure*}

\subsection{What happens during each phase?}
Section~\ref{sec:results-dynamics} characterized the three phase transitions that arise during LLM training, along with interpretive insights drawn from encoding, probing, and benchmark analyses. Nevertheless, the exact internal changes that occur during each training phase remain unclear. To gain a deeper understanding and provide further interpretation, we examine how the relationship with the brain changes at the voxel level and how the internal representations in the LLMs shift at the neuron level in each phase.

\paragraph{Changes in the relationship with the brain}
\label{sec:results-encoding}

To characterize the voxel-level relationship between LLMs and the brain across the three phases, we calculate the difference in brain encoding accuracy for each voxel between two representative training checkpoints that capture each phase transition, and project these changes onto each participant's cerebral cortex. Figure \ref{fig:result-flatmap} shows that, using the activations of OLMo-2, the voxel-level accuracy patterns shift distinctly from one phase to another. Specifically, in Phase 1—where the average prediction accuracy increases sharply—the accuracy gains are broadly distributed throughout the cerebral cortex, especially the temporal, occipital, and frontal cortex. By contrast, in Phase 2—marked by a decrease in overall brain encoding accuracy—the changes tend to vary more on a voxel-by-voxel basis, with some voxels within the temporal cortex still showing improvement. In Phase 3, the accuracy rises again across broad brain regions (see Figure~\ref{fig:subresult-flatmap-olmo-2-en}). Furthermore, comparing the post-transition prediction accuracies in Phase 3 with those in Phase 1 reveals that, in certain voxels within the temporal cortex, the post-Phase 3 accuracies are higher than the post–Phase 1 accuracies (see Figure~\ref{fig:subresult-flatmap-olmo-2-en}).

Overall, these findings indicate that once the LLMs begin following the task instructions (Phase 1), their internal representations exhibit better global alignment with brain activity. During the subsequent stagnation phase (Phase 2), this global alignment partially recedes, and after Phase 3, the LLMs realign with the brain, reinforcing the similarity to brain regions involved in semantic processing, in particular. Notably, these patterns only emerge when the input language matches one on which the model has been trained. Further details for all participants, ROIs, other phases, and LLMs can be found in Figures~\ref{fig:subresult-flatmap-olmo-2-en}--\ref{fig:subresult-flatmap-llmjp-ja}.

\begin{figure}[t]
\centering
\includegraphics[width=1.0\linewidth]{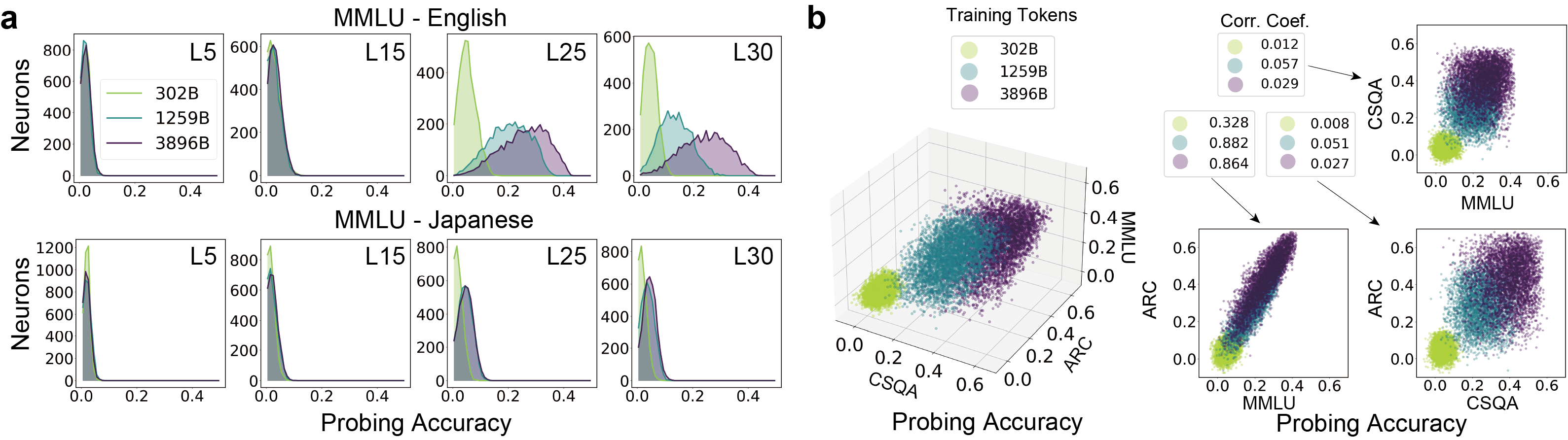}
\caption{\textbf{Probing results.} \textbf{a} Evolution of probing accuracy for English/Japanese MMLU throughout the training process, assessed at layers 5, 15, 25, and 30 of OLMo-2. The horizontal axis indicates the probing accuracy. The vertical axis indicates the number of neurons that fall within each 0.01 accuracy bin. The legends corresponds to the number of training tokens. \textbf{b} Relationship between probing accuracy in the neurons of OLMo-2 (layer 25) across English MMLU, CSQA, and ARC. Each axis denotes the probing accuracy for the respective task, and the color gradient reflects the number of training tokens. The legend shows the correlation coefficient between certain two tasks.}
\label{fig:result-probing}
\end{figure}

\paragraph{Evolution of LLM internal representations for downstream tasks}
\label{sec:results-probing}
We now investigate how the neuron-level internal representations evolve over the course of training. For this purpose, we examine changes in the distribution of probing accuracy for each neuron at the early, middle, and final checkpoints (302B, 1259B, 3896B training tokens in OLMo-2) of Phase 3. Figure~\ref{fig:result-probing}a presents the results for neurons in layers 5, 15, 25, and 30 of OLMo-2 when using MMLU. From these results, we find that, in the English MMLU setting, later-layer neurons progressively increase in probing accuracy, indicating that, as the LLMs acquire downstream task proficiency, they also develop more specialized neuron activations. Figures~\ref{fig:subresult-probing-histplot-olmo2}--\ref{fig:subresult-probing-histplot-llmjp} show the results corresponding to all downstream tasks for OLMo-2, OLMo-0724, and LLM-jp, confirming that a similar trend emerges across every downstream task when each model is provided with the language on which it has been trained.

We can also explore whether different sets of neurons acquire specialized representations for distinct downstream tasks, focusing on the same checkpoints examined in Figure~\ref{fig:result-probing}a. Figure~\ref{fig:result-probing}b illustrates the relationships among the per-neuron probing accuracies of OLMo-2 for the English MMLU, CSQA, and ARC. Across all three tasks, it is evident that certain neurons begin to manifest representations specialized for each downstream task as training progresses. Intriguingly, some neurons develop such specialized representations for all tasks, while others remain specific to only a subset of tasks. Moreover, the per-neuron accuracies for MMLU and ARC are highly correlated (r = 0.864), whereas those for CSQA demonstrate weak correlations with the other two tasks (r = 0.029 with MMLU, r = 0.027 with ARC). Furthermore, focusing on OLMo-0724—which, much like OLMo-2, demonstrates high benchmark accuracy on downstream task performance—we confirm that OLMo-0724 yields results that are analogous to those of OLMo-2 (Figure~\ref{fig:subresult-probing-scatter}). This implies that substantial transformations, specific to each task's characteristics, are emerging within these models. The results for all downstream tasks, including HellaSwag, are discussed in Section~\ref{sec:additional_results-probing}.

\subsection{Changes in the nature of activations}
\label{sec:result-llms-latent-analysis}

Based on brain encoding and probing analyses, the preceding results have demonstrated that shifts in neuronal activation occur over the course of training. We now examine whether these alterations reflect fundamental shifts in the activations of the LLMs themselves.

\begin{wrapfigure}{r}{0.5\textwidth}
 \centering
   \includegraphics[width=1\linewidth]{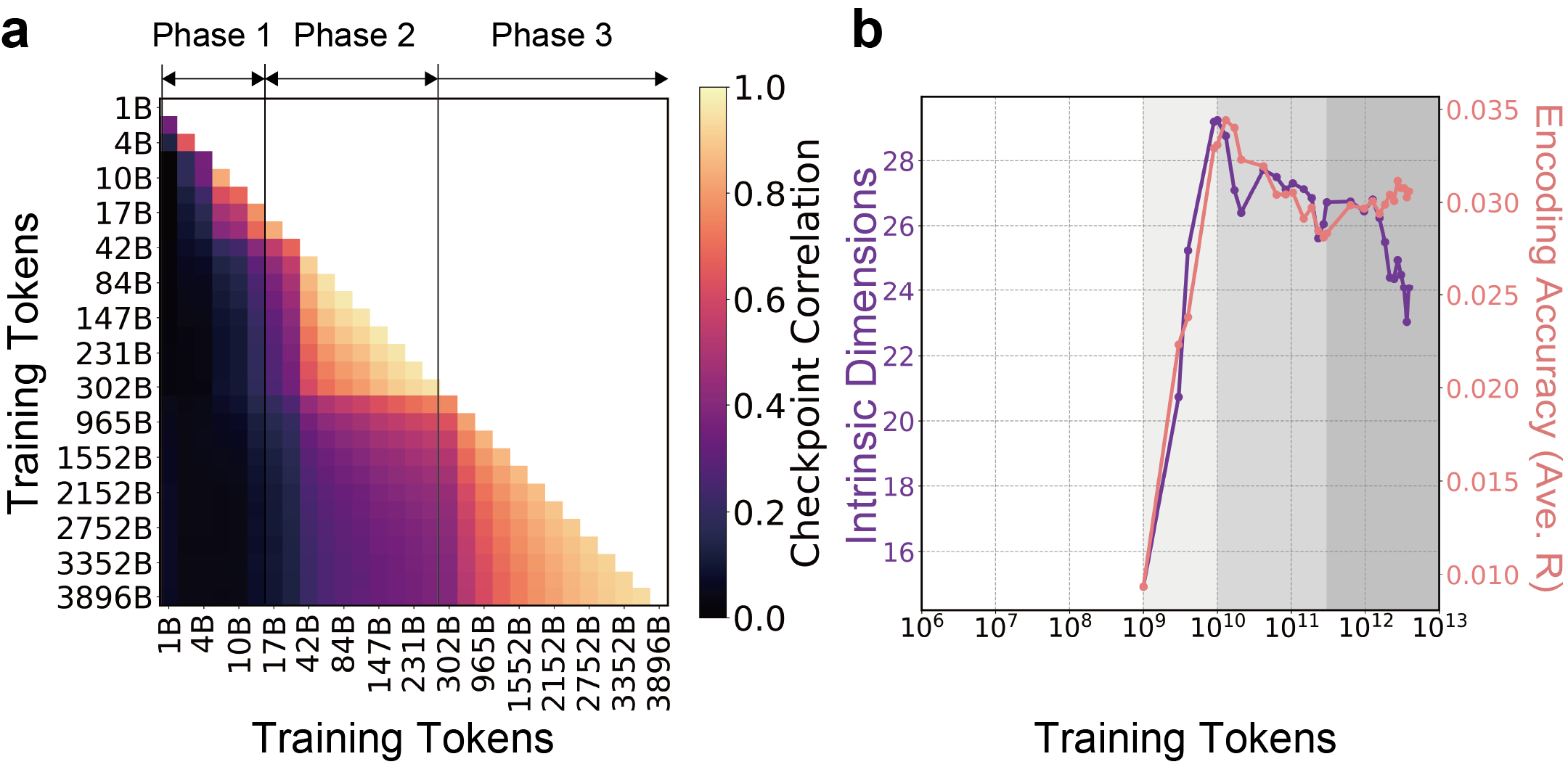}
 \caption{\textbf{The Nature of Activations.} \textbf{a} Variations in correlation coefficients of the activations of OLMo-2 across checkpoints. \textbf{b} IDs (purple line) and average encoding accuracy for all voxels of a single participant (DM06) (red line) across checkpoints.}
 \label{fig:result-ckpt-cc-id}
\end{wrapfigure}

As shown in Figure~\ref{fig:result-ckpt-cc-id}a, within-phase activations are highly similar, but substantial changes emerge at each phase transition. Interestingly, although the brain encoding accuracy does not vary dramatically between Phases 2 and 3, the underlying activations differ considerably. Next, we quantify the dimensionality of these activations using the IDs defined by GRIDE. Figure~\ref{fig:result-ckpt-cc-id}b shows that there is a strong similarity between the brain encoding accuracy and IDs during Phases 1 and 2, consistent with earlier findings that single-checkpoint encoding accuracy and IDs are highly correlated~\citep{Cheng2024-nu}. Our findings extend this result by demonstrating a similarly robust correlation across multiple checkpoints. Nevertheless, the brain encoding accuracy and IDs are not perfectly congruent (see Phase 3), leaving open questions about the aspects of activation changes that intrinsic dimensionality alone does not capture.
\section{Discussion and conclusions}
\label{sec:discussion}

In this study, we interpreted the learning dynamics of LLMs from three perspectives: alignment between LLMs and the brain, internal states associated with downstream tasks, and downstream task performance. Our novel approach elucidated a three-stage phase transition—including the precise timing of each transition—that emerges consistently during training, even among LLMs that diverge substantially in both training data and architecture.

The \textit{Brain Alignment and Instruction Following} Phase and \textit{Brain Realignment and Consolidation} Phase resemble the emergent phenomena reported by~\citet{Wei2022-gq,  wei2022chain, caballero2023broken, Olsson2022-of}, whereby models abruptly gain novel abilities after passing a threshold. Prior work has largely focused on changes in outputs or the emergence of specific mechanisms. By integrating human brain alignment and shifts in LLM internal representations, we have shown that these factors progress in unison, highlighting multi-scale indicators behind abrupt gains and stagnations in model performance. Furthermore, the marked emergence of these phenomena in languages extensively represented within the model's training corpus implies that deep internalization of their statistical patterns can trigger significant shifts in both brain alignment and overall performance. Hence, the presence or absence of these phase transitions may indicate whether an LLM has adequately assimilated and comprehended a language's statistical structure. 

Additionally, the learning dynamics that we have revealed are reminiscent of the ``synaptic overproduction and pruning'' posited in neuroscience, where an initial synaptic overgrowth eventually refines the neural circuits~\citep{PETERR1979195, 10.1093/cercor/4.1.78}. These transient changes in the brain suggest the potential for more efficient neural information processing, leading to enhanced cognitive function. Thus, a promising avenue for future work involves examining whether the phenomenon observed in this study arises from the LLM pruning superfluous internal representations in pursuit of more refined ones.

This study has also revealed that voxel-level brain alignment does not simply increase over the training process, but fluctuates across distinct phases. Previous research linking model representations to brain activity \citep{doi:10.1073/pnas.1403112111, G10005, Schrimpf2021-vr, Caucheteux2023-az} has not fully addressed such learning dynamics. While \citet{Antonello2023-pi} reported a rise in alignment with increasing model size, our findings depict a more intricate path: an initial surge in brain alignment, a decline, and a final resurgence. This suggests that LLMs may adopt distinct computational strategies at various stages, rather than gradually acquiring brain-like language representations. Moreover, alignment with semantic regions in the temporal cortex generally rises over the course of training, implying a dynamic reorganization in how models correspond to brain activity.

By capturing shifts in LLM neuron activations related to downstream tasks, we elucidated how task-specific neurons form and how much specialization they share among tasks. These findings align with previous studies showing neurons that encode particular linguistic information~\citep{tenney2018what, dai-etal-2022-knowledge, wang-etal-2022-finding-skill, gurnee2023finding}. We observed that neurons contributing substantially to certain tasks emerge abruptly in later layers as training progresses. Moreover, the variations in neuron sharing across tasks suggest that differences in activation patterns—reflecting task-specific characteristics (e.g., required capabilities, difficulty levels, and answer formats)—are indeed discernible at the neuronal level. Taken together, our results offer a more comprehensive view of LLM neurons during the learning process, building on past work that focused solely on trained models. Further in-depth analyses of these neurons would be a fascinating avenue for future research.

Finally, we extended the findings of~\citet{Cheng2024-nu}, who demonstrated a positive correlation between the dimensionality of LLM activations in trained models and the alignment of those LLMs with the brain. Our results confirm that this correlation remains robust during training, suggesting that the observed phase transitions are tied to changes in activation dimensionality. It is crucial to note that although these metrics confirm that fundamental transformations occur in the representation space, they do not fully explain the specific underlying mechanisms. By examining how these transformations relate to both the brain and downstream tasks, as in this study, a more comprehensive and nuanced understanding of the underlying processes can be achieved.

In summary, this study has elucidated the existence of three distinct phase transitions during LLM training, as evidenced by three perspectives: alignment with the human brain, shifts in internal representations pertinent to downstream tasks, and task performance. We interpreted these transitions as indispensable internal transformations that enable the model to acquire downstream task capabilities. This study is the first to show that changes in brain alignment, internal representation, and model performance advance in tandem by leveraging the human brain—the only known system (aside from LLMs) capable of processing complex language. These findings highlight the critical importance of examining multiple signals to gain a comprehensive understanding of LLM learning dynamics, including emergent phenomena. Furthermore, they suggest the potential to harness human brain activity in the pursuit of explainable and human-aligned language models.
\section*{Acknowledgements}
\label{sec:acknowledgements}
Y.T. was supported by PRESTO Grant Number JP-MJPR23I6. S.N. was supported by KAKENHI JP24H00619 and JST JPMJCR24U2.

\section*{Author contributions}
\label{sec:contributions}
All authors conceived the study. Y.N., K.T., S.Y., and Y.T. analyzed the data. Y.N. wrote the original draft. Y.N. and Y.T. wrote the manuscript with the consultation by the other authors. All authors reviewed the manuscript.

\bibliography{colm2025_conference}

\begin{thebibliography}{64}
\providecommand{\natexlab}[1]{#1}
\providecommand{\url}[1]{\texttt{#1}}
\expandafter\ifx\csname urlstyle\endcsname\relax
  \providecommand{\doi}[1]{doi: #1}\else
  \providecommand{\doi}{doi: \begingroup \urlstyle{rm}\Url}\fi

\bibitem[Ainslie et~al.(2023)Ainslie, Lee-Thorp, de~Jong, Zemlyanskiy, Lebrón, and Sanghai]{ainslie2023gqatraininggeneralizedmultiquery}
Joshua Ainslie, James Lee-Thorp, Michiel de~Jong, Yury Zemlyanskiy, Federico Lebrón, and Sumit Sanghai.
\newblock Gqa: Training generalized multi-query transformer models from multi-head checkpoints, 2023.
\newblock URL \url{https://arxiv.org/abs/2305.13245}.

\bibitem[Antonello \& Huth(2024)Antonello and Huth]{10.1162/nol_a_00087}
Richard Antonello and Alexander Huth.
\newblock Predictive coding or just feature discovery? an alternative account of why language models fit brain data.
\newblock \emph{Neurobiology of Language}, 5\penalty0 (1):\penalty0 64--79, 04 2024.
\newblock ISSN 2641-4368.
\newblock \doi{10.1162/nol_a_00087}.
\newblock URL \url{https://doi.org/10.1162/nol\_a\_00087}.

\bibitem[Antonello et~al.(2023)Antonello, Vaidya, and Huth]{Antonello2023-pi}
Richard Antonello, Aditya Vaidya, and Alexander~G Huth.
\newblock Scaling laws for language encoding models in {fMRI}.
\newblock \emph{arXiv [cs.CL]}, 19~May 2023.

\bibitem[Aw \& Toneva(2023)Aw and Toneva]{aw2023training}
Khai~Loong Aw and Mariya Toneva.
\newblock Training language models to summarize narratives improves brain alignment.
\newblock In \emph{The Eleventh International Conference on Learning Representations}, 2023.
\newblock URL \url{https://openreview.net/forum?id=KzkLAE49H9b}.

\bibitem[Aw et~al.(2023)Aw, Montariol, AlKhamissi, Schrimpf, and Bosselut]{Aw2023-zc}
Khai~Loong Aw, Syrielle Montariol, Badr AlKhamissi, Martin Schrimpf, and Antoine Bosselut.
\newblock Instruction-tuning aligns {LLMs} to the human brain.
\newblock \emph{arXiv [cs.CL]}, 1~December 2023.

\bibitem[Bereska \& Gavves(2024)Bereska and Gavves]{bereska2024mechanisticinterpretabilityaisafety}
Leonard Bereska and Efstratios Gavves.
\newblock Mechanistic interpretability for ai safety -- a review, 2024.
\newblock URL \url{https://arxiv.org/abs/2404.14082}.

\bibitem[Bourgeois et~al.(1994)Bourgeois, Goldman-Rakic, and Rakic]{10.1093/cercor/4.1.78}
Jean-Pierre Bourgeois, Patricia~S. Goldman-Rakic, and Pasko Rakic.
\newblock Synaptogenesis in the prefrontal cortex of rhesus monkeys.
\newblock \emph{Cerebral Cortex}, 4\penalty0 (1):\penalty0 78--96, 01 1994.
\newblock ISSN 1047-3211.
\newblock \doi{10.1093/cercor/4.1.78}.
\newblock URL \url{https://doi.org/10.1093/cercor/4.1.78}.

\bibitem[Caballero et~al.(2023)Caballero, Gupta, Rish, and Krueger]{caballero2023broken}
Ethan Caballero, Kshitij Gupta, Irina Rish, and David Krueger.
\newblock Broken neural scaling laws.
\newblock In \emph{The Eleventh International Conference on Learning Representations}, 2023.
\newblock URL \url{https://openreview.net/forum?id=sckjveqlCZ}.

\bibitem[Caucheteux \& King(2022)Caucheteux and King]{Caucheteux2022}
Charlotte Caucheteux and Jean-R{\'e}mi King.
\newblock Brains and algorithms partially converge in natural language processing.
\newblock \emph{Communications Biology}, 5\penalty0 (1):\penalty0 134, Feb 2022.
\newblock ISSN 2399-3642.
\newblock \doi{10.1038/s42003-022-03036-1}.
\newblock URL \url{https://doi.org/10.1038/s42003-022-03036-1}.

\bibitem[Caucheteux et~al.(2023)Caucheteux, Gramfort, and King]{Caucheteux2023-az}
Charlotte Caucheteux, Alexandre Gramfort, and Jean-Rémi King.
\newblock Evidence of a predictive coding hierarchy in the human brain listening to speech.
\newblock \emph{Nat Hum Behav}, 7\penalty0 (3):\penalty0 430--441, March 2023.

\bibitem[Chen et~al.(2023)Chen, Shwartz-Ziv, Cho, Leavitt, and Saphra]{Chen2023-wl}
Angelica Chen, Ravid Shwartz-Ziv, Kyunghyun Cho, Matthew~L Leavitt, and Naomi Saphra.
\newblock Sudden drops in the loss: Syntax acquisition, phase transitions, and simplicity bias in {MLMs}.
\newblock \emph{arXiv [cs.CL]}, 13~September 2023.

\bibitem[Cheng \& Antonello(2024)Cheng and Antonello]{Cheng2024-nu}
Emily Cheng and Richard~J Antonello.
\newblock Evidence from {fMRI} supports a two-phase abstraction process in language models.
\newblock \emph{arXiv [cs.CL]}, 9~September 2024.

\bibitem[Clark et~al.(2018)Clark, Cowhey, Etzioni, Khot, Sabharwal, Schoenick, and Tafjord]{allenai:arc}
Peter Clark, Isaac Cowhey, Oren Etzioni, Tushar Khot, Ashish Sabharwal, Carissa Schoenick, and Oyvind Tafjord.
\newblock Think you have solved question answering? try arc, the ai2 reasoning challenge.
\newblock \emph{arXiv:1803.05457v1}, 2018.

\bibitem[Dai et~al.(2022)Dai, Dong, Hao, Sui, Chang, and Wei]{dai-etal-2022-knowledge}
Damai Dai, Li~Dong, Yaru Hao, Zhifang Sui, Baobao Chang, and Furu Wei.
\newblock Knowledge neurons in pretrained transformers.
\newblock In Smaranda Muresan, Preslav Nakov, and Aline Villavicencio (eds.), \emph{Proceedings of the 60th Annual Meeting of the Association for Computational Linguistics (Volume 1: Long Papers)}, pp.\  8493--8502, Dublin, Ireland, May 2022. Association for Computational Linguistics.
\newblock \doi{10.18653/v1/2022.acl-long.581}.
\newblock URL \url{https://aclanthology.org/2022.acl-long.581/}.

\bibitem[de~Varda et~al.(2025)de~Varda, Malik-Moraleda, Tuckute, and Fedorenko]{de-Varda2025.02.01.636044}
Andrea~Gregor de~Varda, Saima Malik-Moraleda, Greta Tuckute, and Evelina Fedorenko.
\newblock Multilingual computational models reveal shared brain responses to 21 languages.
\newblock \emph{bioRxiv}, 2025.
\newblock \doi{10.1101/2025.02.01.636044}.
\newblock URL \url{https://www.biorxiv.org/content/early/2025/02/01/2025.02.01.636044}.

\bibitem[Denti et~al.(2022)Denti, Doimo, Laio, and Mira]{Denti2022}
Francesco Denti, Diego Doimo, Alessandro Laio, and Antonietta Mira.
\newblock The generalized ratios intrinsic dimension estimator.
\newblock \emph{Scientific Reports}, 12\penalty0 (1):\penalty0 20005, Nov 2022.
\newblock ISSN 2045-2322.
\newblock \doi{10.1038/s41598-022-20991-1}.
\newblock URL \url{https://doi.org/10.1038/s41598-022-20991-1}.

\bibitem[Du et~al.(2024)Du, Zeng, Dong, and Tang]{du2024understanding}
Zhengxiao Du, Aohan Zeng, Yuxiao Dong, and Jie Tang.
\newblock Understanding emergent abilities of language models from the loss perspective.
\newblock In \emph{The Thirty-eighth Annual Conference on Neural Information Processing Systems}, 2024.
\newblock URL \url{https://openreview.net/forum?id=35DAviqMFo}.

\bibitem[{Dupré la Tour} et~al.(2022){Dupré la Tour}, Eickenberg, Nunez-Elizalde, and Gallant]{DUPRELATOUR2022119728}
Tom {Dupré la Tour}, Michael Eickenberg, Anwar~O. Nunez-Elizalde, and Jack~L. Gallant.
\newblock Feature-space selection with banded ridge regression.
\newblock \emph{NeuroImage}, 264:\penalty0 119728, 2022.
\newblock ISSN 1053-8119.
\newblock \doi{https://doi.org/10.1016/j.neuroimage.2022.119728}.
\newblock URL \url{https://www.sciencedirect.com/science/article/pii/S1053811922008497}.

\bibitem[Elhage et~al.(2021)Elhage, Nanda, Olsson, Henighan, Joseph, Mann, Askell, Bai, Chen, Conerly, DasSarma, Drain, Ganguli, Hatfield-Dodds, Hernandez, Jones, Kernion, Lovitt, Ndousse, Amodei, Brown, Clark, Kaplan, McCandlish, and Olah]{elhage2021mathematical}
Nelson Elhage, Neel Nanda, Catherine Olsson, Tom Henighan, Nicholas Joseph, Ben Mann, Amanda Askell, Yuntao Bai, Anna Chen, Tom Conerly, Nova DasSarma, Dawn Drain, Deep Ganguli, Zac Hatfield-Dodds, Danny Hernandez, Andy Jones, Jackson Kernion, Liane Lovitt, Kamal Ndousse, Dario Amodei, Tom Brown, Jack Clark, Jared Kaplan, Sam McCandlish, and Chris Olah.
\newblock A mathematical framework for transformer circuits.
\newblock \emph{Transformer Circuits Thread}, 2021.
\newblock https://transformer-circuits.pub/2021/framework/index.html.

\bibitem[Facco et~al.(2017)Facco, d'Errico, Rodriguez, and Laio]{Facco2017-hk}
Elena Facco, Maria d'Errico, Alex Rodriguez, and Alessandro Laio.
\newblock Estimating the intrinsic dimension of datasets by a minimal neighborhood information.
\newblock \emph{Scientific Reports}, 7\penalty0 (1):\penalty0 12140, September 2017.

\bibitem[Ganguli et~al.(2022)Ganguli, Hernandez, Lovitt, Askell, Bai, Chen, Conerly, Dassarma, Drain, Elhage, El~Showk, Fort, Hatfield-Dodds, Henighan, Johnston, Jones, Joseph, Kernian, Kravec, Mann, Nanda, Ndousse, Olsson, Amodei, Brown, Kaplan, McCandlish, Olah, Amodei, and Clark]{10.1145/3531146.3533229}
Deep Ganguli, Danny Hernandez, Liane Lovitt, Amanda Askell, Yuntao Bai, Anna Chen, Tom Conerly, Nova Dassarma, Dawn Drain, Nelson Elhage, Sheer El~Showk, Stanislav Fort, Zac Hatfield-Dodds, Tom Henighan, Scott Johnston, Andy Jones, Nicholas Joseph, Jackson Kernian, Shauna Kravec, Ben Mann, Neel Nanda, Kamal Ndousse, Catherine Olsson, Daniela Amodei, Tom Brown, Jared Kaplan, Sam McCandlish, Christopher Olah, Dario Amodei, and Jack Clark.
\newblock Predictability and surprise in large generative models.
\newblock In \emph{Proceedings of the 2022 ACM Conference on Fairness, Accountability, and Transparency}, FAccT '22, pp.\  1747–1764, New York, NY, USA, 2022. Association for Computing Machinery.
\newblock ISBN 9781450393522.
\newblock \doi{10.1145/3531146.3533229}.
\newblock URL \url{https://doi.org/10.1145/3531146.3533229}.

\bibitem[Geva et~al.(2021)Geva, Schuster, Berant, and Levy]{geva-etal-2021-transformer}
Mor Geva, Roei Schuster, Jonathan Berant, and Omer Levy.
\newblock Transformer feed-forward layers are key-value memories.
\newblock In Marie-Francine Moens, Xuanjing Huang, Lucia Specia, and Scott Wen-tau Yih (eds.), \emph{Proceedings of the 2021 Conference on Empirical Methods in Natural Language Processing}, pp.\  5484--5495, Online and Punta Cana, Dominican Republic, November 2021. Association for Computational Linguistics.
\newblock \doi{10.18653/v1/2021.emnlp-main.446}.
\newblock URL \url{https://aclanthology.org/2021.emnlp-main.446/}.

\bibitem[Goldstein et~al.(2022)Goldstein, Zada, Buchnik, Schain, Price, Aubrey, Nastase, Feder, Emanuel, Cohen, Jansen, Gazula, Choe, Rao, Kim, Casto, Fanda, Doyle, Friedman, Dugan, Melloni, Reichart, Devore, Flinker, Hasenfratz, Levy, Hassidim, Brenner, Matias, Norman, Devinsky, and Hasson]{Goldstein2022-yw}
Ariel Goldstein, Zaid Zada, Eliav Buchnik, Mariano Schain, Amy Price, Bobbi Aubrey, Samuel~A Nastase, Amir Feder, Dotan Emanuel, Alon Cohen, Aren Jansen, Harshvardhan Gazula, Gina Choe, Aditi Rao, Catherine Kim, Colton Casto, Lora Fanda, Werner Doyle, Daniel Friedman, Patricia Dugan, Lucia Melloni, Roi Reichart, Sasha Devore, Adeen Flinker, Liat Hasenfratz, Omer Levy, Avinatan Hassidim, Michael Brenner, Yossi Matias, Kenneth~A Norman, Orrin Devinsky, and Uri Hasson.
\newblock Shared computational principles for language processing in humans and deep language models.
\newblock \emph{Nat. Neurosci.}, 25\penalty0 (3):\penalty0 369--380, March 2022.

\bibitem[Groeneveld et~al.(2024)Groeneveld, Beltagy, Walsh, Bhagia, Kinney, Tafjord, Jha, Ivison, Magnusson, Wang, Arora, Atkinson, Authur, Chandu, Cohan, Dumas, Elazar, Gu, Hessel, Khot, Merrill, Morrison, Muennighoff, Naik, Nam, Peters, Pyatkin, Ravichander, Schwenk, Shah, Smith, Subramani, Wortsman, Dasigi, Lambert, Richardson, Dodge, Lo, Soldaini, Smith, and Hajishirzi]{Groeneveld2023OLMo}
Dirk Groeneveld, Iz~Beltagy, Pete Walsh, Akshita Bhagia, Rodney Kinney, Oyvind Tafjord, Ananya~Harsh Jha, Hamish Ivison, Ian Magnusson, Yizhong Wang, Shane Arora, David Atkinson, Russell Authur, Khyathi Chandu, Arman Cohan, Jennifer Dumas, Yanai Elazar, Yuling Gu, Jack Hessel, Tushar Khot, William Merrill, Jacob Morrison, Niklas Muennighoff, Aakanksha Naik, Crystal Nam, Matthew~E. Peters, Valentina Pyatkin, Abhilasha Ravichander, Dustin Schwenk, Saurabh Shah, Will Smith, Nishant Subramani, Mitchell Wortsman, Pradeep Dasigi, Nathan Lambert, Kyle Richardson, Jesse Dodge, Kyle Lo, Luca Soldaini, Noah~A. Smith, and Hannaneh Hajishirzi.
\newblock Olmo: Accelerating the science of language models.
\newblock \emph{Preprint}, 2024.

\bibitem[G{\"u}{\c c}l{\"u} \& van Gerven(2015)G{\"u}{\c c}l{\"u} and van Gerven]{G10005}
Umut G{\"u}{\c c}l{\"u} and Marcel A.~J. van Gerven.
\newblock Deep neural networks reveal a gradient in the complexity of neural representations across the ventral stream.
\newblock \emph{Journal of Neuroscience}, 35\penalty0 (27):\penalty0 10005--10014, 2015.
\newblock ISSN 0270-6474.
\newblock \doi{10.1523/JNEUROSCI.5023-14.2015}.
\newblock URL \url{https://www.jneurosci.org/content/35/27/10005}.

\bibitem[Gurnee et~al.(2023)Gurnee, Nanda, Pauly, Harvey, Troitskii, and Bertsimas]{gurnee2023finding}
Wes Gurnee, Neel Nanda, Matthew Pauly, Katherine Harvey, Dmitrii Troitskii, and Dimitris Bertsimas.
\newblock Finding neurons in a haystack: Case studies with sparse probing.
\newblock \emph{Transactions on Machine Learning Research}, 2023.
\newblock ISSN 2835-8856.
\newblock URL \url{https://openreview.net/forum?id=JYs1R9IMJr}.

\bibitem[Hendrycks et~al.(2020)Hendrycks, Burns, Basart, Zou, Mazeika, Song, and Steinhardt]{Hendrycks2020-ev}
Dan Hendrycks, Collin Burns, Steven Basart, Andy Zou, Mantas Mazeika, Dawn Song, and Jacob Steinhardt.
\newblock Measuring massive multitask language understanding.
\newblock \emph{arXiv [cs.CY]}, 7~September 2020.

\bibitem[Hoffmann et~al.(2022)Hoffmann, Borgeaud, Mensch, Buchatskaya, Cai, Rutherford, Casas, Hendricks, Welbl, Clark, Hennigan, Noland, Millican, van~den Driessche, Damoc, Guy, Osindero, Simonyan, Elsen, Rae, Vinyals, and Sifre]{Hoffmann2022-ye}
Jordan Hoffmann, Sebastian Borgeaud, Arthur Mensch, Elena Buchatskaya, Trevor Cai, Eliza Rutherford, Diego de~Las Casas, Lisa~Anne Hendricks, Johannes Welbl, Aidan Clark, Tom Hennigan, Eric Noland, Katie Millican, George van~den Driessche, Bogdan Damoc, Aurelia Guy, Simon Osindero, Karen Simonyan, Erich Elsen, Jack~W Rae, Oriol Vinyals, and Laurent Sifre.
\newblock Training compute-optimal large language models.
\newblock \emph{arXiv [cs.CL]}, 29~March 2022.

\bibitem[Huth et~al.(2012)Huth, Nishimoto, Vu, and Gallant]{Huth2012-bh}
Alexander~G Huth, Shinji Nishimoto, An~T Vu, and Jack~L Gallant.
\newblock A continuous semantic space describes the representation of thousands of object and action categories across the human brain.
\newblock \emph{Neuron}, 76\penalty0 (6):\penalty0 1210--1224, 20~December 2012.

\bibitem[Jain \& Huth(2018)Jain and Huth]{Jain2018-st}
Shailee Jain and Alexander~G Huth.
\newblock Incorporating context into language encoding models for {fMRI}.
\newblock In \emph{Proceedings of the 32nd International Conference on Neural Information Processing Systems}, NIPS'18, pp.\  6629--6638, Red Hook, NY, USA, 3~December 2018. Curran Associates Inc.

\bibitem[Kaplan et~al.(2020)Kaplan, McCandlish, Henighan, Brown, Chess, Child, Gray, Radford, Wu, and Amodei]{Kaplan2020-ul}
Jared Kaplan, Sam McCandlish, Tom Henighan, Tom~B Brown, Benjamin Chess, Rewon Child, Scott Gray, Alec Radford, Jeffrey Wu, and Dario Amodei.
\newblock Scaling laws for neural language models.
\newblock \emph{arXiv [cs.LG]}, 22~January 2020.

\bibitem[Kurihara et~al.(2022)Kurihara, Kawahara, and Shibata]{Kurihara2022-qt}
Kentaro Kurihara, Daisuke Kawahara, and Tomohide Shibata.
\newblock {JGLUE}: Japanese general language understanding evaluation.
\newblock In \emph{Proceedings of the Thirteenth Language Resources and Evaluation Conference}, pp.\  2957--2966, 2022.

\bibitem[Liu et~al.(2023)Liu, Qiao, Neiswanger, Wang, Tan, Tao, Li, Wang, Sun, Pangarkar, Fan, Gu, Miller, Zhuang, He, Li, Koto, Tang, Ranjan, Shen, Ren, Iriondo, Mu, Hu, Schulze, Nakov, Baldwin, and Xing]{Liu2023-ky}
Zhengzhong Liu, Aurick Qiao, Willie Neiswanger, Hongyi Wang, Bowen Tan, Tianhua Tao, Junbo Li, Yuqi Wang, Suqi Sun, Omkar Pangarkar, Richard Fan, Yi~Gu, Victor Miller, Yonghao Zhuang, Guowei He, Haonan Li, Fajri Koto, Liping Tang, Nikhil Ranjan, Zhiqiang Shen, Xuguang Ren, Roberto Iriondo, Cun Mu, Zhiting Hu, Mark Schulze, Preslav Nakov, Tim Baldwin, and Eric~P Xing.
\newblock {LLM360}: Towards fully transparent open-source {LLMs}.
\newblock \emph{arXiv [cs.CL]}, 11~December 2023.

\bibitem[LLM-jp et~al.(2024)LLM-jp, :, Aizawa, Aramaki, Chen, Cheng, Deguchi, Enomoto, Fujii, Fukumoto, Fukushima, Han, Harada, Hashimoto, Hiraoka, Hisada, Hosokawa, Jie, Kamata, Kanazawa, Kanezashi, Kataoka, Katsumata, Kawahara, Kawano, Keyaki, Kiryu, Kiyomaru, Kodama, Kubo, Kuga, Kumon, Kurita, Kurohashi, Li, Maekawa, Matsuda, Miyao, Mizuki, Mizuki, Murawaki, Mousterou, Nakamura, Nakamura, Nakayama, Nakazato, Niitsuma, Nishitoba, Oda, Ogawa, Okamoto, Okazaki, Oseki, Ozaki, Ryu, Rzepka, Sakaguchi, Sasaki, Sekine, Suda, Sugawara, Sugiura, Sugiyama, Suzuki, Suzuki, Suzumura, Tachibana, Takagi, Takami, Takeda, Takeshita, Tanaka, Taura, Tolmachev, Ueda, Wan, Yada, Yahata, Yamamoto, Yamauchi, Yanaka, Yokota, and Yoshino]{llmjp2024llmjpcrossorganizationalprojectresearch}
LLM-jp, :, Akiko Aizawa, Eiji Aramaki, Bowen Chen, Fei Cheng, Hiroyuki Deguchi, Rintaro Enomoto, Kazuki Fujii, Kensuke Fukumoto, Takuya Fukushima, Namgi Han, Yuto Harada, Chikara Hashimoto, Tatsuya Hiraoka, Shohei Hisada, Sosuke Hosokawa, Lu~Jie, Keisuke Kamata, Teruhito Kanazawa, Hiroki Kanezashi, Hiroshi Kataoka, Satoru Katsumata, Daisuke Kawahara, Seiya Kawano, Atsushi Keyaki, Keisuke Kiryu, Hirokazu Kiyomaru, Takashi Kodama, Takahiro Kubo, Yohei Kuga, Ryoma Kumon, Shuhei Kurita, Sadao Kurohashi, Conglong Li, Taiki Maekawa, Hiroshi Matsuda, Yusuke Miyao, Kentaro Mizuki, Sakae Mizuki, Yugo Murawaki, Akim Mousterou, Ryo Nakamura, Taishi Nakamura, Kouta Nakayama, Tomoka Nakazato, Takuro Niitsuma, Jiro Nishitoba, Yusuke Oda, Hayato Ogawa, Takumi Okamoto, Naoaki Okazaki, Yohei Oseki, Shintaro Ozaki, Koki Ryu, Rafal Rzepka, Keisuke Sakaguchi, Shota Sasaki, Satoshi Sekine, Kohei Suda, Saku Sugawara, Issa Sugiura, Hiroaki Sugiyama, Hisami Suzuki, Jun Suzuki, Toyotaro Suzumura, Kensuke Tachibana, Yu~Takagi, Kyosuke
  Takami, Koichi Takeda, Masashi Takeshita, Masahiro Tanaka, Kenjiro Taura, Arseny Tolmachev, Nobuhiro Ueda, Zhen Wan, Shuntaro Yada, Sakiko Yahata, Yuya Yamamoto, Yusuke Yamauchi, Hitomi Yanaka, Rio Yokota, and Koichiro Yoshino.
\newblock Llm-jp: A cross-organizational project for the research and development of fully open japanese llms, 2024.
\newblock URL \url{https://arxiv.org/abs/2407.03963}.

\bibitem[Millet et~al.(2022)Millet, Caucheteux, Orhan, Boubenec, Gramfort, Dunbar, Pallier, and King]{millet2022toward}
Juliette Millet, Charlotte Caucheteux, Pierre Orhan, Yves Boubenec, Alexandre Gramfort, Ewan Dunbar, Christophe Pallier, and Jean-Remi King.
\newblock Toward a realistic model of speech processing in the brain with self-supervised learning.
\newblock In Alice~H. Oh, Alekh Agarwal, Danielle Belgrave, and Kyunghyun Cho (eds.), \emph{Advances in Neural Information Processing Systems}, 2022.
\newblock URL \url{https://openreview.net/forum?id=Y6A4-R_Hgsw}.

\bibitem[Moeller et~al.(2010)Moeller, Yacoub, Olman, Auerbach, Strupp, Harel, and Uğurbil]{Moeller2010}
Steen Moeller, Essa Yacoub, Cheryl~A. Olman, Edward Auerbach, John Strupp, Noam Harel, and Kâmil Uğurbil.
\newblock Multiband multislice ge-epi at 7 tesla, with 16-fold acceleration using partial parallel imaging with application to high spatial and temporal whole-brain fmri.
\newblock \emph{Magnetic Resonance in Medicine}, 63\penalty0 (5):\penalty0 1144--1153, 2010.
\newblock \doi{https://doi.org/10.1002/mrm.22361}.
\newblock URL \url{https://onlinelibrary.wiley.com/doi/abs/10.1002/mrm.22361}.

\bibitem[Nakagi et~al.(2024)Nakagi, Matsuyama, Koide-Majima, Yamaguchi, Kubo, Nishimoto, and Takagi]{nakagi-etal-2024-unveiling}
Yuko Nakagi, Takuya Matsuyama, Naoko Koide-Majima, Hiroto~Q. Yamaguchi, Rieko Kubo, Shinji Nishimoto, and Yu~Takagi.
\newblock Unveiling multi-level and multi-modal semantic representations in the human brain using large language models.
\newblock In Yaser Al-Onaizan, Mohit Bansal, and Yun-Nung Chen (eds.), \emph{Proceedings of the 2024 Conference on Empirical Methods in Natural Language Processing}, pp.\  20313--20338, Miami, Florida, USA, November 2024. Association for Computational Linguistics.
\newblock \doi{10.18653/v1/2024.emnlp-main.1133}.
\newblock URL \url{https://aclanthology.org/2024.emnlp-main.1133/}.

\bibitem[Nanda \& Bloom(2022)Nanda and Bloom]{nanda2022transformerlens}
Neel Nanda and Joseph Bloom.
\newblock Transformerlens.
\newblock \url{https://github.com/TransformerLensOrg/TransformerLens}, 2022.

\bibitem[Naselaris et~al.(2011)Naselaris, Kay, Nishimoto, and Gallant]{Naselaris2011-wm}
Thomas Naselaris, Kendrick~N Kay, Shinji Nishimoto, and Jack~L Gallant.
\newblock Encoding and decoding in {fMRI}.
\newblock \emph{Neuroimage}, 56\penalty0 (2):\penalty0 400--410, 15~May 2011.

\bibitem[Nishimoto et~al.(2011)Nishimoto, Vu, Naselaris, Benjamini, Yu, and Gallant]{Nishimoto2011-uc}
Shinji Nishimoto, An~T Vu, Thomas Naselaris, Yuval Benjamini, Bin Yu, and Jack~L Gallant.
\newblock Reconstructing visual experiences from brain activity evoked by natural movies.
\newblock \emph{Curr. Biol.}, 21\penalty0 (19):\penalty0 1641--1646, 11~October 2011.

\bibitem[Nostalgebraist(2020)]{logit-lens}
Nostalgebraist.
\newblock Interpreting gpt: The logit lens.
\newblock 2020.
\newblock URL \url{https://www.lesswrong.com/posts/AcKRB8wDpdaN6v6ru/interpreting-gpt-the-logit-lens}.

\bibitem[OLMo et~al.(2024)OLMo, Walsh, Soldaini, Groeneveld, Lo, Arora, Bhagia, Gu, Huang, Jordan, Lambert, Schwenk, Tafjord, Anderson, Atkinson, Brahman, Clark, Dasigi, Dziri, Guerquin, Ivison, Koh, Liu, Malik, Merrill, Miranda, Morrison, Murray, Nam, Pyatkin, Rangapur, Schmitz, Skjonsberg, Wadden, Wilhelm, Wilson, Zettlemoyer, Farhadi, Smith, and Hajishirzi]{olmo20242olmo2furious}
Team OLMo, Pete Walsh, Luca Soldaini, Dirk Groeneveld, Kyle Lo, Shane Arora, Akshita Bhagia, Yuling Gu, Shengyi Huang, Matt Jordan, Nathan Lambert, Dustin Schwenk, Oyvind Tafjord, Taira Anderson, David Atkinson, Faeze Brahman, Christopher Clark, Pradeep Dasigi, Nouha Dziri, Michal Guerquin, Hamish Ivison, Pang~Wei Koh, Jiacheng Liu, Saumya Malik, William Merrill, Lester James~V. Miranda, Jacob Morrison, Tyler Murray, Crystal Nam, Valentina Pyatkin, Aman Rangapur, Michael Schmitz, Sam Skjonsberg, David Wadden, Christopher Wilhelm, Michael Wilson, Luke Zettlemoyer, Ali Farhadi, Noah~A. Smith, and Hannaneh Hajishirzi.
\newblock 2 olmo 2 furious, 2024.
\newblock URL \url{https://arxiv.org/abs/2501.00656}.

\bibitem[Olsson et~al.(2022)Olsson, Elhage, Nanda, Joseph, DasSarma, Henighan, Mann, Askell, Bai, Chen, Conerly, Drain, Ganguli, Hatfield-Dodds, Hernandez, Johnston, Jones, Kernion, Lovitt, Ndousse, Amodei, Brown, Clark, Kaplan, McCandlish, and Olah]{Olsson2022-of}
Catherine Olsson, Nelson Elhage, Neel Nanda, Nicholas Joseph, Nova DasSarma, Tom Henighan, Ben Mann, Amanda Askell, Yuntao Bai, Anna Chen, Tom Conerly, Dawn Drain, Deep Ganguli, Zac Hatfield-Dodds, Danny Hernandez, Scott Johnston, Andy Jones, Jackson Kernion, Liane Lovitt, Kamal Ndousse, Dario Amodei, Tom Brown, Jack Clark, Jared Kaplan, Sam McCandlish, and Chris Olah.
\newblock In-context learning and induction heads.
\newblock \emph{arXiv [cs.LG]}, 23~September 2022.

\bibitem[Oota et~al.(2022)Oota, Gupta, and Toneva]{Oota2022-gd}
Subba~Reddy Oota, Manish Gupta, and Mariya Toneva.
\newblock Joint processing of linguistic properties in brains and language models.
\newblock \emph{arXiv [cs.CL]}, 15~December 2022.

\bibitem[OpenAI(2024)]{openai-mmmlu}
OpenAI.
\newblock openai/{MMMLU} · datasets at hugging face.
\newblock \url{https://huggingface.co/datasets/openai/MMMLU}, 2024.
\newblock Accessed: 2025-1-1.

\bibitem[Park et~al.(2024)Park, Okawa, Lee, Tanaka, and Lubana]{park2024emergencehiddencapabilitiesexploring}
Core~Francisco Park, Maya Okawa, Andrew Lee, Hidenori Tanaka, and Ekdeep~Singh Lubana.
\newblock Emergence of hidden capabilities: Exploring learning dynamics in concept space, 2024.
\newblock URL \url{https://arxiv.org/abs/2406.19370}.

\bibitem[{Peter R.}(1979)]{PETERR1979195}
Huttenlocher {Peter R.}
\newblock Synaptic density in human frontal cortex — developmental changes and effects of aging.
\newblock \emph{Brain Research}, 163\penalty0 (2):\penalty0 195--205, 1979.
\newblock ISSN 0006-8993.
\newblock \doi{https://doi.org/10.1016/0006-8993(79)90349-4}.
\newblock URL \url{https://www.sciencedirect.com/science/article/pii/0006899379903494}.

\bibitem[Schrimpf et~al.(2021)Schrimpf, Blank, Tuckute, Kauf, Hosseini, Kanwisher, Tenenbaum, and Fedorenko]{Schrimpf2021-vr}
Martin Schrimpf, Idan~Asher Blank, Greta Tuckute, Carina Kauf, Eghbal~A Hosseini, Nancy Kanwisher, Joshua~B Tenenbaum, and Evelina Fedorenko.
\newblock The neural architecture of language: Integrative modeling converges on predictive processing.
\newblock \emph{Proc. Natl. Acad. Sci. U. S. A.}, 118\penalty0 (45), 9~November 2021.

\bibitem[Shazeer(2020)]{shazeer2020gluvariantsimprovetransformer}
Noam Shazeer.
\newblock Glu variants improve transformer, 2020.
\newblock URL \url{https://arxiv.org/abs/2002.05202}.

\bibitem[Su et~al.(2023)Su, Lu, Pan, Murtadha, Wen, and Liu]{su2023roformerenhancedtransformerrotary}
Jianlin Su, Yu~Lu, Shengfeng Pan, Ahmed Murtadha, Bo~Wen, and Yunfeng Liu.
\newblock Roformer: Enhanced transformer with rotary position embedding, 2023.
\newblock URL \url{https://arxiv.org/abs/2104.09864}.

\bibitem[Talmor et~al.(2019)Talmor, Herzig, Lourie, and Berant]{Talmor2019-or}
Alon Talmor, Jonathan Herzig, Nicholas Lourie, and Jonathan Berant.
\newblock {CommonsenseQA}: A question answering challenge targeting commonsense knowledge.
\newblock In \emph{Proceedings of the 2019 Conference of the North}, pp.\  4149--4158, Stroudsburg, PA, USA, 2019. Association for Computational Linguistics.

\bibitem[Tenney et~al.(2019)Tenney, Xia, Chen, Wang, Poliak, McCoy, Kim, Durme, Bowman, Das, and Pavlick]{tenney2018what}
Ian Tenney, Patrick Xia, Berlin Chen, Alex Wang, Adam Poliak, R~Thomas McCoy, Najoung Kim, Benjamin~Van Durme, Sam Bowman, Dipanjan Das, and Ellie Pavlick.
\newblock What do you learn from context? probing for sentence structure in contextualized word representations.
\newblock In \emph{International Conference on Learning Representations}, 2019.
\newblock URL \url{https://openreview.net/forum?id=SJzSgnRcKX}.

\bibitem[Touvron et~al.(2023{\natexlab{a}})Touvron, Lavril, Izacard, Martinet, Lachaux, Lacroix, Rozière, Goyal, Hambro, Azhar, Rodriguez, Joulin, Grave, and Lample]{touvron2023llamaopenefficientfoundation}
Hugo Touvron, Thibaut Lavril, Gautier Izacard, Xavier Martinet, Marie-Anne Lachaux, Timothée Lacroix, Baptiste Rozière, Naman Goyal, Eric Hambro, Faisal Azhar, Aurelien Rodriguez, Armand Joulin, Edouard Grave, and Guillaume Lample.
\newblock Llama: Open and efficient foundation language models, 2023{\natexlab{a}}.
\newblock URL \url{https://arxiv.org/abs/2302.13971}.

\bibitem[Touvron et~al.(2023{\natexlab{b}})Touvron, Martin, Stone, Albert, Almahairi, Babaei, Bashlykov, Batra, Bhargava, Bhosale, Bikel, Blecher, Ferrer, Chen, Cucurull, Esiobu, Fernandes, Fu, Fu, Fuller, Gao, Goswami, Goyal, Hartshorn, Hosseini, Hou, Inan, Kardas, Kerkez, Khabsa, Kloumann, Korenev, Koura, Lachaux, Lavril, Lee, Liskovich, Lu, Mao, Martinet, Mihaylov, Mishra, Molybog, Nie, Poulton, Reizenstein, Rungta, Saladi, Schelten, Silva, Smith, Subramanian, Tan, Tang, Taylor, Williams, Kuan, Xu, Yan, Zarov, Zhang, Fan, Kambadur, Narang, Rodriguez, Stojnic, Edunov, and Scialom]{touvron2023llama2openfoundation}
Hugo Touvron, Louis Martin, Kevin Stone, Peter Albert, Amjad Almahairi, Yasmine Babaei, Nikolay Bashlykov, Soumya Batra, Prajjwal Bhargava, Shruti Bhosale, Dan Bikel, Lukas Blecher, Cristian~Canton Ferrer, Moya Chen, Guillem Cucurull, David Esiobu, Jude Fernandes, Jeremy Fu, Wenyin Fu, Brian Fuller, Cynthia Gao, Vedanuj Goswami, Naman Goyal, Anthony Hartshorn, Saghar Hosseini, Rui Hou, Hakan Inan, Marcin Kardas, Viktor Kerkez, Madian Khabsa, Isabel Kloumann, Artem Korenev, Punit~Singh Koura, Marie-Anne Lachaux, Thibaut Lavril, Jenya Lee, Diana Liskovich, Yinghai Lu, Yuning Mao, Xavier Martinet, Todor Mihaylov, Pushkar Mishra, Igor Molybog, Yixin Nie, Andrew Poulton, Jeremy Reizenstein, Rashi Rungta, Kalyan Saladi, Alan Schelten, Ruan Silva, Eric~Michael Smith, Ranjan Subramanian, Xiaoqing~Ellen Tan, Binh Tang, Ross Taylor, Adina Williams, Jian~Xiang Kuan, Puxin Xu, Zheng Yan, Iliyan Zarov, Yuchen Zhang, Angela Fan, Melanie Kambadur, Sharan Narang, Aurelien Rodriguez, Robert Stojnic, Sergey Edunov, and Thomas
  Scialom.
\newblock Llama 2: Open foundation and fine-tuned chat models, 2023{\natexlab{b}}.
\newblock URL \url{https://arxiv.org/abs/2307.09288}.

\bibitem[Tuckute et~al.(2024{\natexlab{a}})Tuckute, Kanwisher, and Fedorenko]{annurev:/content/journals/10.1146/annurev-neuro-120623-101142}
Greta Tuckute, Nancy Kanwisher, and Evelina Fedorenko.
\newblock Language in brains, minds, and machines.
\newblock \emph{Annual Review of Neuroscience}, 47\penalty0 (Volume 47, 2024):\penalty0 277--301, 2024{\natexlab{a}}.
\newblock ISSN 1545-4126.
\newblock \doi{https://doi.org/10.1146/annurev-neuro-120623-101142}.
\newblock URL \url{https://www.annualreviews.org/content/journals/10.1146/annurev-neuro-120623-101142}.

\bibitem[Tuckute et~al.(2024{\natexlab{b}})Tuckute, Sathe, Srikant, Taliaferro, Wang, Schrimpf, Kay, and Fedorenko]{Tuckute2024}
Greta Tuckute, Aalok Sathe, Shashank Srikant, Maya Taliaferro, Mingye Wang, Martin Schrimpf, Kendrick Kay, and Evelina Fedorenko.
\newblock Driving and suppressing the human language network using large language models.
\newblock \emph{Nature Human Behaviour}, 8\penalty0 (3):\penalty0 544--561, Mar 2024{\natexlab{b}}.
\newblock ISSN 2397-3374.
\newblock \doi{10.1038/s41562-023-01783-7}.
\newblock URL \url{https://doi.org/10.1038/s41562-023-01783-7}.

\bibitem[Vaswani et~al.(2017)Vaswani, Shazeer, Parmar, Uszkoreit, Jones, Gomez, Kaiser, and Polosukhin]{NIPS2017_3f5ee243}
Ashish Vaswani, Noam Shazeer, Niki Parmar, Jakob Uszkoreit, Llion Jones, Aidan~N Gomez, \L~ukasz Kaiser, and Illia Polosukhin.
\newblock Attention is all you need.
\newblock In I.~Guyon, U.~Von Luxburg, S.~Bengio, H.~Wallach, R.~Fergus, S.~Vishwanathan, and R.~Garnett (eds.), \emph{Advances in Neural Information Processing Systems}, volume~30. Curran Associates, Inc., 2017.
\newblock URL \url{https://proceedings.neurips.cc/paper_files/paper/2017/file/3f5ee243547dee91fbd053c1c4a845aa-Paper.pdf}.

\bibitem[Wang et~al.(2022)Wang, Wen, Zhang, Hou, Liu, and Li]{wang-etal-2022-finding-skill}
Xiaozhi Wang, Kaiyue Wen, Zhengyan Zhang, Lei Hou, Zhiyuan Liu, and Juanzi Li.
\newblock Finding skill neurons in pre-trained transformer-based language models.
\newblock In Yoav Goldberg, Zornitsa Kozareva, and Yue Zhang (eds.), \emph{Proceedings of the 2022 Conference on Empirical Methods in Natural Language Processing}, pp.\  11132--11152, Abu Dhabi, United Arab Emirates, December 2022. Association for Computational Linguistics.
\newblock \doi{10.18653/v1/2022.emnlp-main.765}.
\newblock URL \url{https://aclanthology.org/2022.emnlp-main.765/}.

\bibitem[Wei et~al.(2022{\natexlab{a}})Wei, Tay, Bommasani, Raffel, Zoph, Borgeaud, Yogatama, Bosma, Zhou, Metzler, Chi, Hashimoto, Vinyals, Liang, Dean, and Fedus]{Wei2022-gq}
Jason Wei, Yi~Tay, Rishi Bommasani, Colin Raffel, Barret Zoph, Sebastian Borgeaud, Dani Yogatama, Maarten Bosma, Denny Zhou, Donald Metzler, Ed~H. Chi, Tatsunori Hashimoto, Oriol Vinyals, Percy Liang, Jeff Dean, and William Fedus.
\newblock Emergent abilities of large language models.
\newblock \emph{Transactions on Machine Learning Research}, 2022{\natexlab{a}}.
\newblock ISSN 2835-8856.
\newblock URL \url{https://openreview.net/forum?id=yzkSU5zdwD}.
\newblock Survey Certification.

\bibitem[Wei et~al.(2022{\natexlab{b}})Wei, Wang, Schuurmans, Bosma, brian ichter, Xia, Chi, Le, and Zhou]{wei2022chain}
Jason Wei, Xuezhi Wang, Dale Schuurmans, Maarten Bosma, brian ichter, Fei Xia, Ed~H. Chi, Quoc~V Le, and Denny Zhou.
\newblock Chain of thought prompting elicits reasoning in large language models.
\newblock In Alice~H. Oh, Alekh Agarwal, Danielle Belgrave, and Kyunghyun Cho (eds.), \emph{Advances in Neural Information Processing Systems}, 2022{\natexlab{b}}.
\newblock URL \url{https://openreview.net/forum?id=_VjQlMeSB_J}.

\bibitem[Wolf et~al.(2020)Wolf, Debut, Sanh, Chaumond, Delangue, Moi, Cistac, Rault, Louf, Funtowicz, Davison, Shleifer, von Platen, Ma, Jernite, Plu, Xu, Scao, Gugger, Drame, Lhoest, and Rush]{wolf-etal-2020-transformers}
Thomas Wolf, Lysandre Debut, Victor Sanh, Julien Chaumond, Clement Delangue, Anthony Moi, Pierric Cistac, Tim Rault, Rémi Louf, Morgan Funtowicz, Joe Davison, Sam Shleifer, Patrick von Platen, Clara Ma, Yacine Jernite, Julien Plu, Canwen Xu, Teven~Le Scao, Sylvain Gugger, Mariama Drame, Quentin Lhoest, and Alexander~M. Rush.
\newblock Transformers: State-of-the-art natural language processing.
\newblock In \emph{Proceedings of the 2020 Conference on Empirical Methods in Natural Language Processing: System Demonstrations}, pp.\  38--45, Online, October 2020. Association for Computational Linguistics.
\newblock URL \url{https://www.aclweb.org/anthology/2020.emnlp-demos.6}.

\bibitem[Yamaguchi et~al.(2024)Yamaguchi, Koide-Majima, Kubo, Nakai, and Nishimoto]{Yamaguchi2024-is}
Hiroto~Q Yamaguchi, Naoko Koide-Majima, Rieko Kubo, Tomoya Nakai, and Shinji Nishimoto.
\newblock Narrative movie {fMRI} dataset, 4~October 2024.

\bibitem[Yamins et~al.(2014)Yamins, Hong, Cadieu, Solomon, Seibert, and DiCarlo]{doi:10.1073/pnas.1403112111}
Daniel L.~K. Yamins, Ha~Hong, Charles~F. Cadieu, Ethan~A. Solomon, Darren Seibert, and James~J. DiCarlo.
\newblock Performance-optimized hierarchical models predict neural responses in higher visual cortex.
\newblock \emph{Proceedings of the National Academy of Sciences}, 111\penalty0 (23):\penalty0 8619--8624, 2014.
\newblock \doi{10.1073/pnas.1403112111}.
\newblock URL \url{https://www.pnas.org/doi/abs/10.1073/pnas.1403112111}.

\bibitem[Zellers et~al.(2019)Zellers, Holtzman, Bisk, Farhadi, and Choi]{zellers2019hellaswag}
Rowan Zellers, Ari Holtzman, Yonatan Bisk, Ali Farhadi, and Yejin Choi.
\newblock Hellaswag: Can a machine really finish your sentence?
\newblock In \emph{Proceedings of the 57th Annual Meeting of the Association for Computational Linguistics}, 2019.

\end{thebibliography}
\bibliographystyle{colm2025_conference}
\appendix
\renewcommand{\thefigure}{A.\arabic{figure}}
\setcounter{figure}{0}
\section{Additional methods}
\label{appendix: additional_methods}

\subsection{Large language models}
\label{appendix:additional_methods:llms}

We used the allenai/OLMo-2-1124-7B, allenai/OLMo-7B-0724-hf, llm-jp/llm-jp-3-7.2b, and LLM360/Amber models available on Hugging Face for OLMo-2, OLMo-0724, LLM-jp, and Amber. All checkpoints of LLM-jp will be made publicly available, although it has only released its final checkpoint. Tables \ref{tab:method-llms} and \ref{tab:appendix-method-checkpoints} present an overview of the LLMs and detailed information on their respective training checkpoints used in this study. We used 28 checkpoints for OLMo-2 (1B-3,896B training tokens), 23 for OLMo-0724 (4B–2,724B), 27 for LLM-jp (4.2M–1,258B), and 18 for Amber (3.5B–1,259B). In selecting these checkpoints, we took particular care to ensure that the number of training tokens was as closely aligned as possible across the four LLMs. 

\begin{table}[h]
\centering
\tabcolsep 2.0pt
\begin{tabular}[width=0.7\linewidth]{lccccc}
\toprule
Model & Layers & Width & Params. & Vocab. sizes & Trn. Tokens (Ckpts.)\\
\midrule
OLMo-2 & 32 & 4096 & 7.3B & 100352 & 1B - 3896B (28)\\
OLMo-0724 & 32 & 4096 & 6.89B & 50304 & 4B - 2724B (23)\\
LLM-jp & 32 & 4096 & 7.29B & 99584 & 4.2M - 1258B (27)\\
Amber & 32 & 4096 & 6.74B & 32000 & 3.5B - 1259B (18)\\
\bottomrule
\end{tabular}
\caption{Overview of LLMs}
\label{tab:method-llms}
\end{table}

\begin{table*}[h]
\centering
\tabcolsep 1pt
\begin{tabular}{p{2.1cm}p{11cm}}
\toprule
Model & Checkpoints\\
\midrule \midrule

\multirow{3}{*}{OLMo-2} & 150, 600, 900, 2K, 3K, 4K, 5K, 10K, 15K, 20K, 25K, 35K, 45K, 55K, 65K, 72K, 150K, 230K, 300K, 370K, 441K, 513K, 584K, 656K, 727K, 799K, 870K, 928.646K (Training step) \\
\midrule
\multirow{3}{*}{OLMo-0724} & 1K, 2K, 2.5K, 3.5K, 4.5K, 5K, 10K, 15K, 20.5K, 25.5K, 35.5K, 45.5K, 55.5K, 65K, 72K, 149.5K, 230K, 300K, 370K, 442K, 514K, 585K, 649.65K (Training step) \\
\midrule
\multirow{2}{*}{LLM-jp} & 1, 2, 4, 8, 20, 30, 60, 100, 300, 500, 1K, 2K, 3K, 4K, 5K, 10K, 15K, 20K, 25K, 35K, 45K, 55K, 65K, 72K, 150K, 230K, 300K (Training step) \\
\midrule
Amber & 1, 2, 3, 4, 5, 6, 12, 18, 24, 30, 42, 54, 66, 78, 86, 179, 275, main (Checkpoint)\\
\bottomrule
\end{tabular}
\caption{Details of the training checkpoints}
\label{tab:appendix-method-checkpoints}
\end{table*}

The architectural variations among the LLMs used in this study encompass layer normalization, activation functions, positional embeddings, and attention mechanisms. All models are derived from a decoder-only Transformer \citep{NIPS2017_3f5ee243} architecture, albeit with several critical modifications:
\begin{enumerate}
\item LLM-jp is built upon Llama 2 \citep{touvron2023llama2openfoundation}, whereas Amber is derived from LLaMA \citep{touvron2023llamaopenefficientfoundation}.
\item In OLMo-0724, LLM-jp, and Amber, layer normalization is applied before the self-attention and MLP sublayers; in OLMo-2, layer normalization is applied after these sublayers.
\item Regarding activation normalization, OLMo-2, LLM-jp, and Amber use RMSNorm, whereas OLMo-0724 adopts a nonparametric norm.
\item In all models, the output of the self-attention mechanism is added to the residual stream preceding the MLP.
\item In all models, the ReLU activation function is replaced by the SwiGLU activation function \citep{shazeer2020gluvariantsimprovetransformer}.
\item All models substitute absolute positional embeddings with rotary positional embeddings \citep{su2023roformerenhancedtransformerrotary}.
\item To simplify the self-attention computations, LLM-jp uses grouped query attention \citep{ainslie2023gqatraininggeneralizedmultiquery}.
\item For enhanced training stability, OLMo-2 and OLMo-0724 both use QKV Clipping.
\item Finally, to prevent excessively large attention logits—and consequently prevent the training loss from diverging—OLMo-2 normalizes the Key and Query projections via RMSNorm before computing the attention.
\end{enumerate}

\subsection{fMRI datasets}
\label{appendix:additional_methods:fmri_datasets}

MRI data were acquired using a 3T MAGNETOM Vida scanner (Siemens, Germany) with a standard Siemens 64-channel volume coil. Functional brain images based on the blood oxygenation level-dependent (BOLD) signal were collected via a multiband gradient echo-planar imaging sequence \citep{Moeller2010} (TR = 1,000 ms, TE = 30 ms, flip angle = 60$^\circ$, voxel size = 2 \(\times\) 2 \(\times\) 2 mm\(^3\), matrix size = 96 \(\times\) 96, 72 slices with a thickness of 2 mm, slice gap 0 mm, FOV = 192 \(\times\) 192 mm\(^2\), bandwidth 1736 Hz/pixel, partial Fourier 6/8, multiband acceleration factor 6). Anatomical data were acquired using the same 3T scanner using T1-weighted MPRAGE (TR = 2530 ms, TE = 3.26 ms, flip angle = 9$^\circ$, voxel size = 1 \(\times\) 1 \(\times\) 1 mm\(^3\), FOV = 256 \(\times\) 256 mm\(^2\)). The preprocessing of the fMRI data included motion correction, coregistration, and detrending. All participants are right-handed, native Japanese speakers and provided written informed consent for this study, which was conducted under the approval of the relevant ethics and safety committee.

This dataset comprises nine videos of movies or drama series as experimental stimuli (ten episodes in total). The videos span a diverse range of genres: eight international movies or dramas and one Japanese animation. The average duration across the ten episodes is 49.98 min (minimum 21 min, maximum 125 min). Each episode is segmented into 2-9 parts, each lasting approximately 10 min. These segments were administered as fMRI stimuli.

This dataset provides three types of natural language annotations describing the stimulus videos: \textit{Objective Information}, \textit{Speech Transcription}, and \textit{Narrative Content (Story)}. Each type of annotation captures distinct semantic content relevant to narrative comprehension. We used the \textit{Narrative Content (Story)} annotation for the main analysis and the \textit{Objective Information} annotation for the control analysis. All annotations were originally described in Japanese. They were translated into English and back-translated into Japanese using DeepL.

\subsection{Brain encoding models}
\label{appendix:additional_methods:brain-encoding-model}

The dataset used in this study comprises nine movies or dramas, and therefore the regularization parameters were tuned during training, using sessions from two or three movies or dramas as validation data and the remaining sessions as training data. This procedure was iterated for cross-validation. For the evaluation, we computed the Pearson's correlation coefficients between the predicted and measured fMRI signals. Statistical significance was assessed using a blockwise permutation test. Specifically, to generate a null distribution, we shuffled the voxel's measured response time course before calculating the Pearson's correlation between the predicted response time course and the permuted response time course. During this process, we shuffled the measured response time course in blocks of 10 TRs to preserve the temporal correlation between slices. We identified voxels having scores significantly higher than those expected by chance in the null distribution.

On the basis of the encoding analysis results in Section \ref{sec:methods-encoding}, we performed an analysis using the regions of interest (ROIs) included in the DeusTex atlas. We selected focal ROIs that (1) exhibited a trend of three-phase transitions in encoding accuracy throughout training (though the specific timing of these shifts varied by participant), (2) contained voxels showing a pronounced manifestation of these three-phase transitions, and (3) demonstrated comparatively high encoding accuracy for every participant.

All encoding (Section \ref{sec:methods-encoding}) and probing (Section \ref{sec:methods-probing}) analyses were conducted using the \textit{himalaya} library\footnote{\url{https://github.com/gallantlab/himalaya}} \citep{DUPRELATOUR2022119728} and the \textit{drama2brain} library\footnote{\url{https://github.com/yu-takagi/drama2brain}} \citep{nakagi-etal-2024-unveiling}. To extract latent representations from the MLP layers of OLMo-2, we modified the code from the \textit{Transformers} library\footnote{\url{https://github.com/huggingface/transformers}} \citep{wolf-etal-2020-transformers}. To extract latent representations from the MLP layers of OLMo-0724, LLM-jp, and Amber, we modified the code from the \textit{TransformerLens} library\footnote{\url{https://github.com/TransformerLensOrg/TransformerLens}} \citep{nanda2022transformerlens}. We will make our source code and training data for the encoding, probing (See Section~\ref{sec:methods-probing}), and benchmark (See Section~\ref{sec:methods-benchmark}) analyses publicly available on acceptance.

\subsection{Downstream datasets}
\label{appendix:additional_methods:Downstream_datasets}
MMLU assesses broad knowledge and problem-solving abilities using multidisciplinary coverage of 57 subjects, CSQA tests everyday conceptual commonsense reasoning, ARC probes elementary-level scientific knowledge, and HellaSwag assesses contextual commonsense reasoning in typical scenarios. For MMLU, we use the original English dataset from \citet{Hendrycks2020-ev} and its Japanese translation from \citet{openai-mmmlu}. For control analysis, we use the Chinese translation from \citet{openai-mmmlu}. Each of these datasets (English/Japanese/Chinese) comprises 13,571 samples. For CSQA, we use the original English dataset from \citet{Talmor2019-or} and the Japanese dataset from \citet{Kurihara2022-qt}, which contain 10,957 and 8,934 samples, respectively. For ARC (both the ARC-Challenge and ARC-Easy subsets), we use the original English dataset from \citet{allenai:arc} and its Japanese translation, resulting in 7,778 samples for both the English and Japanese versions. For HellaSwag, we use the original English dataset from \citet{zellers2019hellaswag} and its Japanese translation, resulting in 9,658 samples for both the English and Japanese versions. We use the OpenAI API (GPT 4o-mini) for translation.

In the probing analysis, each dataset is split into training and test datasets at a 4:1 ratio. Because MMLU comprises multiple subject areas, we split the dataset by subject. Furthermore, during the optimization of regularization parameters described in Section~\ref{sec:methods-brain-inspired-Probing}, to mitigate the subject-based bias of MMLU, we shuffle the training indices, and then perform cross-validation to ensure balanced distributions in each fold.

\subsection{Intrinsic dimensions}
\label{appendix:activaions-dimensionality-analysis}
We used GRIDE~\citep{Denti2022} to compute the IDs. GRIDE extends the TwoNN estimator~\citep{Facco2017-hk} to general scales.

\paragraph{Estimation procedure using GRIDE}
GRIDE employs the following ratio as its fundamental component:
\begin{equation}
  \mu_{i,2k,k} = \frac{r_{i,2k}}{r_{i,k}}\notag
\end{equation}
where \(r_{i,j}\) denotes the Euclidean distance between point \(i\) and its \(j\)-th nearest neighbor. Under the assumption of a locally uniform density distribution, these ratios \(\mu_{i,2k,k}\) are shown to follow a generalized Pareto distribution:
\begin{equation}
  f_{\mu_{i,2k,k}}(\mu) 
  = \frac{d \left(\mu^{\,d-1}\right)^{k-1}}{B(k,k)\mu^{\,d(2k-1)+1}}\notag
\end{equation}
where \(B(\cdot,\cdot)\) is the beta function. Furthermore, assuming independence among the ratios \(\mu_{i,2k,k}\) from different points, the likelihood of this distribution can be numerically maximized to obtain the ID. In this study, for each model checkpoint, we selected the value of \(k\) at which the mean ID across layers stabilized at its maximum, thereby determining the estimated IDs.

\subsection{Determining the layers of interest}
\label{appendix:additional_methods:select-layer}
In interpreting the learning dynamics of LLMs from three distinct perspectives, we determined which layers merited attention based on (1) each layer's encoding accuracy, (2) each layer's probing accuracy, and (3) each layer's benchmark accuracy. We obtained the encoding accuracy and the probing accuracy according to the methods described in Sections \ref{sec:methods-encoding} and \ref{sec:methods-probing}, respectively. We computed each layer's benchmark accuracy using Logit Lens \citep{logit-lens}.

\paragraph{Logit lens}
In the output layer of an LLM, an unembedding matrix is employed to convert vectors into tokens by projecting the hidden-layer vectors within the model onto the vocabulary dimension. A softmax function (or similar) is then applied to compute probabilities and generate the output tokens. This process is referred to as ``unembedding''.

The hidden-layer vectors within the model have the same dimensionality as the vectors in the output layer, and therefore  the unembedding procedure can be applied to the hidden-layer vectors, thereby gaining insight into the intermediate processes. Logit Lens is a tool specifically devised for this purpose.

\paragraph{Measuring benchmark accuracy by layer}
Using Logit Lens, we extracted the probability distribution over the final token predicted from each intermediate layer, and designated the token assigned the highest probability as that layer's output. We then computed the proportion of correctly answered questions by dividing the number of correct answers by the total number of questions, analogous to the procedure described in Section \ref{sec:methods-benchmark}. This proportion was treated as the benchmark accuracy for that layer.

In Section \ref{sec:additional_results:select-layer}, we presented the layer-wise accuracy of OLMo-2, OLMo-0724, and LLM-jp with respect to the three metrics at each training checkpoint, thereby determining which layers were to be examined in greater detail.

\subsection{Examples of 5-shot prompts}
\label{appendix:additional_methods:prompts-examples}
\begin{figure}[htbp]
  \centering
  \begin{subfigure}[b]{0.48\linewidth}
    \centering
    \includegraphics[width=\linewidth]{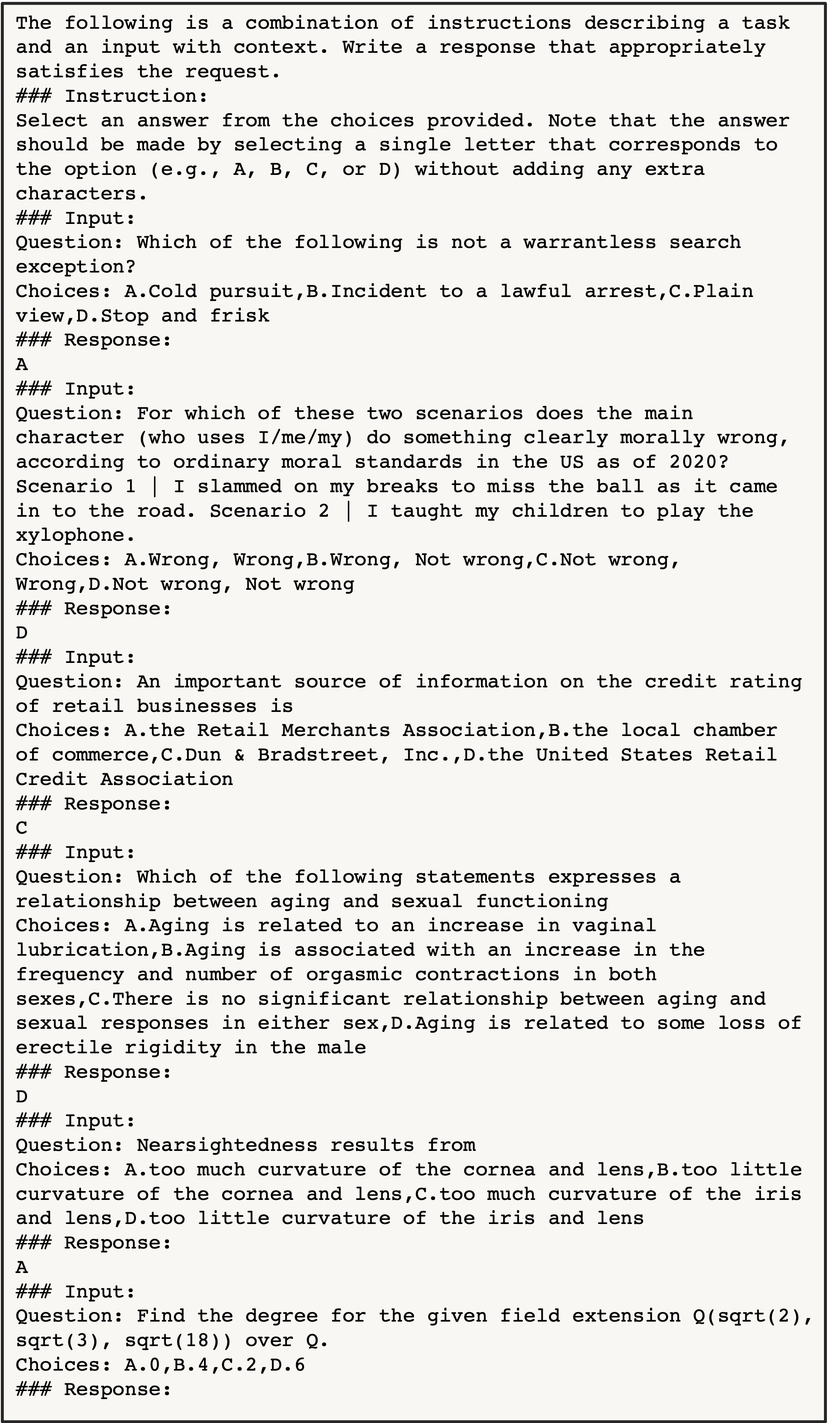}
  \end{subfigure}
  \hfill
  \begin{subfigure}[b]{0.48\linewidth}
    \centering
    \includegraphics[width=\linewidth]{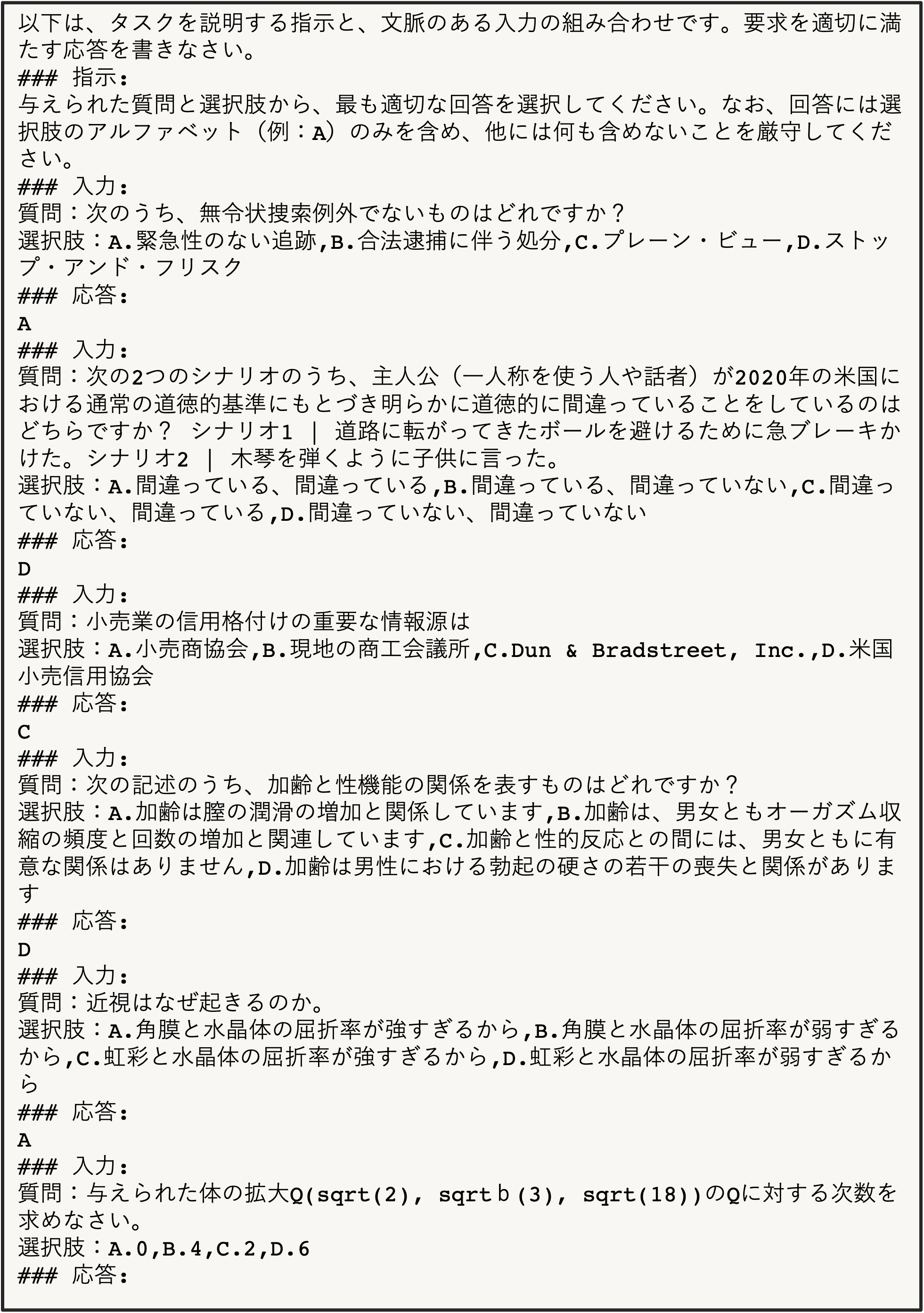}
  \end{subfigure}
  \caption{Example of MMLU (English and Japanese) prompt.}
  \label{fig:prompt-example-mmlu}
\end{figure}

\begin{figure}[htbp]
  \centering
  \begin{subfigure}{0.48\linewidth}
    \centering
    \includegraphics[width=\linewidth]{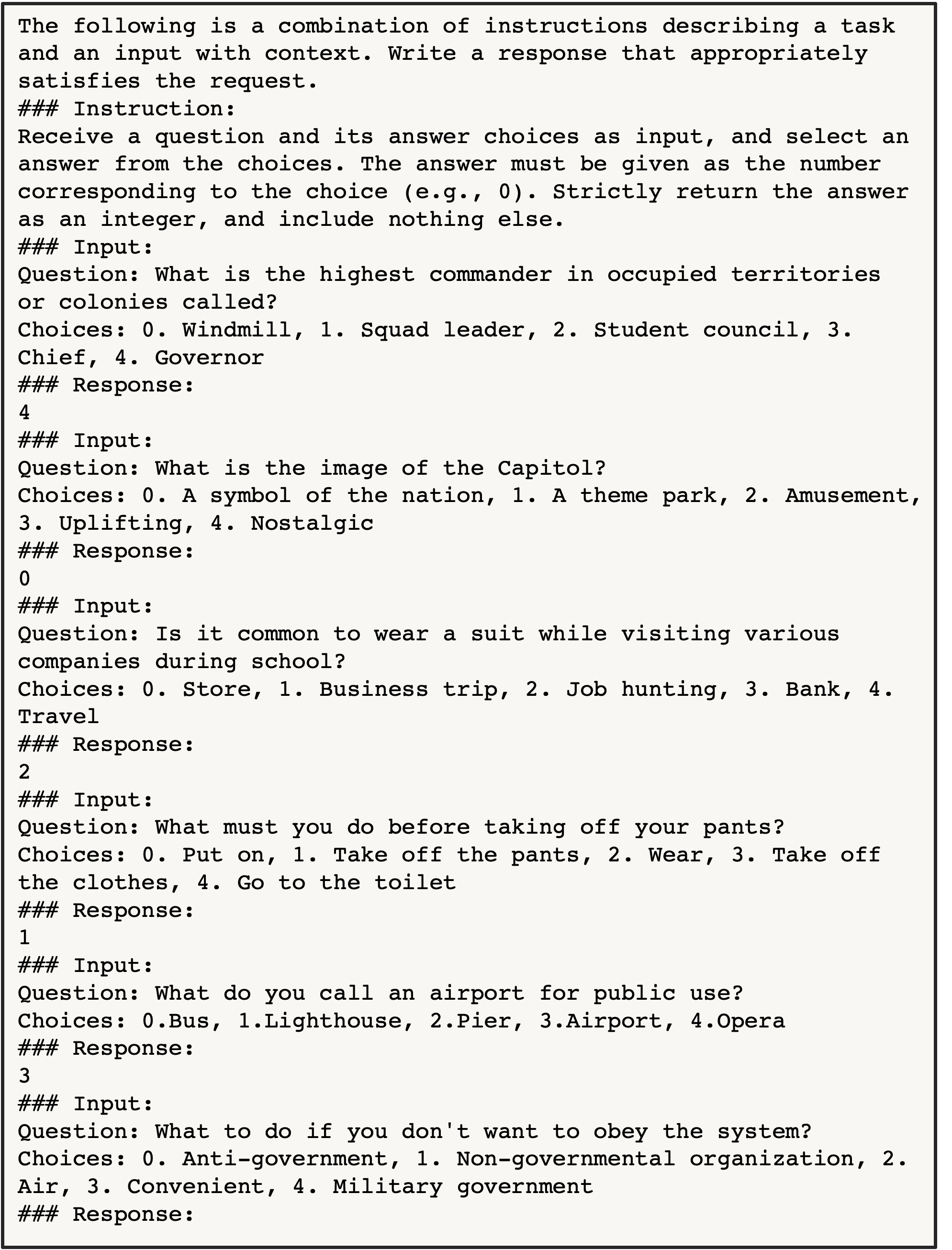}
  \end{subfigure}
  \hfill
  \begin{subfigure}{0.48\linewidth}
    \centering
    \includegraphics[width=\linewidth]{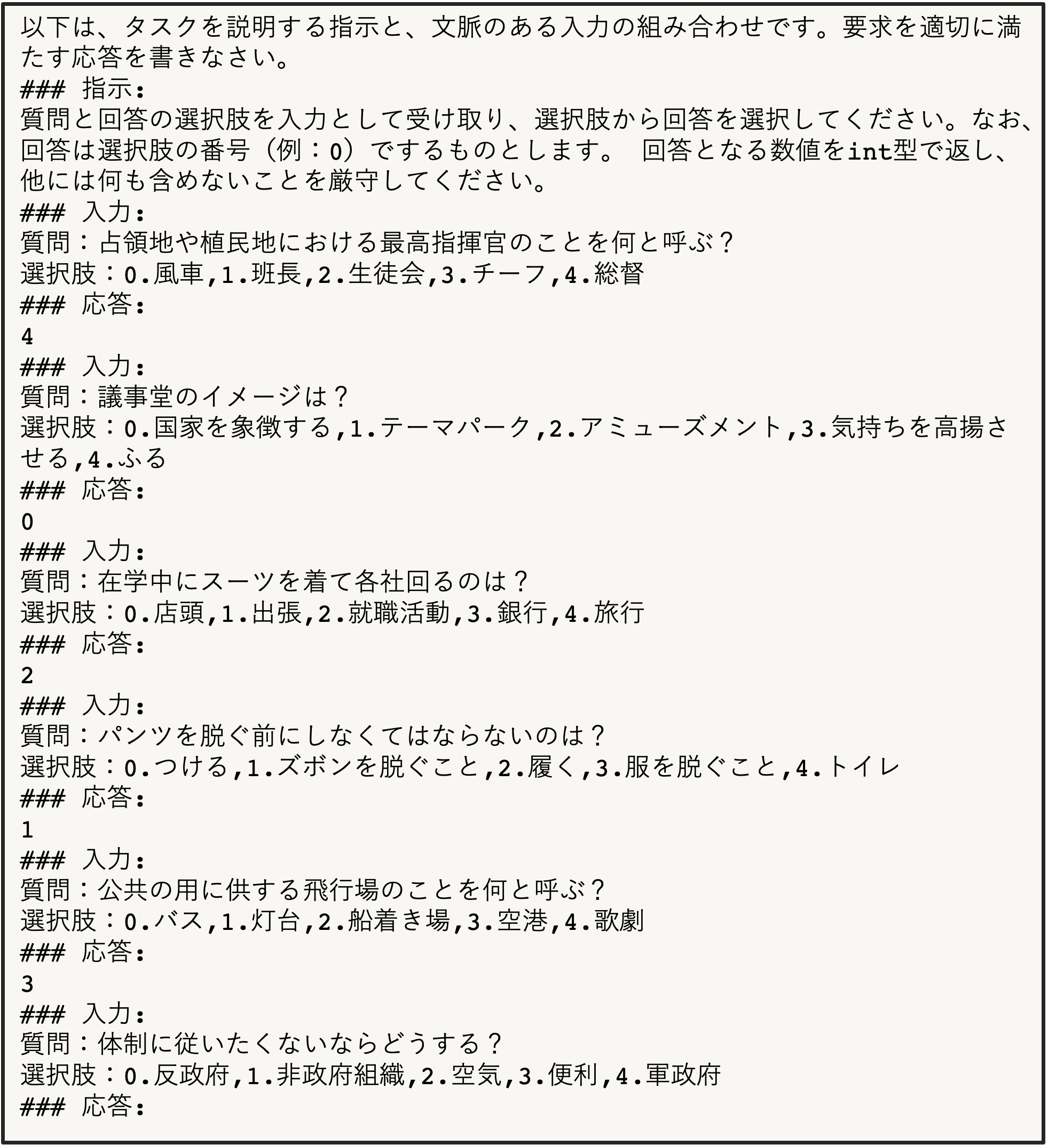}
  \end{subfigure}
  \caption{Example of CSQA (English and Japanese) prompt.}
  \label{fig:prompt-example-csqa}
\end{figure}

\begin{figure}[htbp]
  \centering
  \begin{subfigure}{0.48\linewidth}
    \centering
    \includegraphics[width=\linewidth]{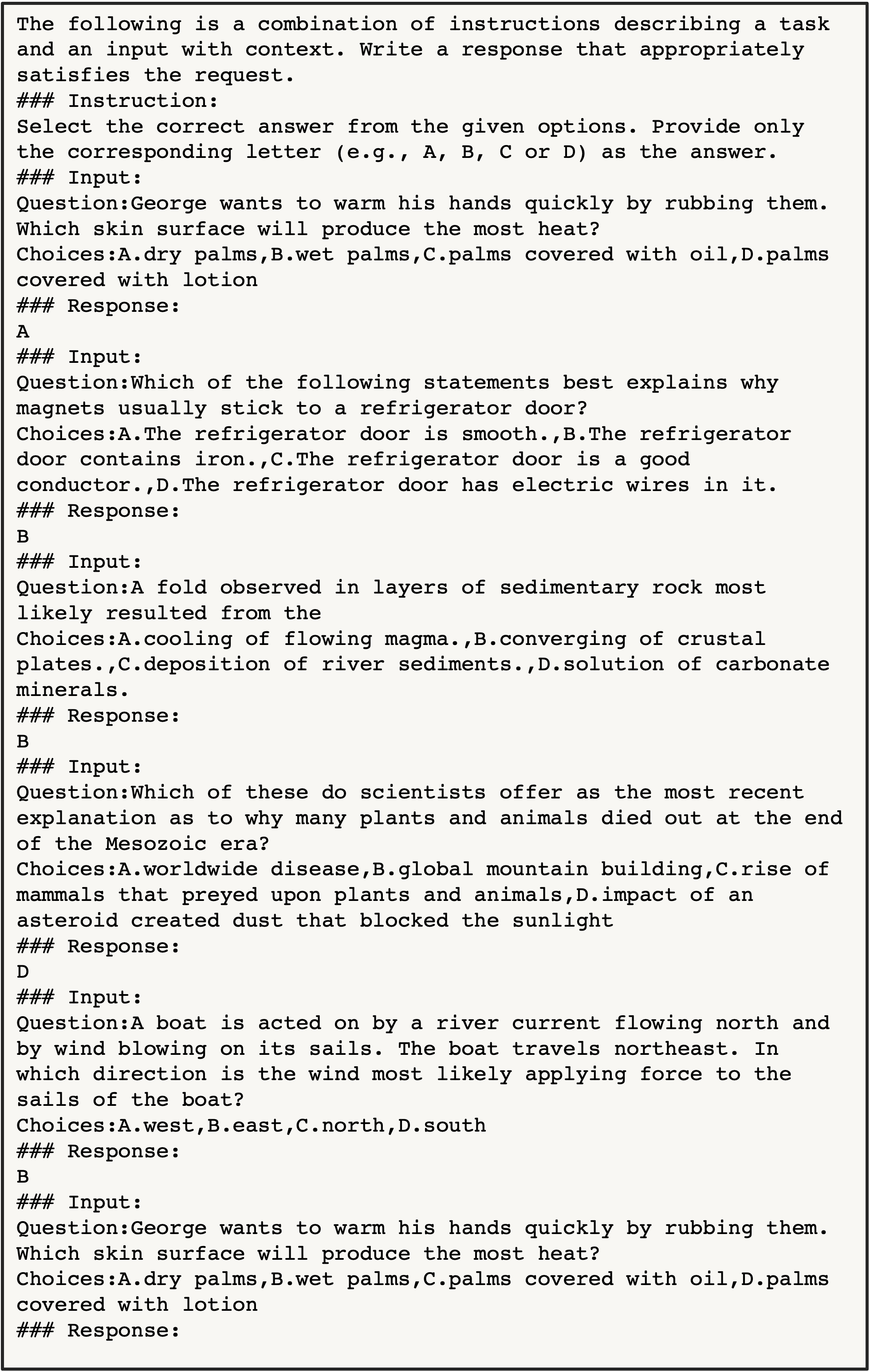}
  \end{subfigure}
  \hfill
  \begin{subfigure}{0.48\linewidth}
    \centering
    \includegraphics[width=\linewidth]{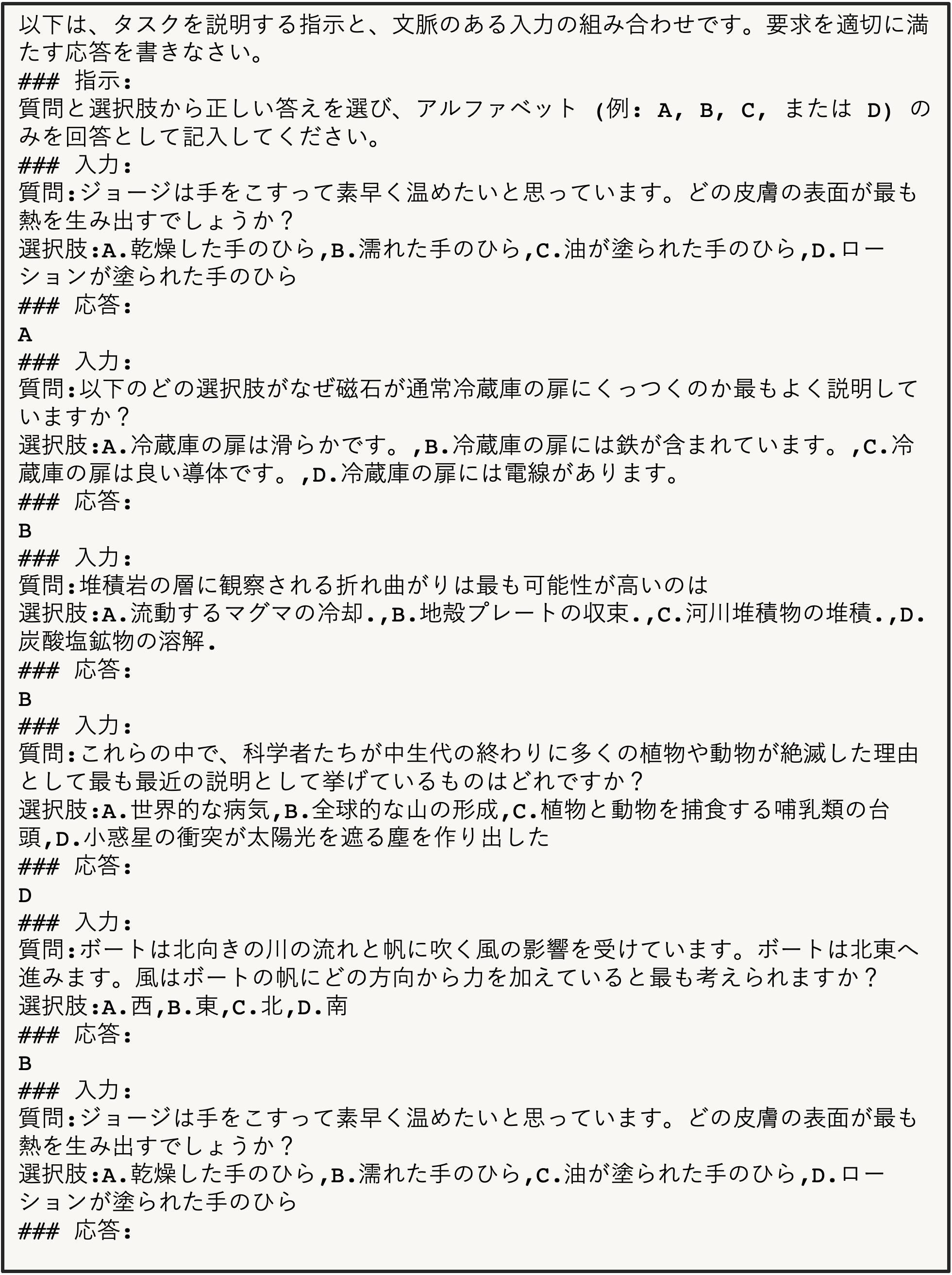}
  \end{subfigure}
  \caption{Example of ARC (English and Japanese) prompt.}
  \label{fig:prompt-example-arc}
\end{figure}

\begin{figure}[htbp]
  \centering
  \begin{subfigure}{0.48\linewidth}
    \centering
    \includegraphics[width=\linewidth]{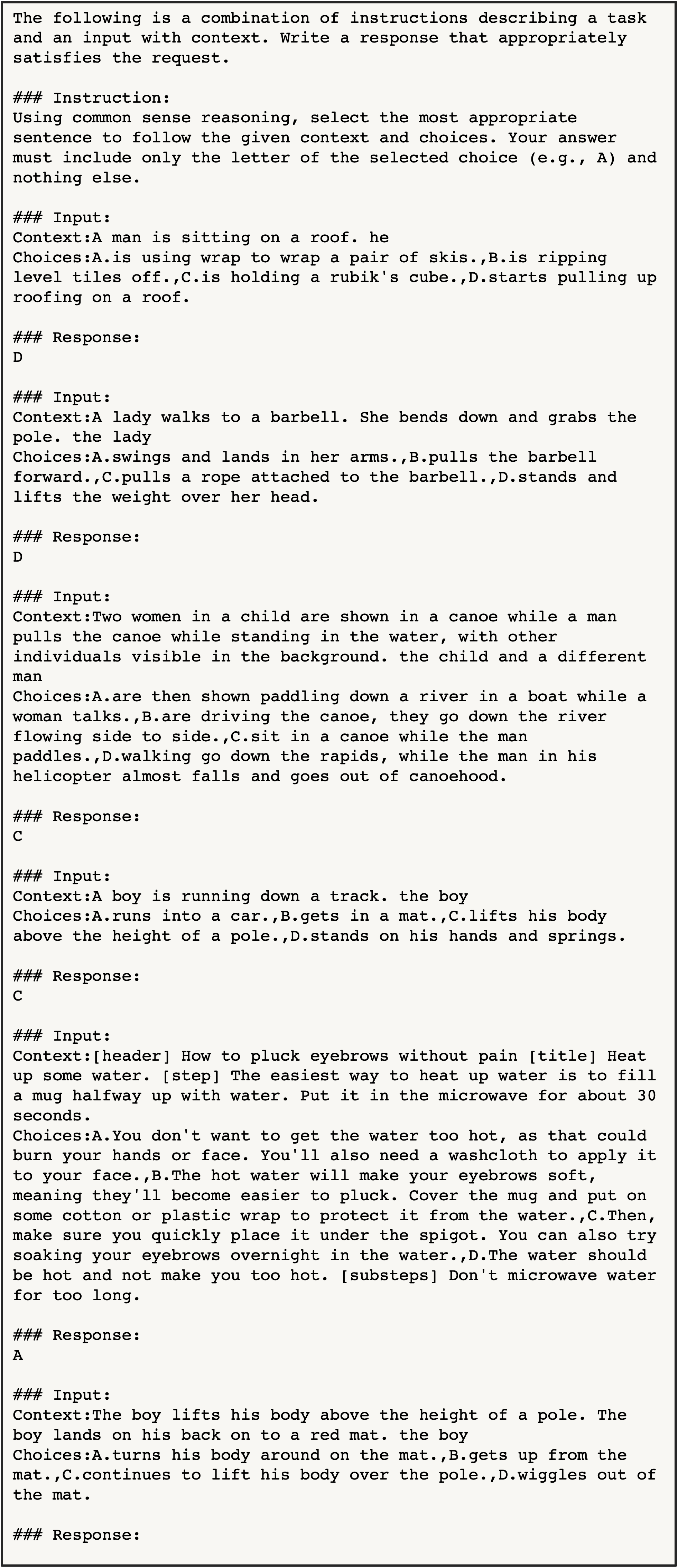}
  \end{subfigure}
  \hfill
  \begin{subfigure}{0.48\linewidth}
    \centering
    \includegraphics[width=\linewidth]{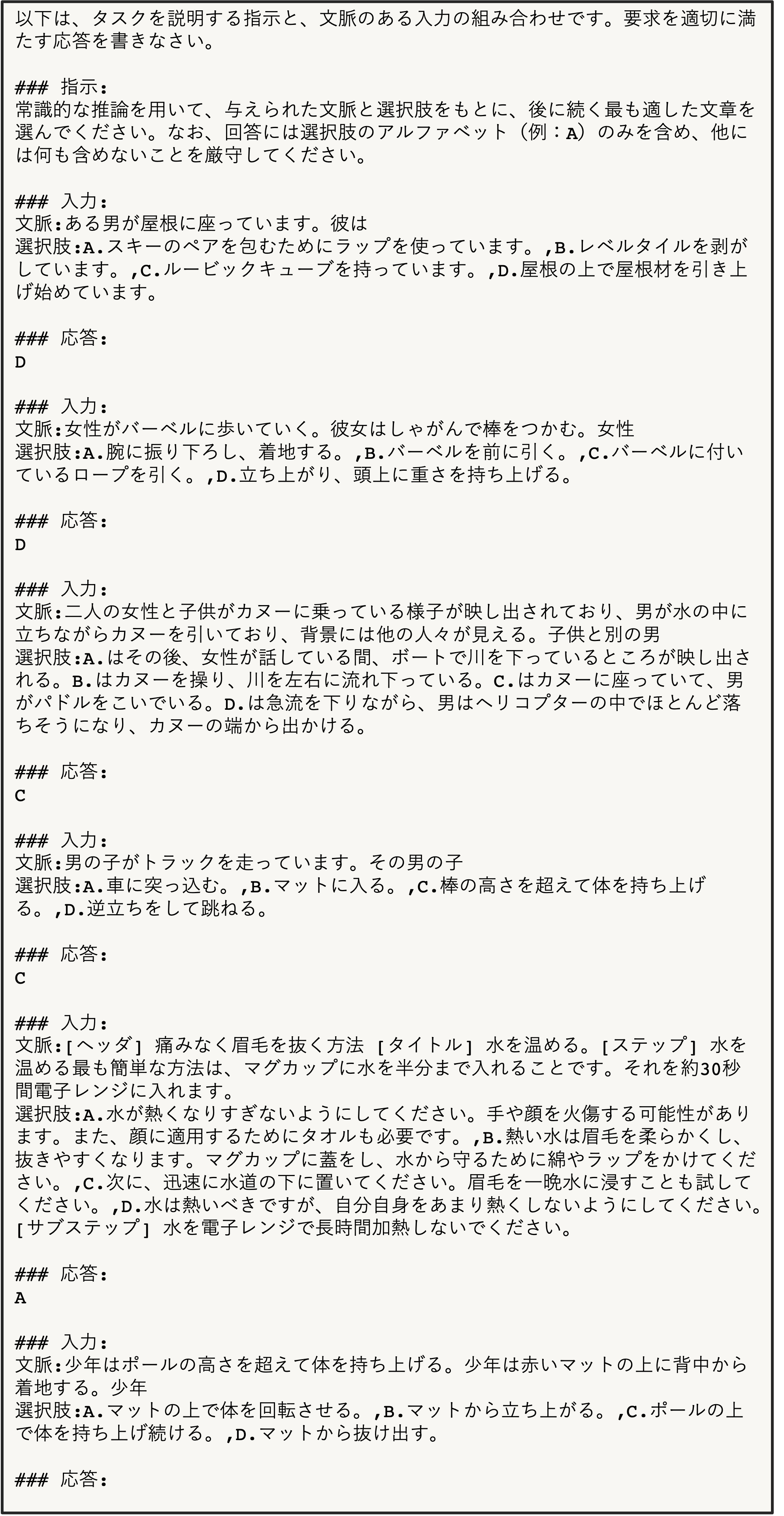}
  \end{subfigure}
  \caption{Example of HellaSwag (English and Japanese) prompt.}
  \label{fig:prompt-example-hellaswag}
\end{figure}

Figures \ref{fig:prompt-example-mmlu}, \ref{fig:prompt-example-csqa}, \ref{fig:prompt-example-arc}, and \ref{fig:prompt-example-hellaswag} show examples of the English and Japanese 5-shot prompts used for each downstream task described in Section \ref{sec:methods-downstream-datasets}.

\renewcommand{\thefigure}{B.\arabic{figure}}
\setcounter{figure}{0}
\clearpage
\section{Additional results}
\label{appendix:additional_results}

\subsection{Learning dynamics of LLMs}
\label{appendix:additional_results:dyamics}
\begin{figure*}[h]
\centering
\includegraphics[width=1\linewidth]{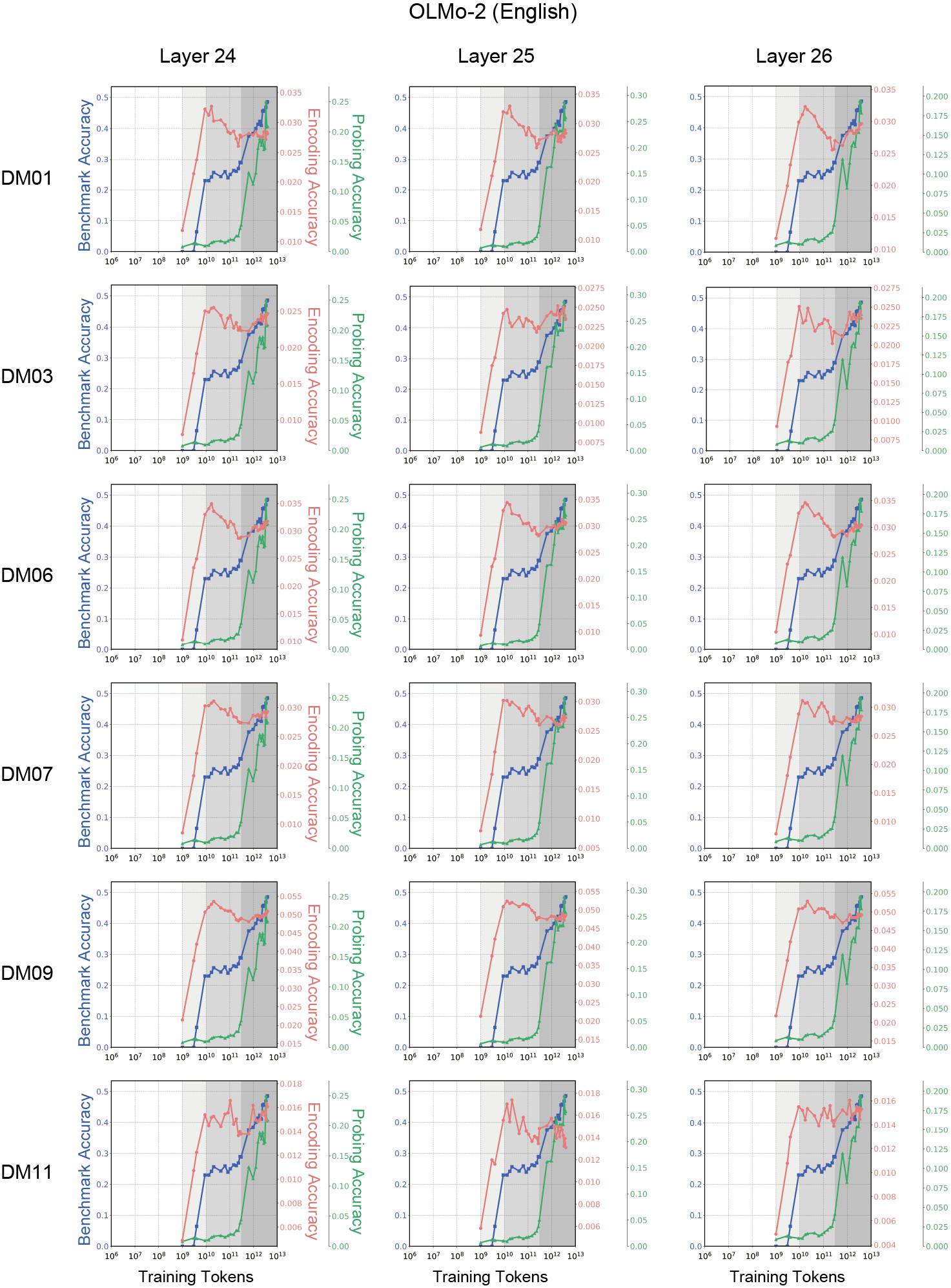}
\caption{Results for all participants regarding learning dynamics of layers 24, 25, 26 of OLMo-2 exhibiting three phase transitions when using English annotation and MMLU.}
\label{fig:subresult-phase-transitions-otherlayers-olmo2-en}
\end{figure*}

\begin{figure*}[h]
\centering
\includegraphics[width=1\linewidth]{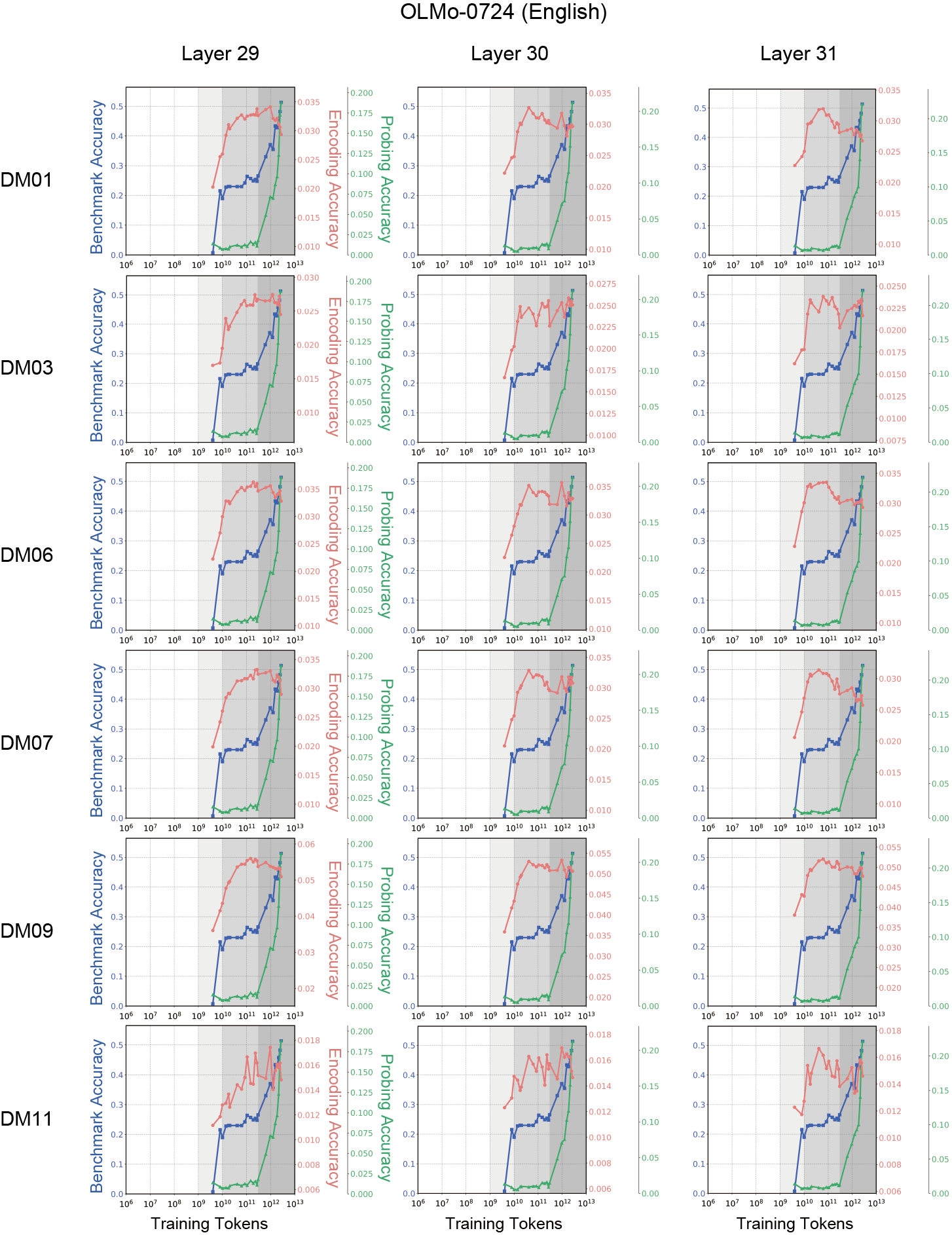}
\caption{Results for all participants regarding learning dynamics of layers 29, 30, 31 of OLMo-0724 exhibiting three phase transitions when using English annotation and MMLU.}
\label{fig:subresult-phase-transitions-otherlayers-olmo0724-en}
\end{figure*}

\begin{figure*}[h]
\centering
\includegraphics[width=1\linewidth]{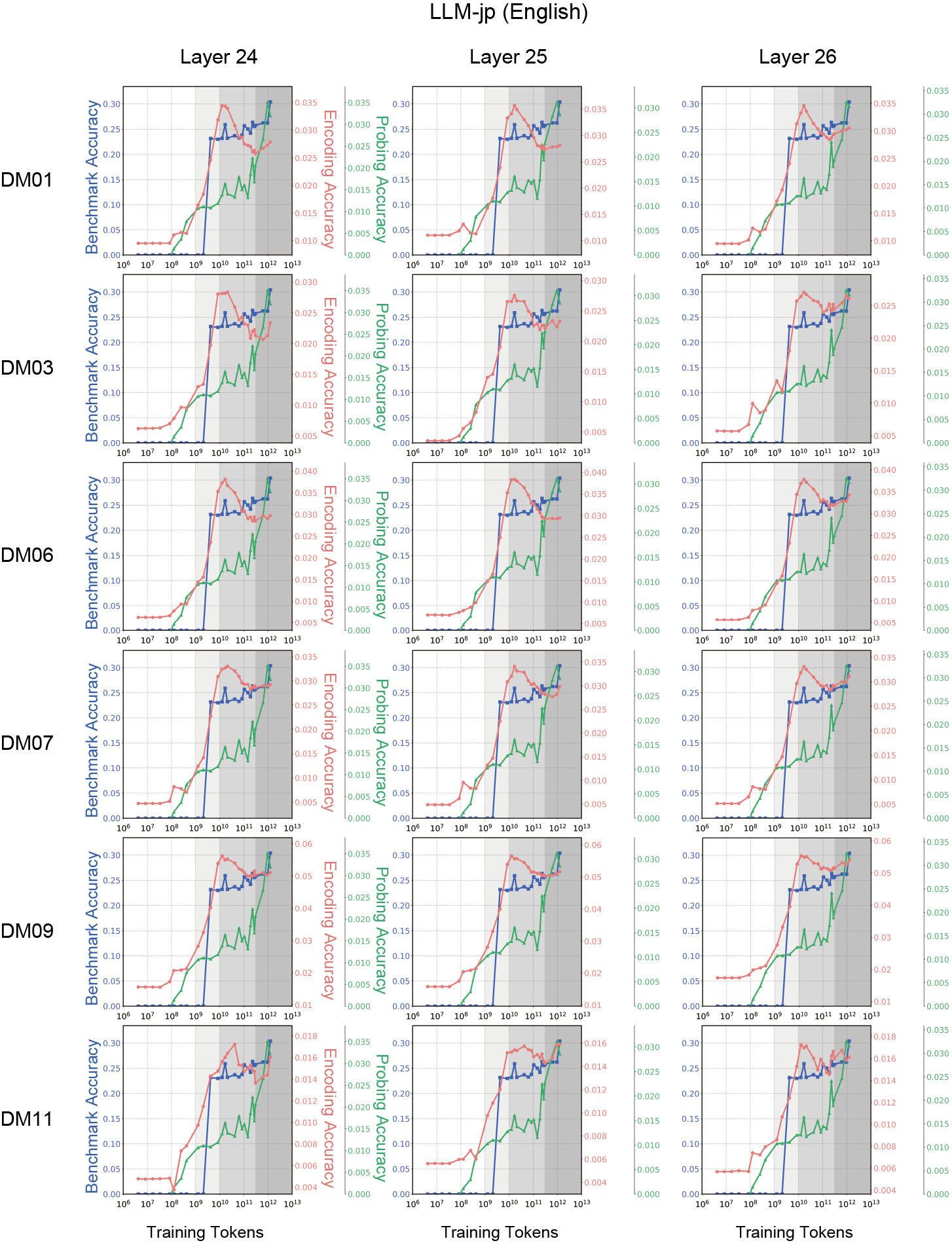}
\caption{Results for all participants regarding learning dynamics of layers 24, 25, 26 of LLM-jp exhibiting three phase transitions when using English annotation and MMLU.}
\label{fig:subresult-phase-transitions-otherlayers-llmjp-en}
\end{figure*}

\begin{figure*}[h]
\centering
\includegraphics[width=1\linewidth]{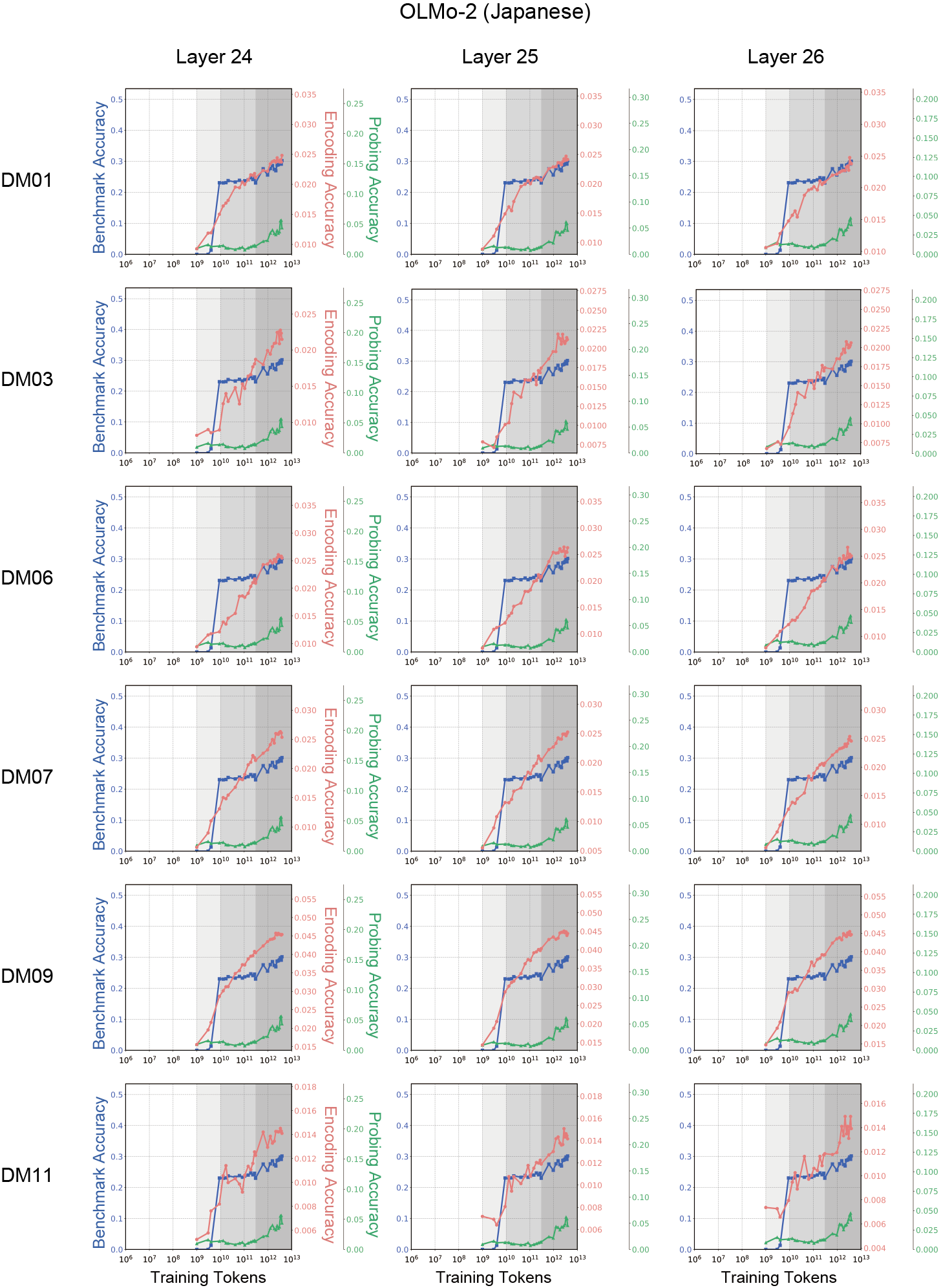}
\caption{Results for all participants regarding learning dynamics of layers 24, 25, 26 of OLMo-2 when using Japanese annotation and MMLU.}
\label{fig:subresult-phase-transitions-otherlayers-olmo2-ja}
\end{figure*}

\begin{figure*}[h]
\centering
\includegraphics[width=1\linewidth]{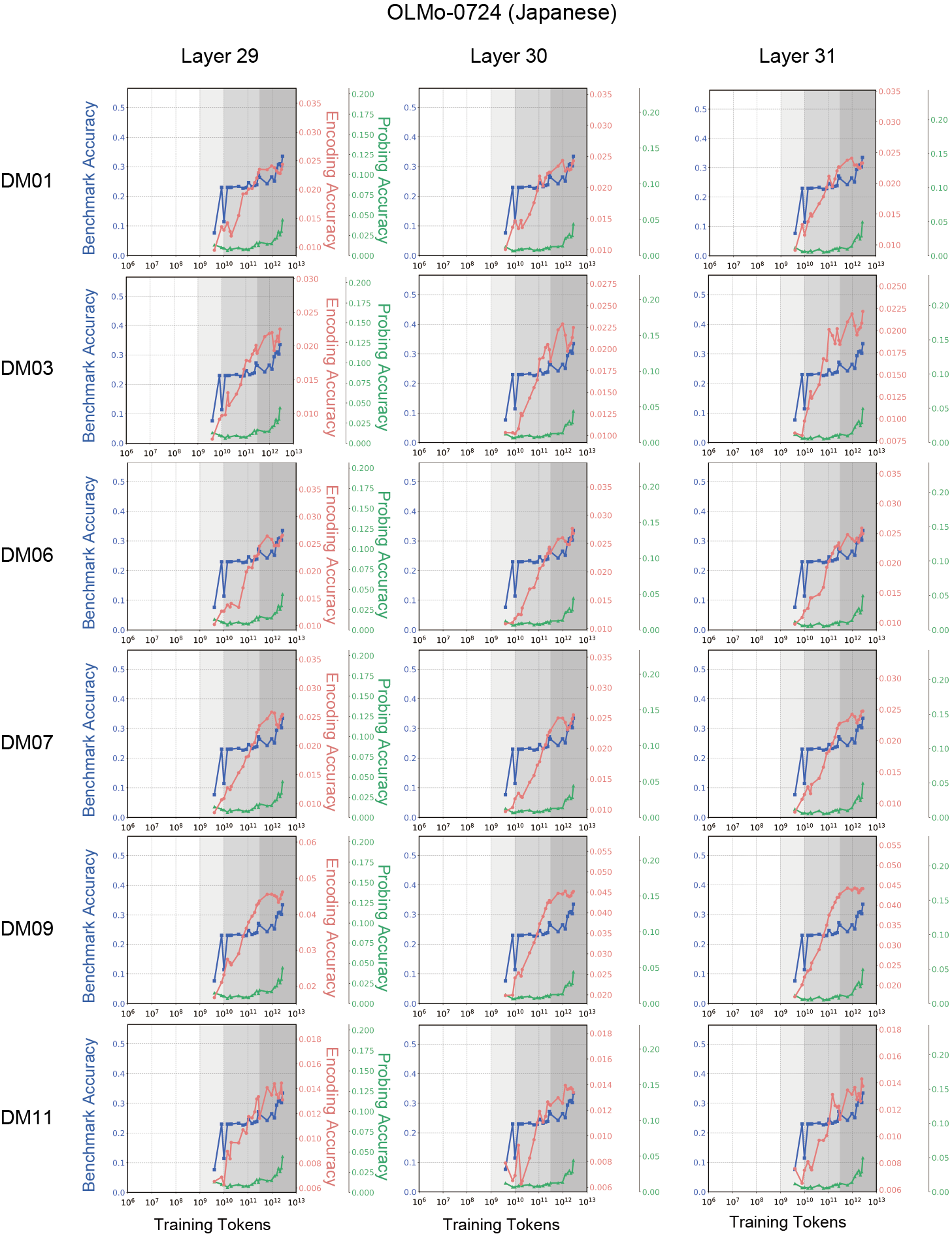}
\caption{Results for all participants regarding learning dynamics of layers 29, 30, 31 of OLMo-0724 when using Japanese annotation and MMLU.}
\label{fig:subresult-phase-transitions-otherlayers-olmo0724-ja}
\end{figure*}

\begin{figure*}[h]
\centering
\includegraphics[width=1\linewidth]{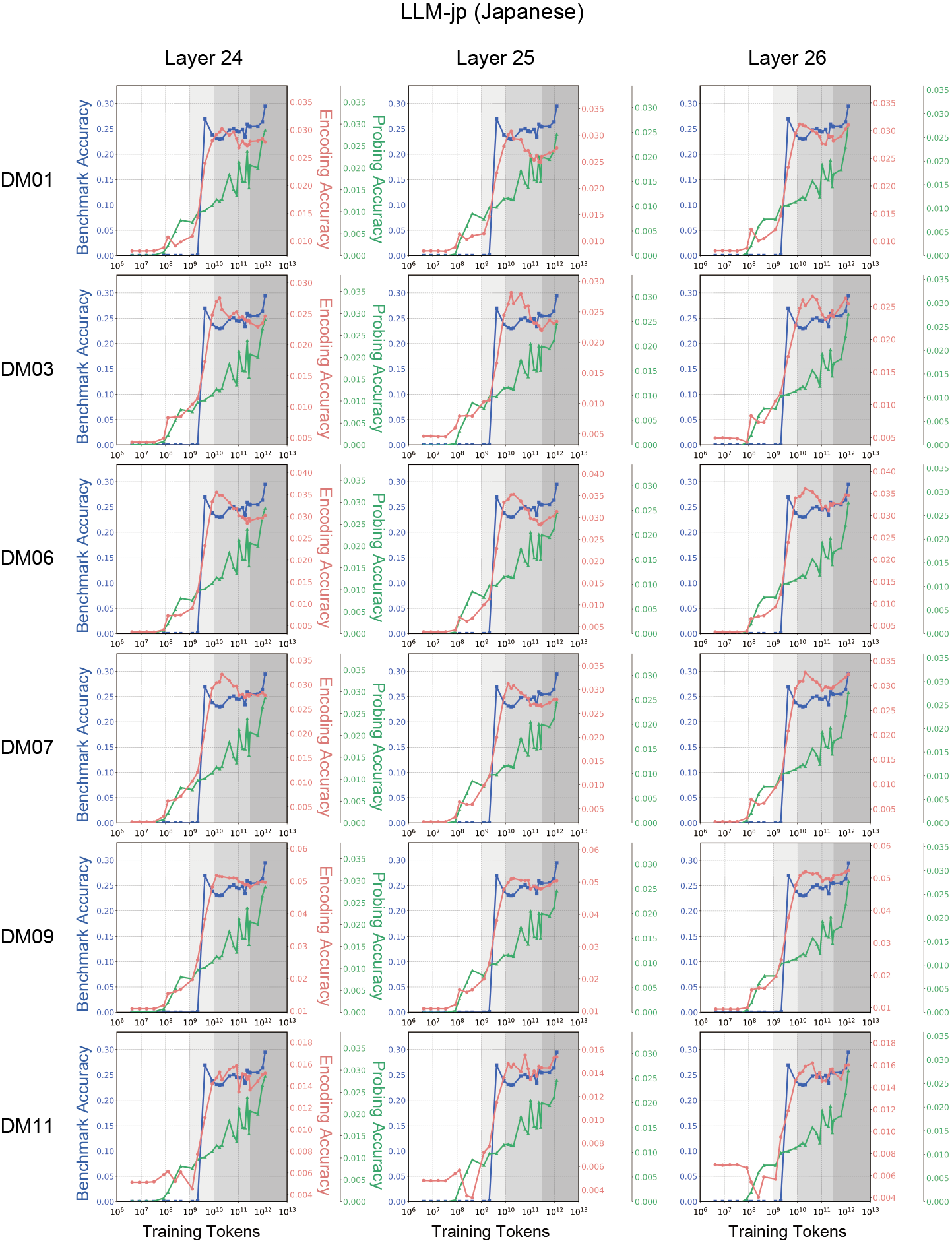}
\caption{Results for all participants regarding learning dynamics of layers 24, 25, 26 of LLM-jp exhibiting three phase transitions when using Japanese annotation and MMLU.}
\label{fig:subresult-phase-transitions-otherlayers-llmjp-ja}
\end{figure*}

\begin{figure*}[h]
\centering
\includegraphics[width=0.85\linewidth]{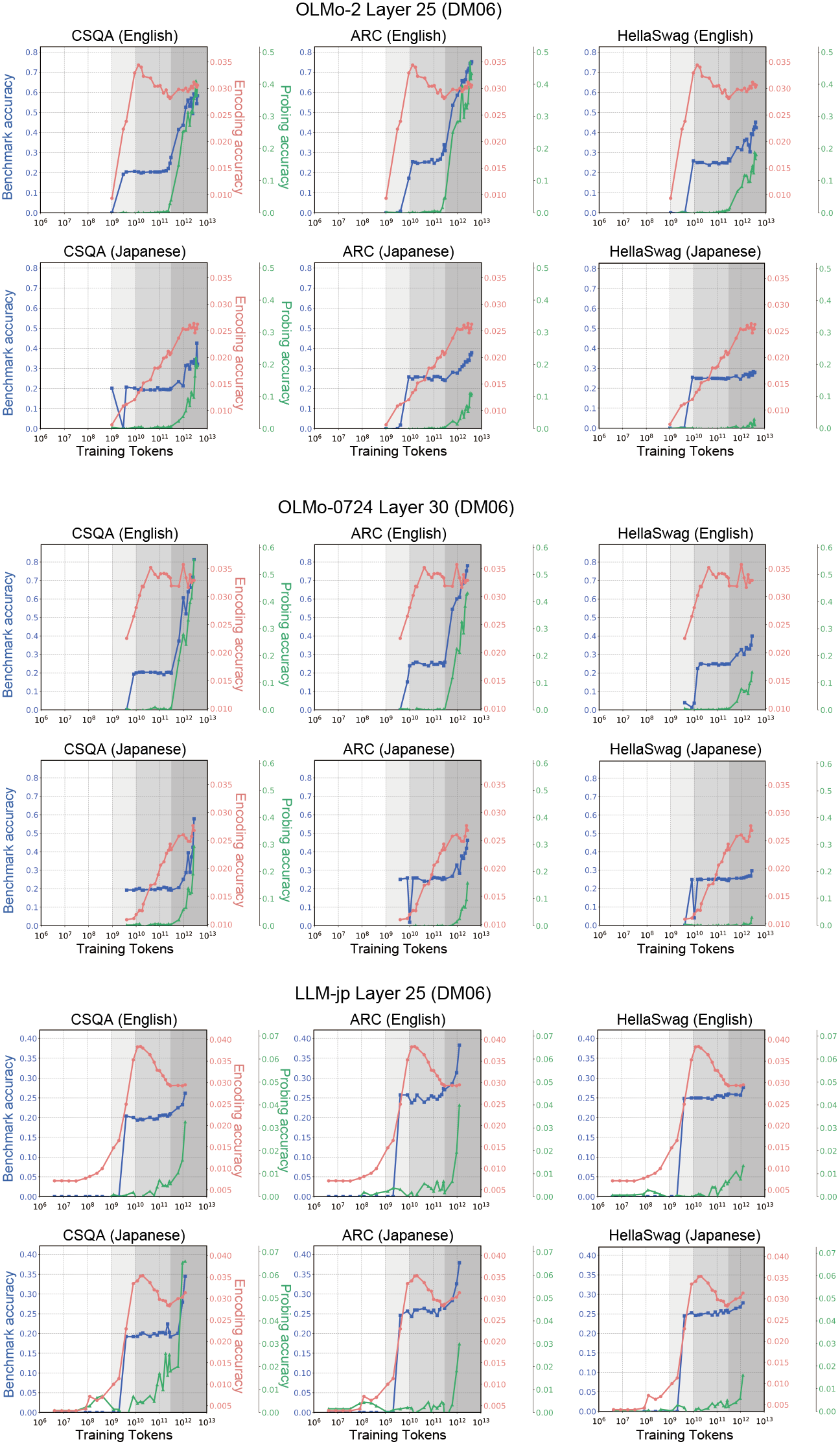}
\caption{Results from a single participant (DM06) for learning dynamics of layer 25 of OLMo-2 and LLM-jp, layer 30 of OLMo-0724 exhibiting three phase transitions when using other tasks.}
\label{fig:subresult-phase-transitions-othertasks}
\end{figure*}

\begin{figure*}[h]
\centering
\includegraphics[width=0.8\linewidth]{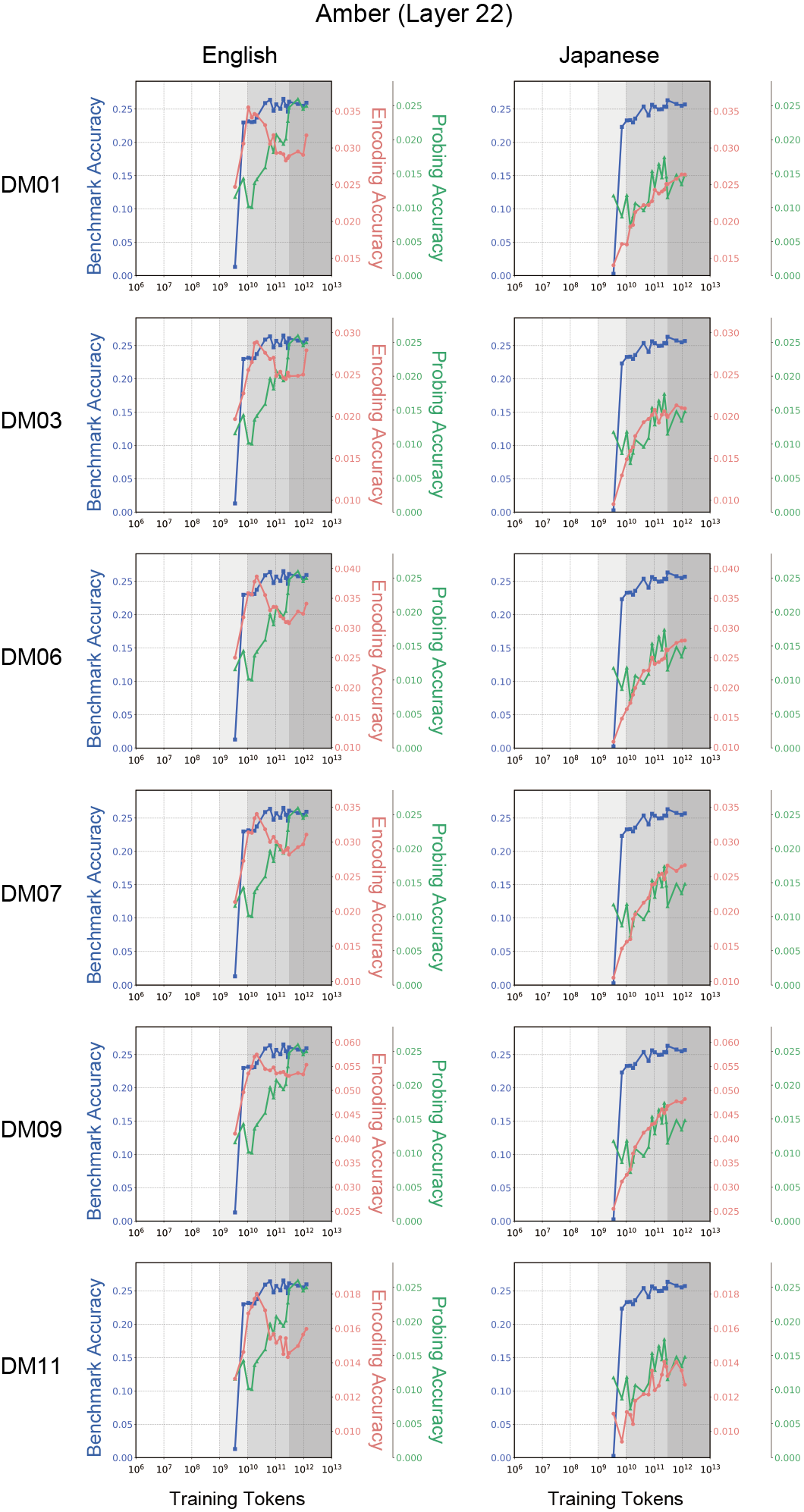}
\caption{Results for all participants regarding learning dynamics of layer 22 of Amber exhibiting three phase transitions when using English/Japanese annotation and MMLU.}
\label{fig:subresult-phase-transitions-othermodel}
\end{figure*}

\begin{figure*}[h]
\centering
\includegraphics[width=1\linewidth]{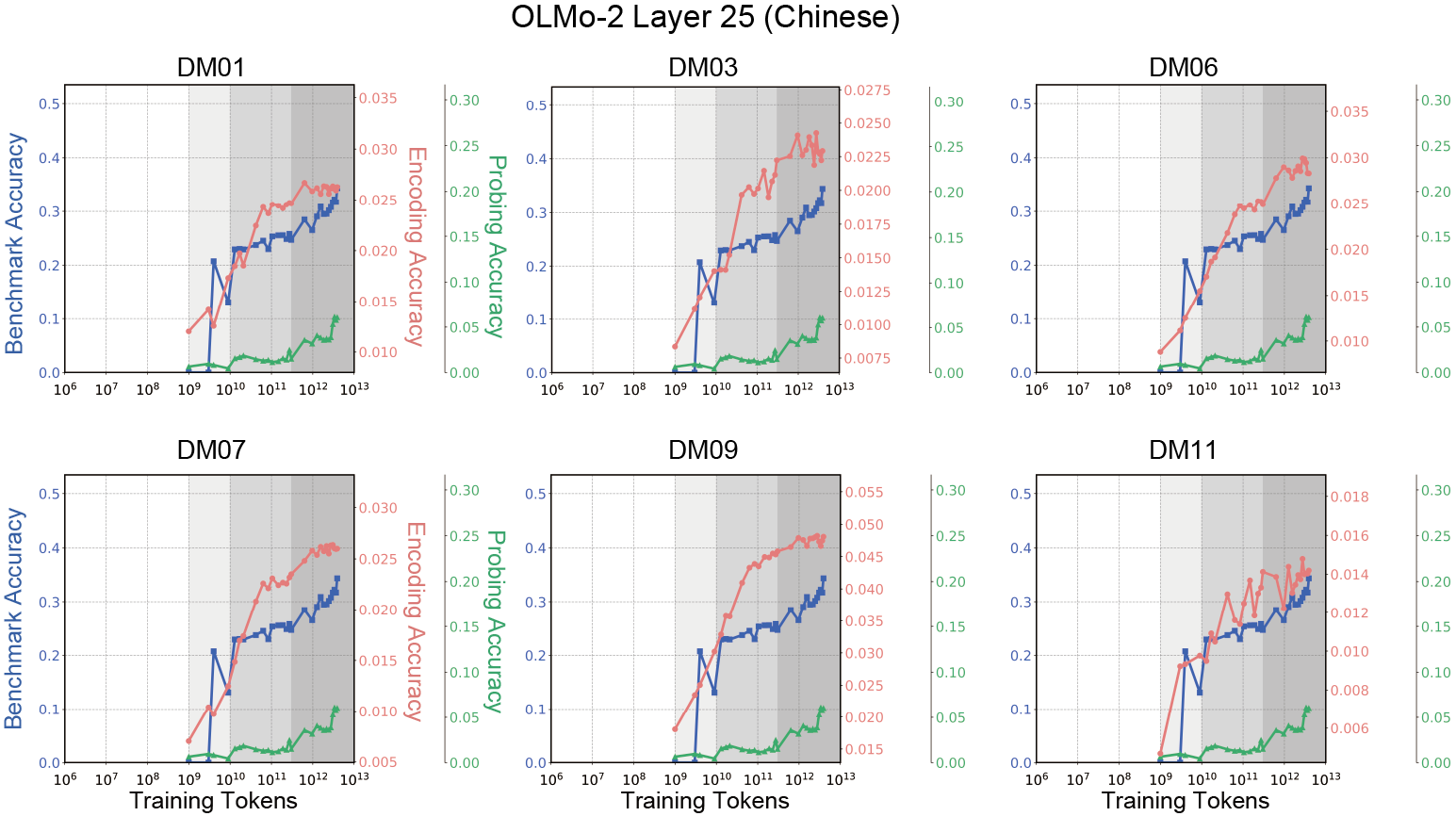}
\caption{Results for all participants regarding learning dynamics of layer 25 of OLMo-2 when using Chinese annotation and MMLU.}
\label{fig:subresult-phase-transitions-otherlanguage}
\end{figure*}

\begin{figure*}[h]
\centering
\includegraphics[width=1\linewidth]{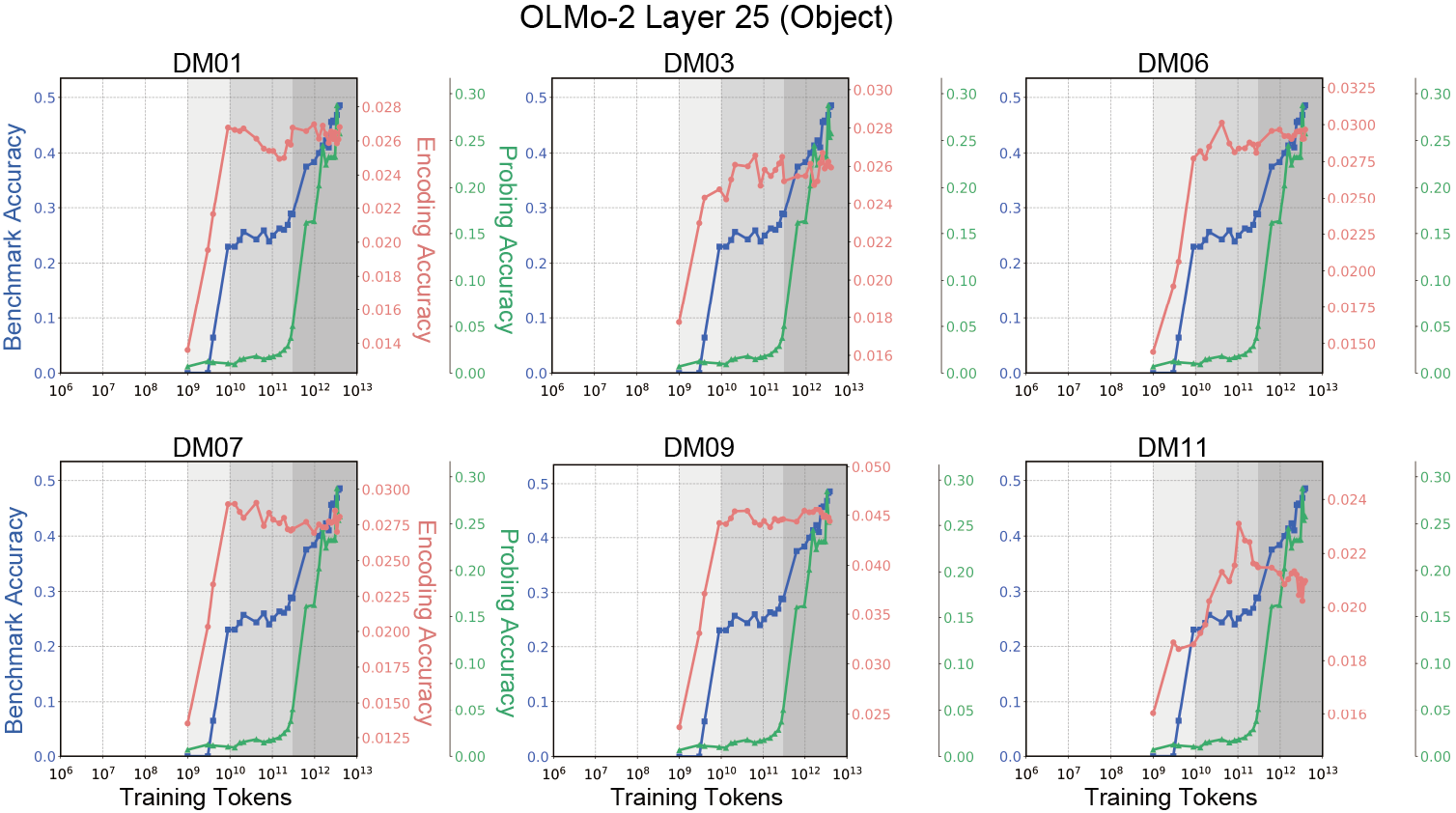}
\caption{Results for all participants regarding learning dynamics of layer 25 of OLMo-2 when using English \textit{Object} annotation and MMLU.}
\label{fig:subresult-phase-transitions-otherannotation}
\end{figure*}

Here, we present supplementary results corresponding to Section~\ref{sec:results-dynamics}. Figures~\ref{fig:subresult-phase-transitions-otherlayers-olmo2-en}, \ref{fig:subresult-phase-transitions-otherlayers-olmo0724-en}, \ref{fig:subresult-phase-transitions-otherlayers-llmjp-en}, \ref{fig:subresult-phase-transitions-otherlayers-olmo2-ja}, \ref{fig:subresult-phase-transitions-otherlayers-olmo0724-ja}, and \ref{fig:subresult-phase-transitions-otherlayers-llmjp-ja} illustrate outcomes for all participants and main LLMs (OLMo-2, OLMo-0724, LLM-jp) when using layers adjacent to those of Figure~\ref{fig:result-phase-transtions}. We demonstrate that comparable encoding results can be obtained using these layers. Figure~\ref{fig:subresult-phase-transitions-othertasks} displays the outcomes of DM06 and the main LLMs (OLMo-2, OLMo-0724, LLM-jp) on different downstream tasks (CSQA, ARC, HellaSwag) to those of Figure~\ref{fig:result-phase-transtions}, showing that similar benchmark and probing results are achieved for these alternate tasks. Figure~\ref{fig:subresult-phase-transitions-othermodel} presents the results obtained using Amber, a different LLM from those employed for Figure~\ref{fig:result-phase-transtions}. We confirm a phase transition in encoding accuracy around layer 22; however, downstream task performance by Amber only extends to instruction-following, and thus the increase in probing accuracy observed in Phase~3 is slightly attenuated. Figure~\ref{fig:subresult-phase-transitions-otherlanguage} reports the results obtained when the LLMs are given input in Chinese, a different language to that used for Figure~\ref{fig:result-phase-transtions}. This figure indicates that the phase transition is not observed in a language on which the model has not been trained. Finally, Figure~\ref{fig:subresult-phase-transitions-otherannotation} presents the results of using the \textit{Object} annotation, which differs from that used for Figure~\ref{fig:result-phase-transtions}; here, too, we confirm a similar phase transition.

\subsection{Changes in the relationship with the brain}
\label{sec:additional_results-encoding}
\begin{figure*}[t]
\centering
\includegraphics[width=1\linewidth]{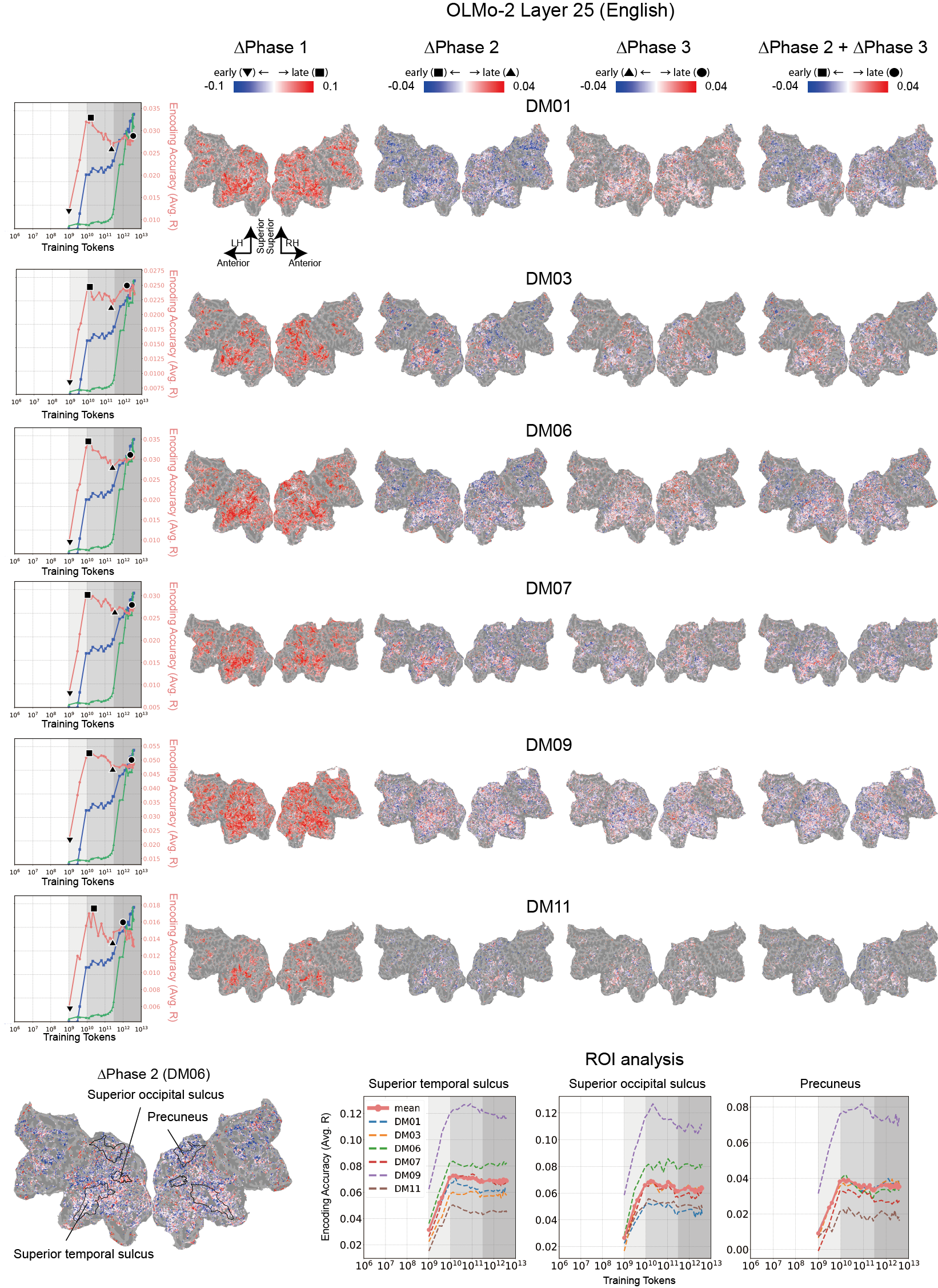}
\caption{Results for all participants regarding changes in the relationship with the brain using layer 25 of OLMo-2 and English annotation and MMLU. The line graph below shows the results of the ROI-level analysis. The vertical axis denotes the average encoding accuracy of significant voxels within each ROI. The solid line represents the mean across participants, and the dashed lines represent individual participant results. Black contours show several representative, anatomically defined ROIs.}
\label{fig:subresult-flatmap-olmo-2-en}
\end{figure*}

\begin{figure*}[t]
\centering
\includegraphics[width=1\linewidth]{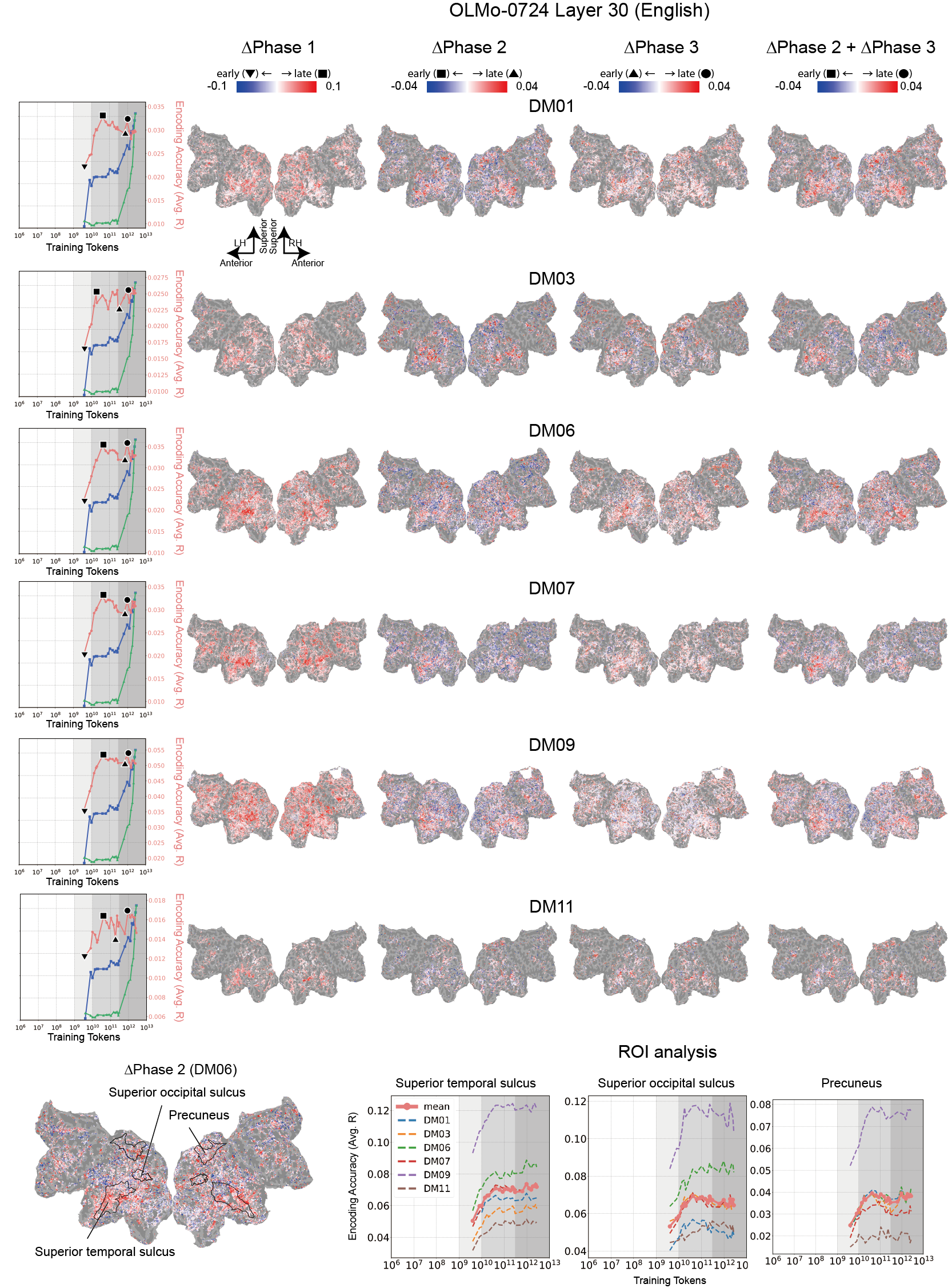}
\caption{Results for all participants regarding changes in the relationship with the brain using layer 30 of OLMo-0724 and English annotation and MMLU.}
\label{fig:subresult-flatmap-olmo-0724-en}
\end{figure*}

\begin{figure*}[t]
\centering
\includegraphics[width=1\linewidth]{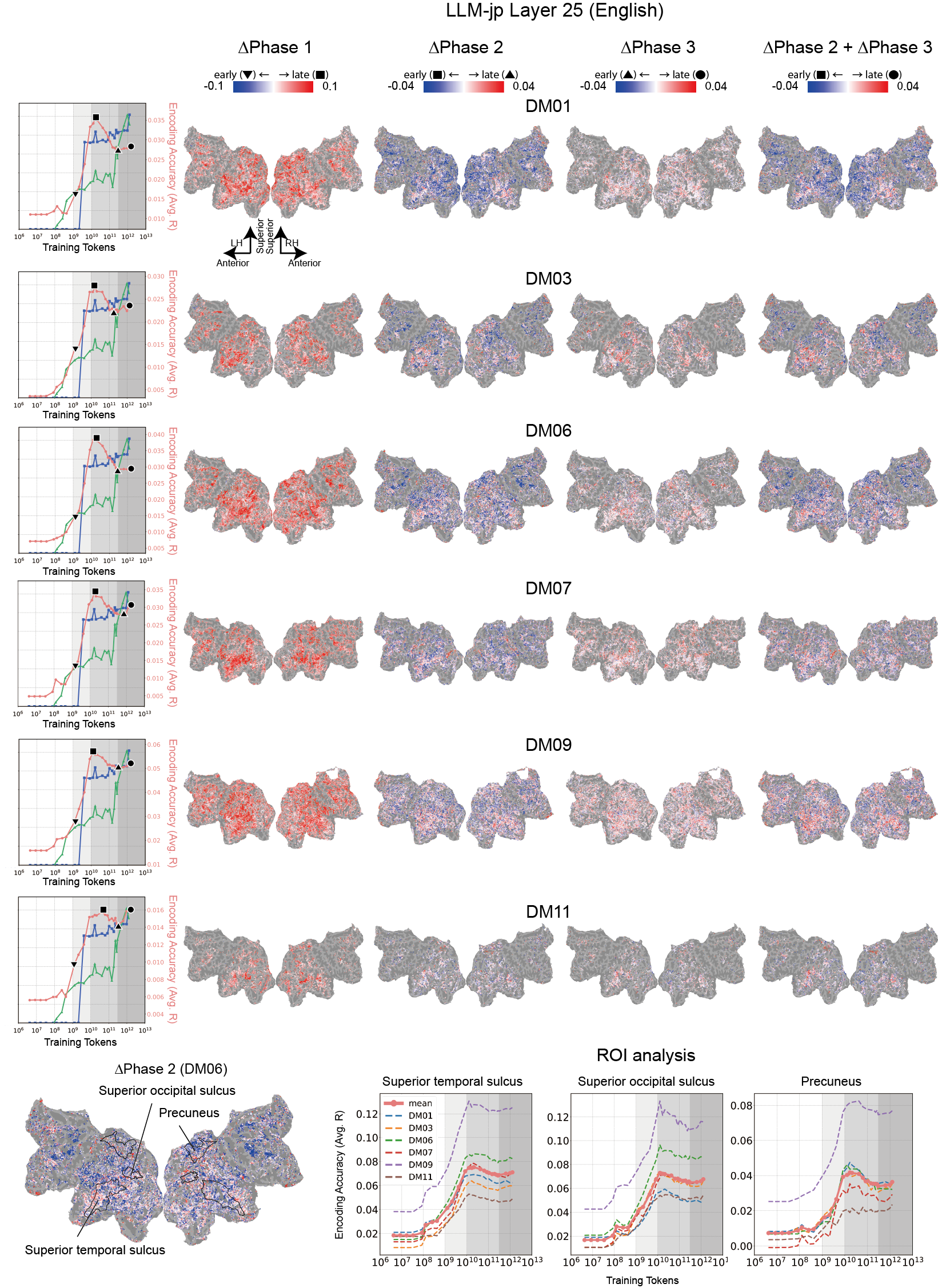}
\caption{Results for all participants regarding changes in the relationship with the brain using layer 25 of LLM-jp and English annotation and MMLU.}
\label{fig:subresult-flatmap-llmjp-en}
\end{figure*}

\begin{figure*}[t]
\centering
\includegraphics[width=1\linewidth]{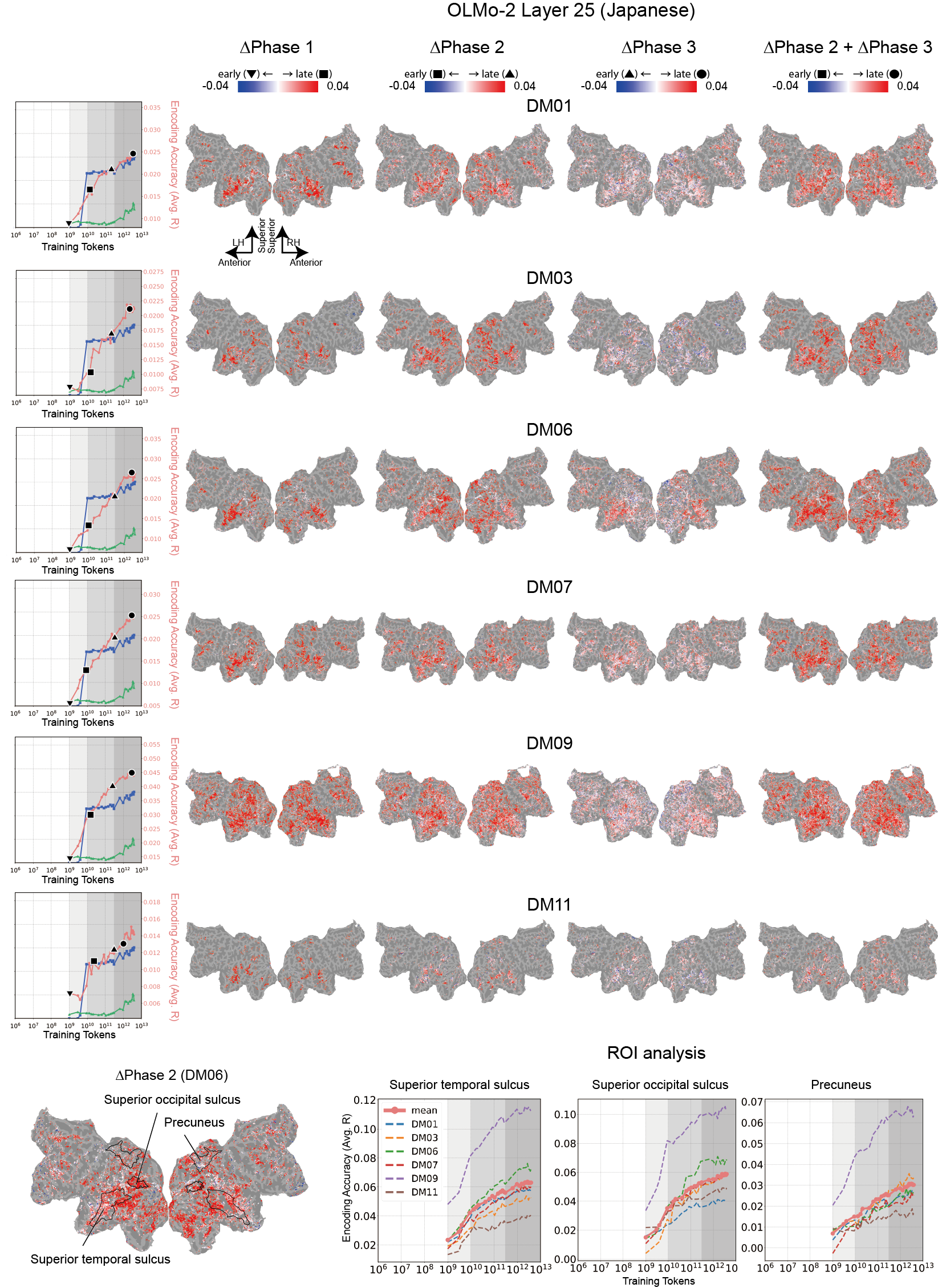}
\caption{Results for all participants regarding changes in the relationship with the brain using layer 25 of OLMo-2 and Japanese annotation and MMLU.}
\label{fig:subresult-flatmap-olmo-2-ja}
\end{figure*}

\begin{figure*}[t]
\centering
\includegraphics[width=1\linewidth]{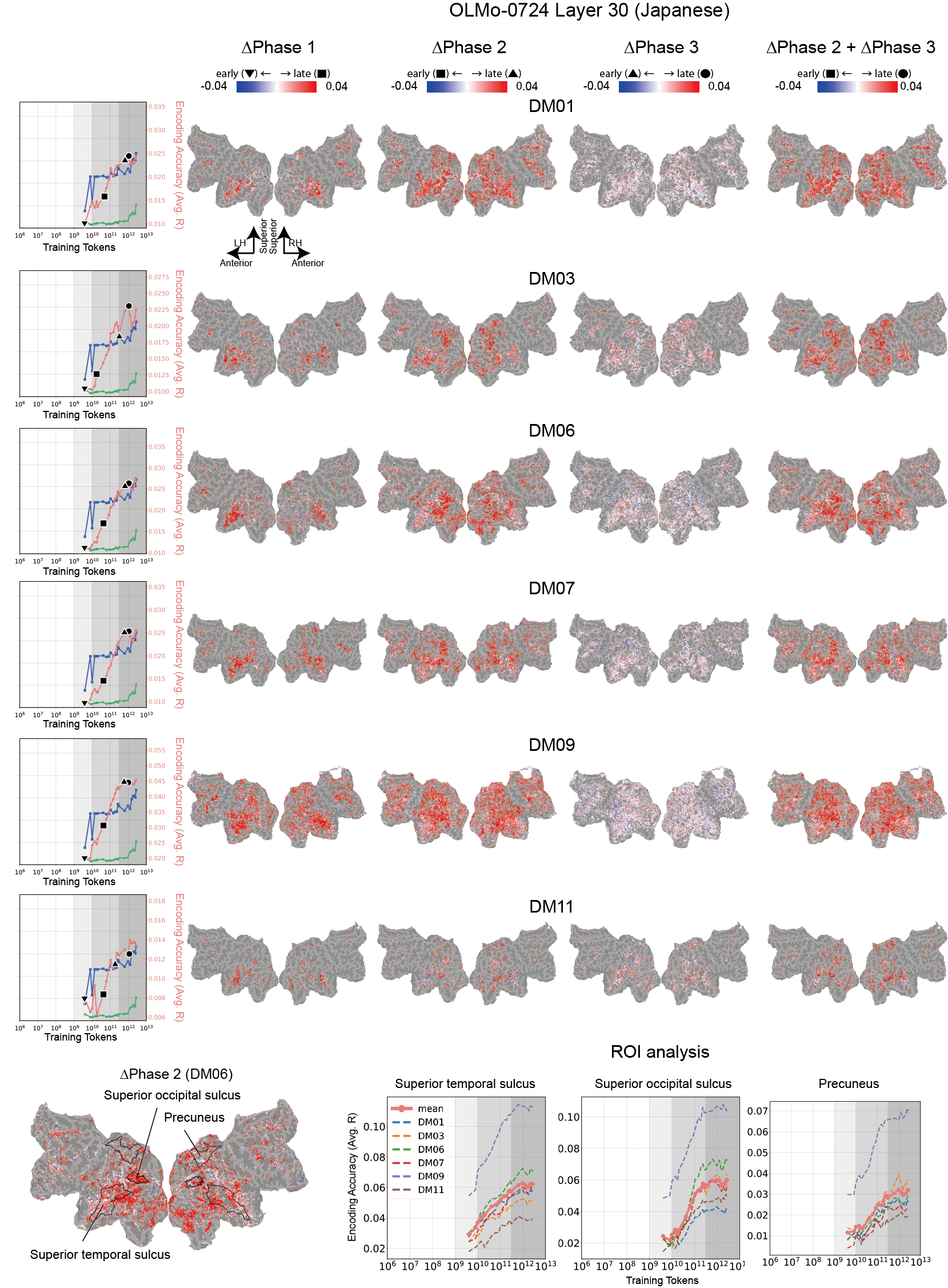}
\caption{Results for all participants regarding changes in the relationship with the brain using layer 30 of OLMo-0724 and Japanese annotation and MMLU.}
\label{fig:subresult-flatmap-olmo-0724-ja}
\end{figure*}

\begin{figure*}[t]
\centering
\includegraphics[width=1\linewidth]{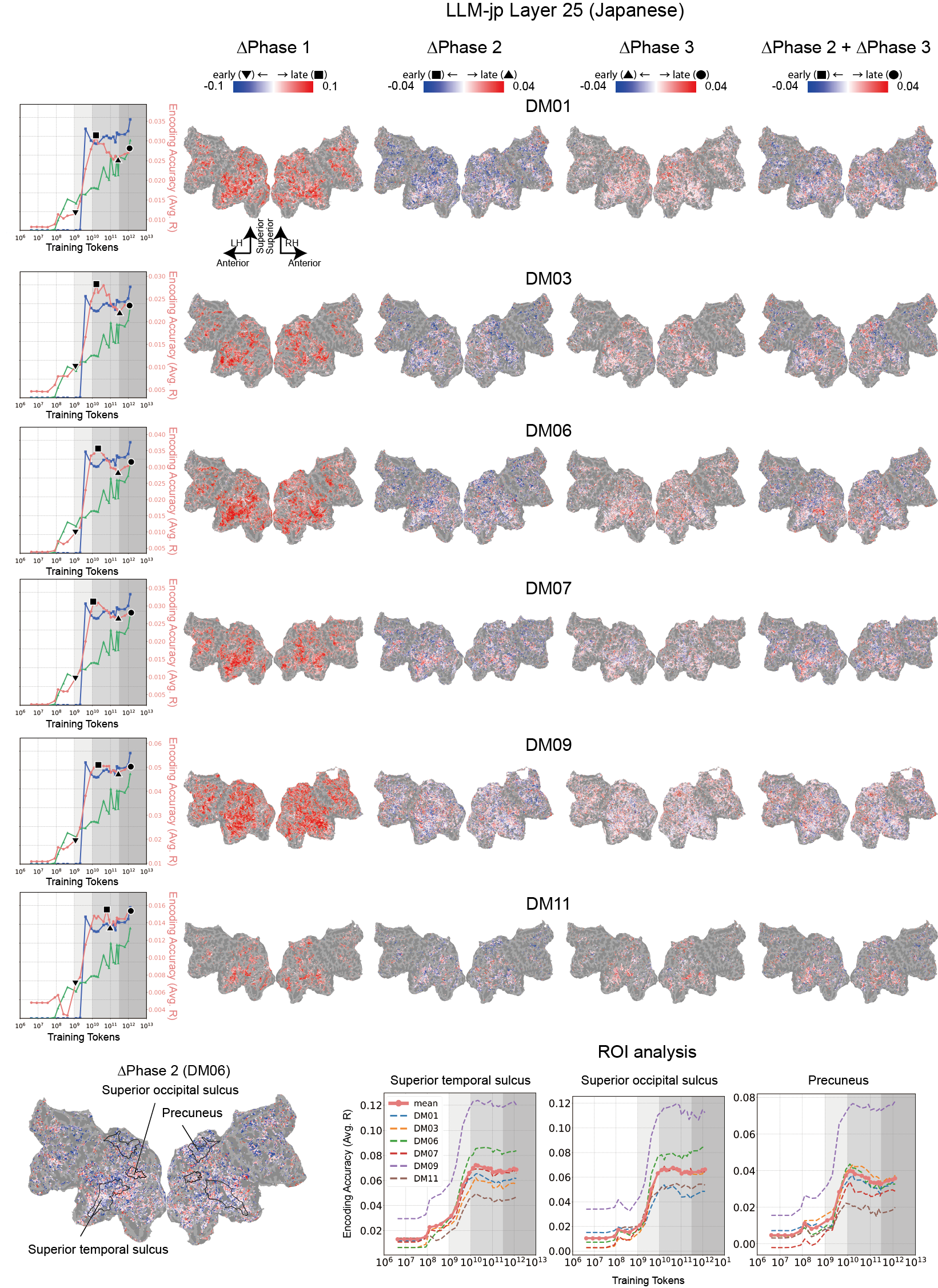}
\caption{Results for all participants regarding changes in the relationship with the brain using layer 25 of LLM-jp and Japanese annotation and MMLU.}
\label{fig:subresult-flatmap-llmjp-ja}
\end{figure*}

In this section, we present supplementary findings related to Section \ref{sec:results-encoding}. Figures \ref{fig:subresult-flatmap-olmo-2-en}, \ref{fig:subresult-flatmap-olmo-0724-en}, \ref{fig:subresult-flatmap-llmjp-en}, \ref{fig:subresult-flatmap-olmo-2-ja}, \ref{fig:subresult-flatmap-olmo-0724-ja}, and \ref{fig:subresult-flatmap-llmjp-ja} illustrate the outcomes for all participants and main LLMs (OLMo-2, OLMo-0724, LLM-jp). We show that similar results are observed for every participant. Furthermore, when the language in which the LLM was trained is used as input, the results are similar to those depicted in Figure \ref{fig:result-flatmap} across the three LLMs. 

The line graphs presented in Figures \ref{fig:subresult-flatmap-olmo-2-en}--\ref{fig:subresult-flatmap-llmjp-ja} illustrate how the voxel-wise changes aggregate within three major ROIs. Across participants, Phase 1 shows that there is a sharp increase in accuracy for all ROIs, while Phases 2 and 3 exhibit more localized or modest changes. Although some voxels within the temporal and occipital cortex exhibit higher prediction accuracy after Phase 3 than after Phase 1, this trend is less evident at the ROI level. This discrepancy suggests that finer-grained voxel-level analyses may be more sensitive than coarse ROI-level approaches in capturing such effects.

\subsection{Evolution of LLM internal representations for downstream tasks}
\label{sec:additional_results-probing}
\begin{figure*}[t]
\centering
\includegraphics[width=1\linewidth]{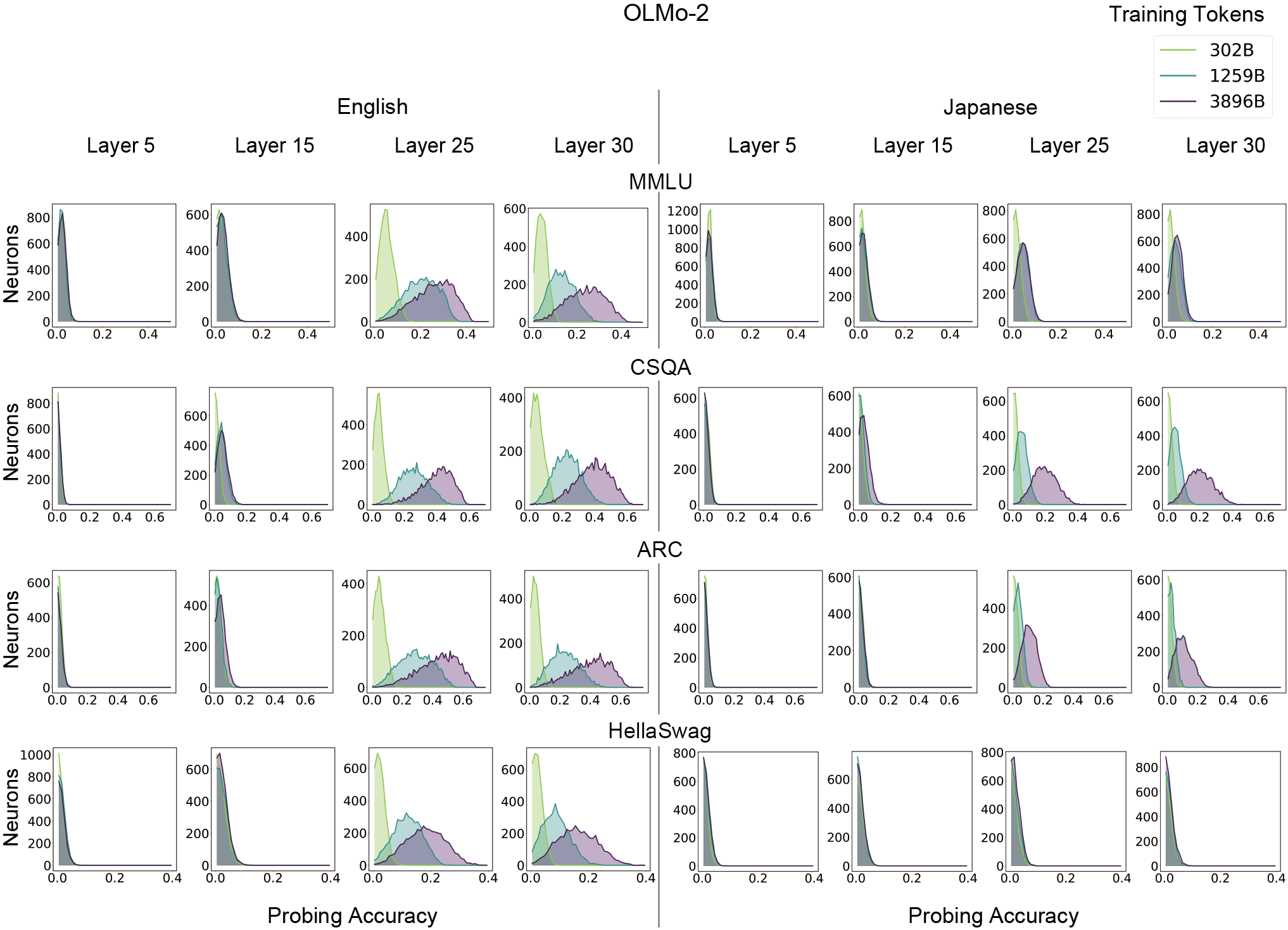}
\caption{Results for all downstream tasks regarding changes in the activations of OLMo-2.}
\label{fig:subresult-probing-histplot-olmo2}
\end{figure*}

\begin{figure*}[t]
\centering
\includegraphics[width=1\linewidth]{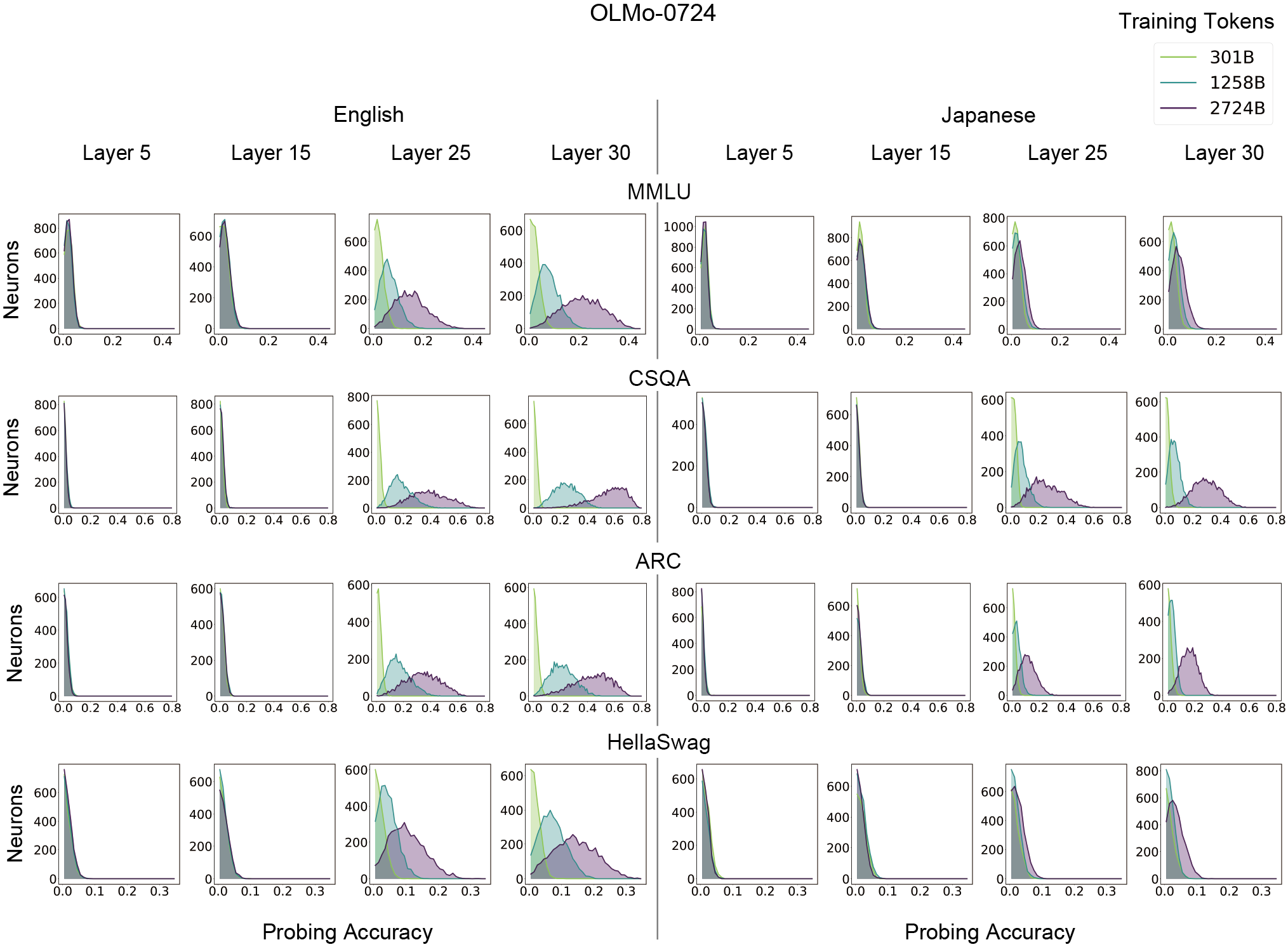}
\caption{Results for all downstream tasks regarding changes in the activations of OLMo-0724.}
\label{fig:subresult-probing-histplot-olmo0724}
\end{figure*}

\begin{figure*}[t]
\centering
\includegraphics[width=1\linewidth]{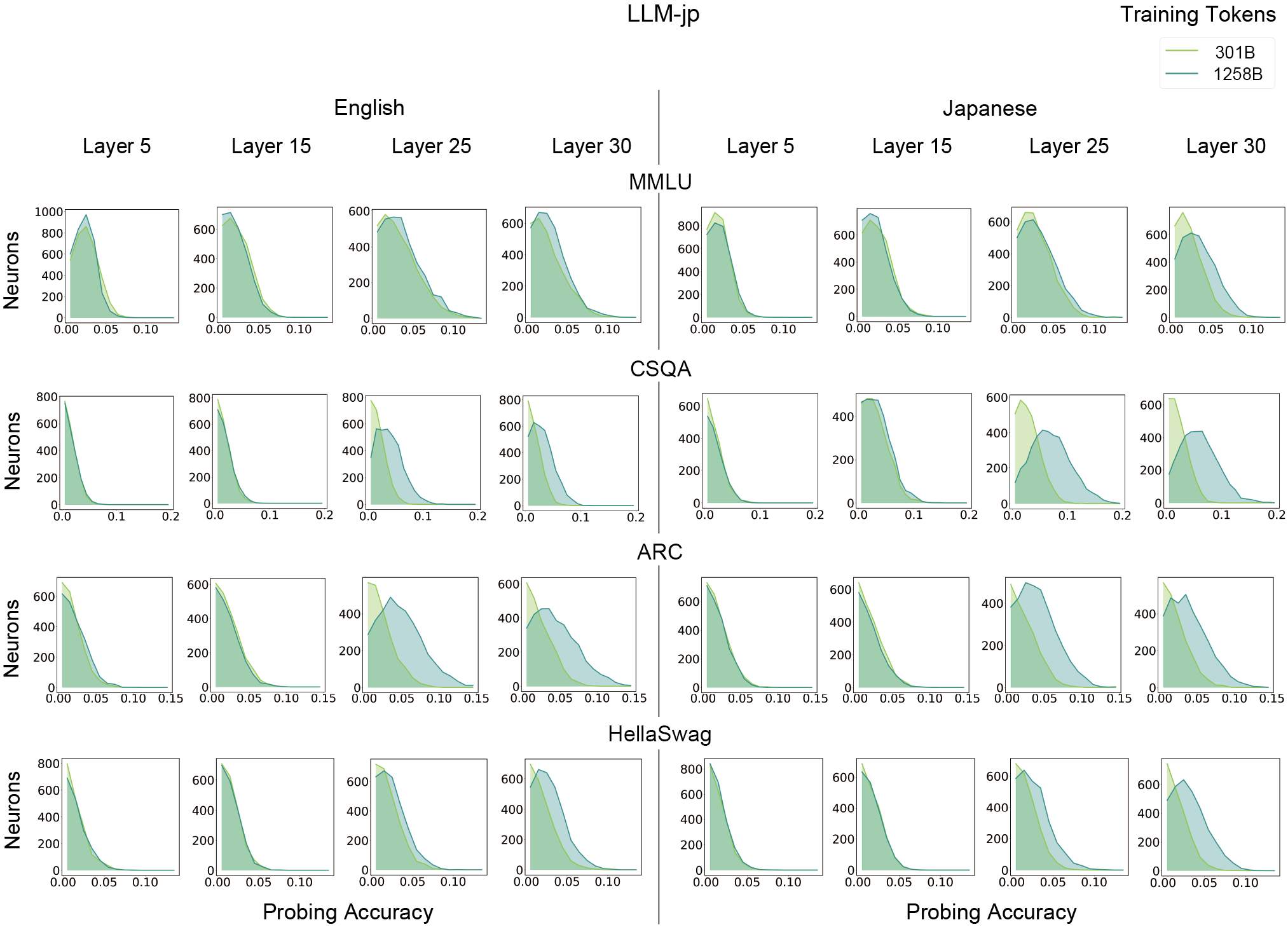}
\caption{Results for all downstream tasks regarding changes in the activations of LLM-jp.}
\label{fig:subresult-probing-histplot-llmjp}
\end{figure*}

\begin{figure*}[t]
\centering
\includegraphics[width=1\linewidth]{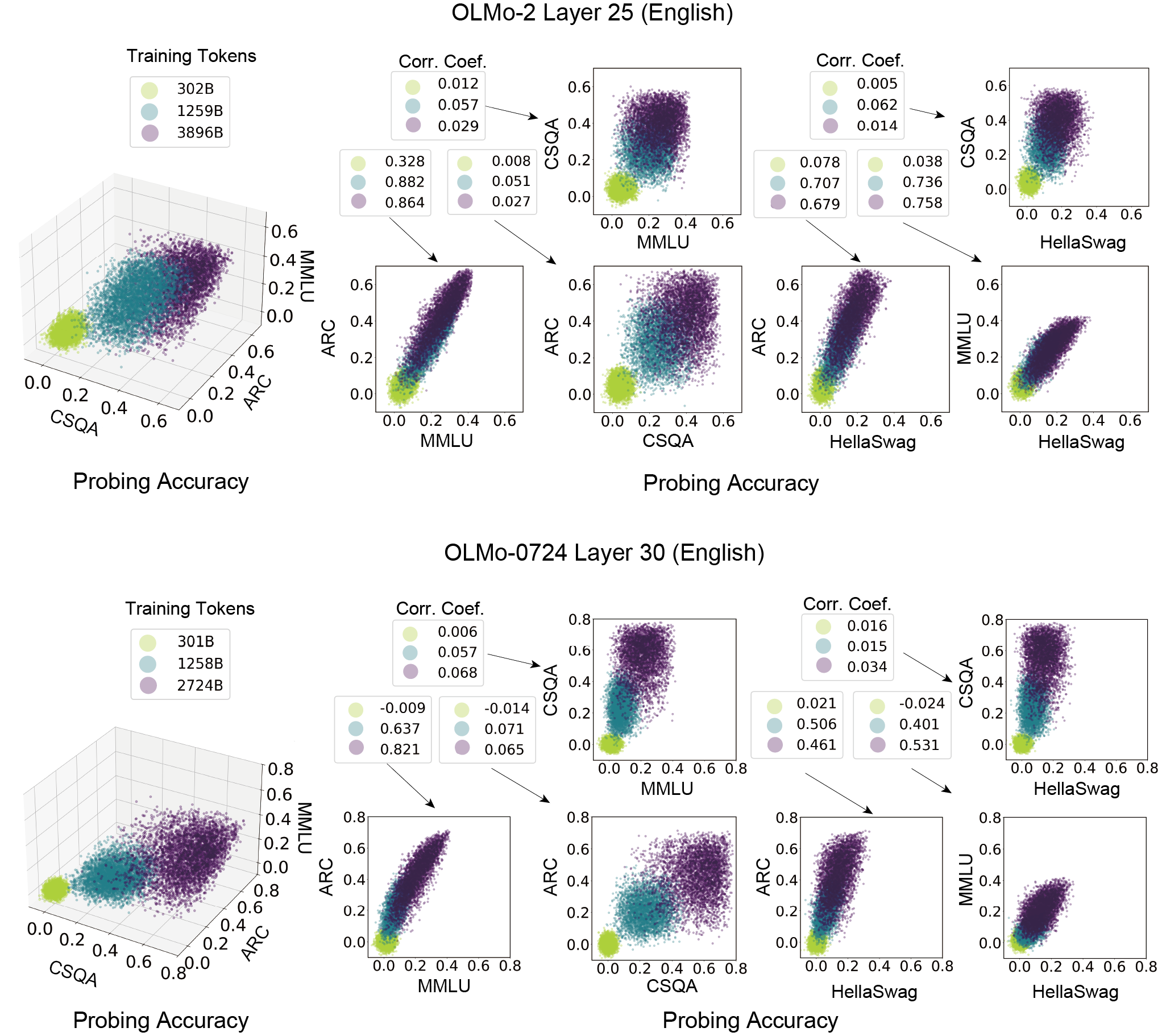}
\caption{Relationship between probing accuracies in OLMo-2 (layer 25) and OLMo-0724 (layer 30) across English MMLU, CSQA, ARC, and HellaSwag.}
\label{fig:subresult-probing-scatter}
\end{figure*}

In this section, we present supplementary findings related to Section~\ref{sec:results-probing}. Figures~\ref{fig:subresult-probing-histplot-olmo2},  \ref{fig:subresult-probing-histplot-olmo0724}, and \ref{fig:subresult-probing-histplot-llmjp} correspond to Figure~\ref{fig:result-probing}a for the main LLMs (OLMo-2, OLMo-0724, LLM-jp) and all downstream tasks (MMLU, CSQA, ARC, HellaSwag). The results confirm that when the learned language is fed into the LLM, its neurons progressively acquired robust representations pertinent to each task. Figure~\ref{fig:subresult-probing-scatter} corresponds to Figure~\ref{fig:result-probing}b for OLMo-2/OLMo-0724 and all downstream tasks. We observe a similar tendency in both LLMs. Additionally, we examine the relationships between HellaSwag and the other three tasks, observing neurons that specialize in both tasks as well as neurons dedicated to a single task.

Summarizing these findings alongside our primary results (Section~\ref{sec:results-probing}), the neuron-wise probing accuracy for MMLU and ARC exhibits a remarkably high correlation, followed by a moderately positive correlation between HellaSwag and those two tasks (MMLU and ARC). By contrast, CSQA displays no correlation with any of the tasks. These observations suggest that the way each neuron in an LLM acquires its representations varies according to the nature of the task (e.g., required capabilities and answer formats).

\subsection{Changes in the nature of activations}
\begin{figure*}[t]
\centering
\includegraphics[width=1\linewidth]{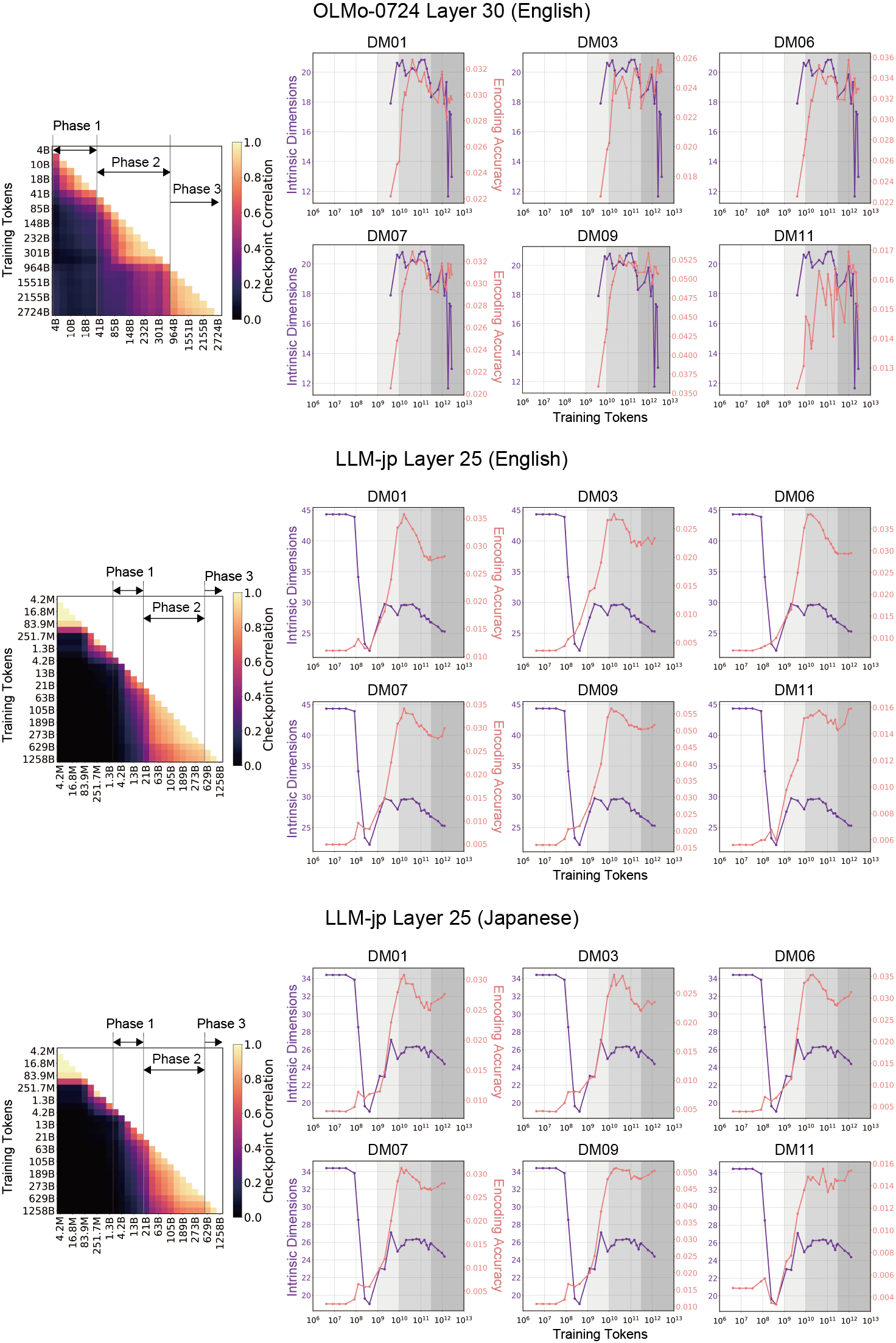}
\caption{Variations in correlation coefficients (left), encoding accuracy, and IDs (right) of the activations of OLMo-0724 (layer 30)/LLM-jp  (layer 25) using learned languages across checkpoints.}
\label{fig:subresult-ckpt-cc-id}
\end{figure*}

Figure~\ref{fig:subresult-ckpt-cc-id} shows additional results corresponding to those in Figure~\ref{fig:result-ckpt-cc-id} produced by the other LLMs (OLMo-0724, LLM-jp) when provided with the learned language data. When the number of training tokens exceeds 10\textsuperscript{9}, the changes in IDs throughout the training processes of the other LLMs exhibit a high correlation with variations in encoding accuracy, although some exhibited distinct patterns. By contrast, when the number of training tokens is less than 10\textsuperscript{9} (which is not possible for OLMo-2 or OLMo-0724), there is a precipitous drop from initially very high ID values in the activations of LLM-jp.

\subsection{Layers of interest}
\label{sec:additional_results:select-layer}
\begin{figure*}[t]
\centering
\includegraphics[width=1\linewidth]{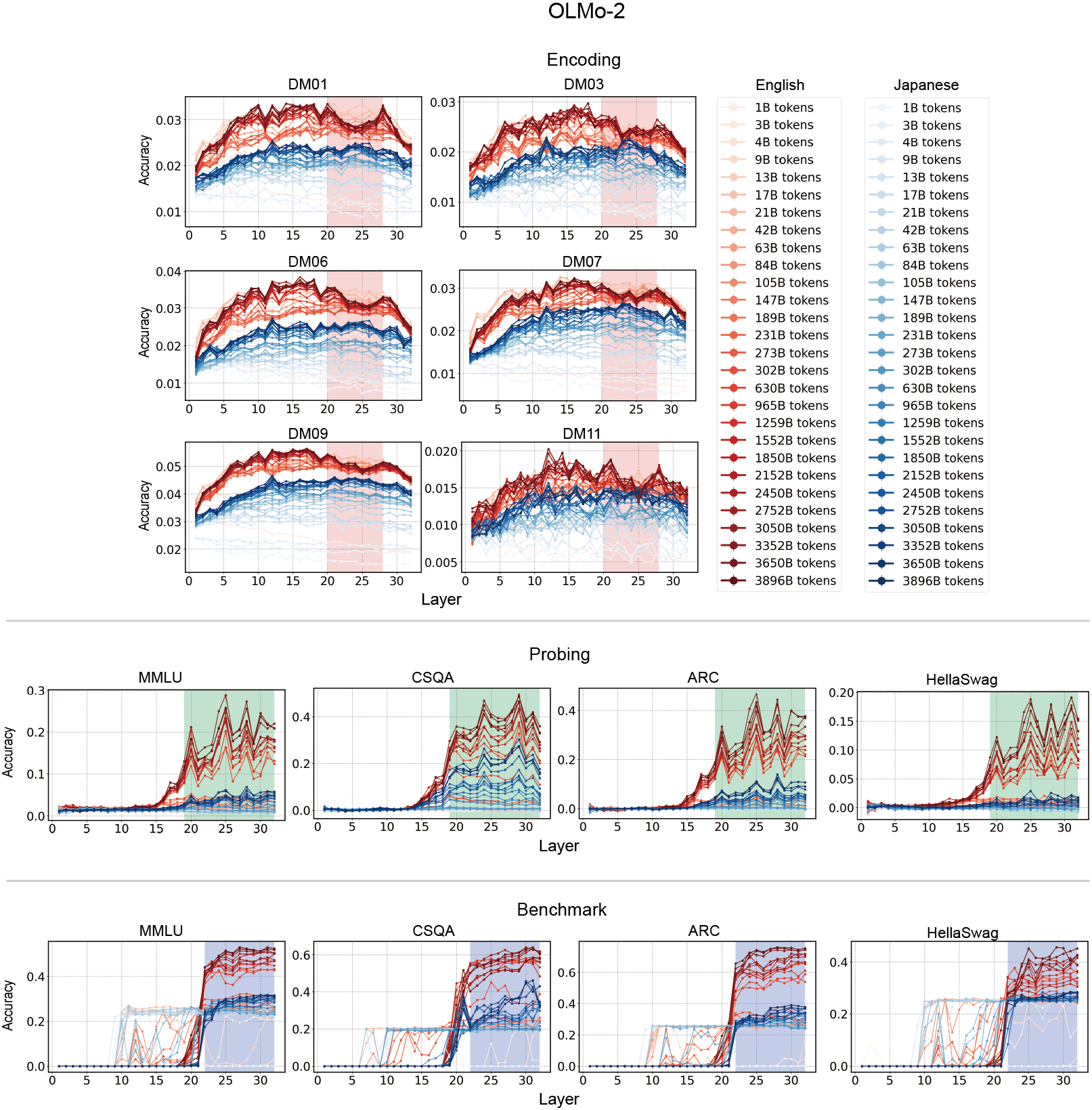}
\caption{Layers of interest for OLMo-2.}
\label{fig:subresult-layer-detail-olmo2}
\end{figure*}

\begin{figure*}[t]
\centering
\includegraphics[width=1\linewidth]{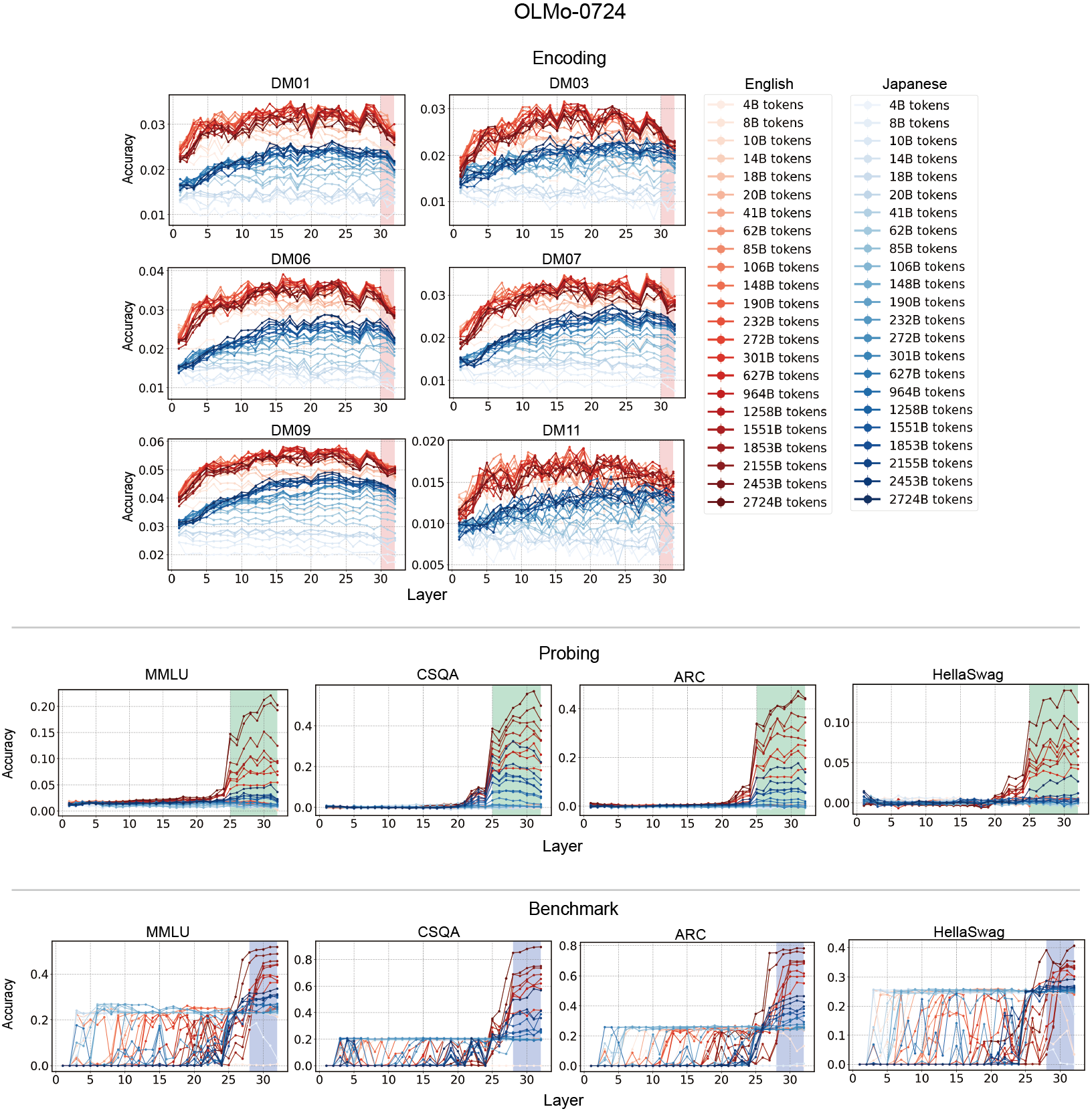}
\caption{Layers of interest for OLMo-0724.}
\label{fig:subresult-layer-detail-olmo0724}
\end{figure*}

\begin{figure*}[t]
\centering
\includegraphics[width=1\linewidth]{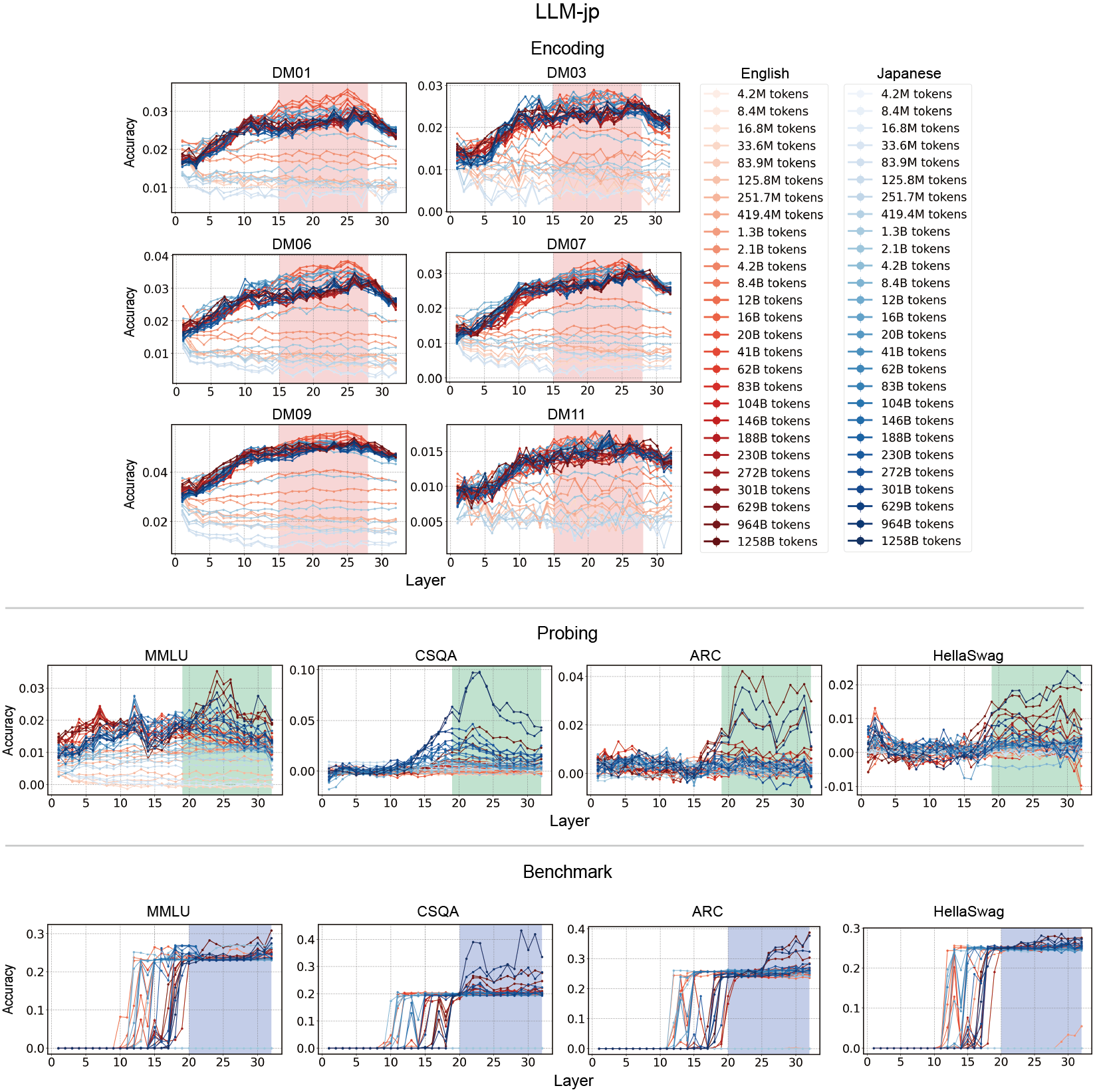}
\caption{Layers of interest for LLM-jp.}
\label{fig:subresult-layer-detail-llm-jp}
\end{figure*}

Figures~\ref{fig:subresult-layer-detail-olmo2}, \ref{fig:subresult-layer-detail-olmo0724}, and \ref{fig:subresult-layer-detail-llm-jp} show the layer-wise encoding, probing, and benchmark accuracies of OLMo-2, OLMo-0724, and LLM-jp at each training checkpoint, thereby determining which layers to examine in greater detail.

We can observe the phase transition in encoding accuracy (particularly the transition from Phase 1 to Phase 2) in layers 20--28 of OLMo-2, layers 30--32 of OLMo-0724, and layers 15--28 of LLM-jp. We can further identify the phase transition in probing accuracy (specifically the transition from Phase 2 to Phase 3) in layers 19--32 of OLMo-2, layers 25--32 of OLMo-0724, and layers 19--32 of LLM-jp. Finally, we can detect the phase transition in benchmark accuracy in layers 22--32 of OLMo-2, layers 28--32 of OLMo-0724, and layers 20--32 of LLM-jp. In this study, we focused on layers in which all three of these transitions emerge. Consequently, we confirmed this tendency in the later layers, namely in layers 22--28 for OLMo-2, layers 30--32 for OLMo-0724, and layers 20--28 for LLM-jp. Moreover, because OLMo-2 and LLM-jp exhibited the three transitions most prominently at layer 25, and the transitions were most prominent for OLMo-0724 at layer 30, we present the principal analytical results for these layers.

\end{document}